%% file: main.tex
\documentclass[10pt,twocolumn,letterpaper]{article}

\usepackage[pagenumbers]{cvpr}

\input{preamble}

\definecolor{cvprblue}{rgb}{0.21,0.49,0.74}
\usepackage[pagebackref,breaklinks,colorlinks,allcolors=cvprblue]{hyperref}

\def\confName{CVPR}
\def\confYear{2027}

\title{Disagree to Prune: Reallocating Visual Tokens Where a Small VLM Dissents}

\author{
Yanchao Li$^{1,2}$\quad
Jiaqing Xie$^{3,2}$\quad
Ben Gao$^{4}$\quad
Yusu Zhang$^{5}$\quad
Beilun Wang$^{6}$\\
Yanbo Wang$^{7}$\quad
Xiaohua Xu$^{8}$\quad
Zhongyu Huang$^{2}$\quad
Tianfan Fu$^{1,2,*}$\quad
Nanqing Dong$^{2,9,*}$\\[0.5em]
{\small $^{1}$Nanjing University\quad
$^{2}$Shanghai Artificial Intelligence Laboratory\quad
$^{3}$Fudan University\quad
$^{4}$Wuhan University}\\
{\small $^{5}$Shanghai Jiao Tong University\quad
$^{6}$Southeast University\quad
$^{7}$North University of China}\\
{\small $^{8}$University of Science and Technology of China\quad
$^{9}$Shanghai Innovation Institute}\\[0.3em]
{\small $^{*}$Corresponding authors}
}

\begin{document}
\maketitle
\input{sections/0_abstract}
\input{sections/1_intro}
\input{sections/2_related}
\input{sections/3_method}
\input{sections/4_experiments}
\input{sections/5_conclusion}
\input{sections/6_llm_usage}
{
    \small
    \bibliographystyle{ieeenat_fullname}
    \bibliography{ref}
}

\clearpage
\onecolumn
{\centering\Large\textbf{\thetitle}\\\vspace{0.5em}Supplementary Material\\\vspace{1.0em}\par}
\input{sections/9_appendix}

\input{sections/9b_appendix_exp}

\end{document}

%% file: preamble.tex
\usepackage{cuted} %

\usepackage{currfile} %

\usepackage{caption} %
\usepackage{amsmath,amssymb,amsthm}
\usepackage{dsfont}
\usepackage{booktabs}
\usepackage{multirow}
\newtheoremstyle{named}{\topsep}{\topsep}{\itshape}{0pt}{\bfseries}{.}{.5em}{\thmname{#1}\thmnumber{ #2}\thmnote{. #3}}
\theoremstyle{named}
\newtheorem{proposition}{Proposition}
\newtheorem{corollary}{Corollary}

\usepackage{colortbl}
\definecolor{oursbg}{HTML}{FBEAE5}
\usepackage{etoolbox}
\usepackage{placeins} %
\newcommand{\tabunderline}[1]{\setbox0\hbox{#1}\rlap{\rule[-0.9pt]{\wd0}{0.35pt}}\box0}
\AtBeginEnvironment{table}{\let\underline\tabunderline}
\AtBeginEnvironment{table*}{\let\underline\tabunderline}
\usepackage{algorithm}
\usepackage{algpseudocode}

\makeatletter
\newbox\disagree@labels
\let\disagree@ShipoutPicture\@ShipoutPicture
\renewcommand{\@ShipoutPicture}{%
  \global\let\disagree@ifinlabel\if@inlabel
  \global\setbox\disagree@labels\box\@labels
  \global\@inlabelfalse
  \disagree@ShipoutPicture
  \global\setbox\@labels\box\disagree@labels
  \global\let\if@inlabel\disagree@ifinlabel}
\makeatother

%% file: sections/0_abstract.tex
\begin{abstract}
Visual token pruning speeds up large vision-language models (VLMs) by keeping only a part of their visual tokens.
Most existing methods improve the ranking of the tokens and then answer every question once with the same budget.
With strong rankings, pruning hurts only a few questions, and signals computed on the kept tokens see little of what it removed.
We find that, on most of these questions, a small VLM given the full image disagrees with the pruned answer, since it still sees the missing evidence.
We propose PRUDENT, which first answers with about half the budget and spends the saved tokens on re-answering the disagreeing questions at twice the budget.
We show when such re-answering pays and why the small VLM should read the full image.
On 18 benchmarks, PRUDENT outperforms eight existing pruning methods at the same average budget of visual tokens.
Its lead widens as the budget tightens, and holds across five InternVL sizes.
PRUDENT also has the highest average in two more model families.
The disagreement of the small VLM flags most pruning failures of all eight methods, and re-answering on it also improves the two strongest with their own rankings.
Other checkers give similar gains, and the extra tokens go where the first pass hurts the answer.
These results suggest that identifying which pruned answers are trustworthy provides an additional lever for visual token pruning beyond the ranking of tokens.
Code and data are available at \href{https://github.com/LiZaiyuan0619/PRUDENT}{\nolinkurl{github.com/LiZaiyuan0619/PRUDENT}}.
\end{abstract}

%% file: sections/1_intro.tex
\section{Introduction}
\label{sec:intro}
\begin{figure}[t]
\centering
\includegraphics{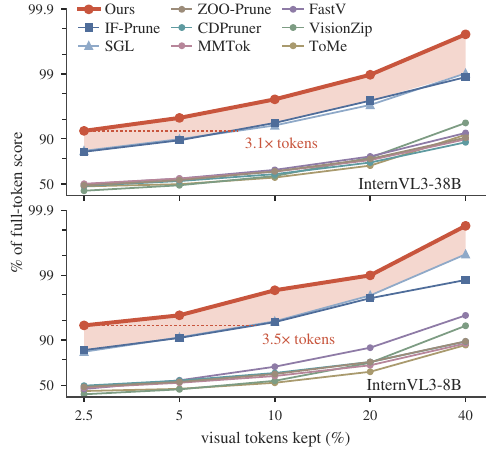}
\caption{\textbf{PRUDENT (Ours) and eight existing pruning methods.} Each curve averages the five benchmarks that the IF-Prune and SGL papers both report. The vertical axis is logarithmic in the share lost, and Appendix~\ref{app:main} plots each benchmark on a linear axis.}
\label{fig:hero}
\end{figure}
Large vision-language models (VLMs) encode an image into hundreds or thousands of visual tokens, which take most of the computation of answering a question~\cite{3666122.3667638,zhu2025internvl3,bai2025qwen3}.
Visual token pruning keeps only part of them and is a common way to cut this cost~\cite{fastv,visionzip,cdpruner,mmtok}.
Existing methods mostly refine how tokens are ranked, using attention, similarity and diversity, a learned importance, or a small VLM.

Better rankings have narrowed the gap, but at tight budgets, the pruned answers still lose much of the full-token score.
The loss is uneven. With a strong ranking, it falls on a few questions, mostly those whose answer rests on a small detail such as a word or digit.
Yet most existing methods spend the same budget on each question and keep its pruned answer, a practice we call one-pass pruning.
Detectors such as the large VLM's confidence read only the kept tokens, while the cause of a failure lies in the pruned part.
What is missing is a way to find these questions at test time, so that they get more tokens at the same budget.

We find that a small VLM that reads the full image can point them out.
When pruning removes the evidence, the large VLM tends to answer wrongly, while the small VLM still sees the evidence and often answers correctly.
The two answers then disagree, and they mostly agree when the evidence survives.
Combined with the large VLM's own confidence, this disagreement singles out pruning failures better than the confidence alone.
The small VLM is weaker than the large one, but it sees what pruning removed.

We turn this disagreement into \textbf{PRUDENT} (\textbf{PRU}ning with \textbf{D}iss\textbf{ENT}), which gives each question its own share of a fixed average budget.
PRUDENT first answers every question at about half the budget, and a small VLM, the checker, answers it from the full image.
The large VLM re-answers the questions with a disagreement at twice the budget, up to a quota set by the average budget.
The large VLM gives every final answer, and PRUDENT without re-answers is one-pass pruning.
We show that re-answering pays only when it repairs more than the first pass gives up, and that random re-answering does not.
Detectors on the kept tokens cannot tell apart two images that share these tokens, while the checker's disagreement errs at most as often as the checker departs from the full-token answer.
Under two monotone conditions, the best order by the disagreement and the confidence puts the disagreeing questions first, each group from low to high confidence.

Against eight representative baselines, PRUDENT scores highest on every benchmark in the main tables, with two ties, and its margin widens at tighter budgets (Figure~\ref{fig:hero}).
PRUDENT thus lets a fixed budget follow the questions that need more tokens.

Our main contributions are summarized as follows.
\begin{itemize}
\item We identify why detectors on the kept tokens miss pruning failures, and find that the disagreement of a small VLM that reads the full image flags most of them for each of eight existing methods.
\item We propose PRUDENT, which re-answers the questions with this disagreement at twice the budget and keeps the average budget of one-pass pruning.
\item We show in three propositions when re-answering is worth its cost, what information a checker needs, and how to order the re-answers, each supported by measurements.
\item We evaluate PRUDENT on up to 18 benchmarks across 5 model sizes and 3 model families. It outperforms eight existing methods and lifts the two strongest of them with their own rankings.
\item We trace its gain to the questions that pruning hurts, on which the checker disagrees and the re-answers keep the evidence.
\end{itemize}

%% file: sections/2_related.tex
\section{Related Work}
\label{sec:related}
\input{sections/figs_method}

Visual token pruning methods differ mainly in how they rank the tokens.
Some rank them by the attention inside the large VLM or its vision encoder~\cite{fastv,visionzip,pmlr-v267-zhang25s,xing2024pyramiddrop,11445607,zhu2026hawk}.
Others select or merge tokens by their similarity or diversity~\cite{tome,wen-etal-2025-stop,11095009,cdpruner,mmtok,11444044}.
Still others use the sensitivity of the output to each token~\cite{zooprune} or an importance estimate learned on a small model, as in IF-Prune~\cite{ifprune}.
Small VLM Guidance for accelerating Large VLMs (SGL) ranks the tokens by the attention of a small VLM~\cite{sgl}.
These methods give every question the same budget and keep the pruned answer.
PRUDENT answers some questions again with more tokens, keeping the average budget.

Another research line adapts the computation to each input.
Some methods choose a budget or a resolution for each image and question before answering~\cite{Ye_2025_CVPR,zhang2025trimtokenatoradaptivevisualtoken,kimhi-etal-2026-cares,li-etal-2026-sieveserve,cai2025matryoshka,he2026energy,li2026occamtoken}, and others let the model zoom into regions it selects~\cite{zhang2025adaptivechainoffocusreasoningdynamic,zhang2025mllms,Lin_2026_CVPR}.
Cascades pass an input to a larger model when cheaper models disagree or are unsure~\cite{kolawole2025agreementbasedcascadingefficientinference,soiffer-etal-2025-semantic,3524938.3525594,yue2024large,aggarwal2024automix}.
SGL also lets its small VLM answer alone when confident.
Speculative decoding lets a large model verify the draft of a small one~\cite{3618408.3619203,huang2025specvlmfastspeculativedecoding}.
These methods decide from the input or from models that read the same input.
PRUDENT decides after the first answer of the large VLM, from the disagreement between two views of the same question, the tokens kept for the large VLM and the full image read by the checker.
The checker only selects the questions that the large VLM answers again.

The loss from compressing a network concentrates on a few examples~\cite{hooker2021compresseddeepneuralnetworks}.
Recent work abstains when compressing a context removes the evidence of an answer~\cite{khodabandehlou-krishnamachari-2026-compression}.
Disagreement among models or samples is a common signal of error~\cite{3295222.3295387,manakul2023selfcheckgpt,kuhn2023semantic,ding2026llmsagreerightauditing}.
PRUDENT reads the disagreement between the two views as a sign that pruning removed the evidence, and re-answers such questions with more tokens.

%% file: sections/figs_method.tex
\begin{figure}[t]
\centering
\includegraphics{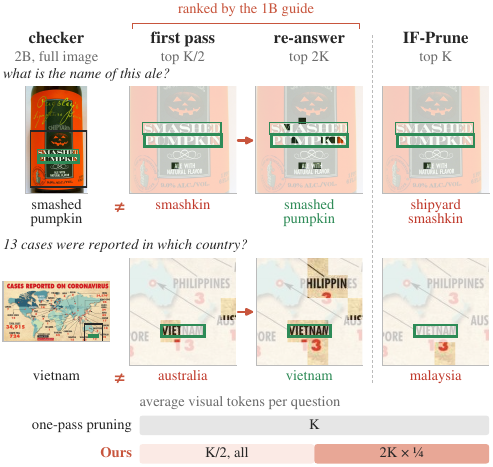}
\caption{\textbf{PRUDENT and IF-Prune on two questions.} First answer and checker disagree, so the large VLM re-answers at $2K$. IF-Prune answers at $K$. Green boxes mark the evidence.}
\label{fig:b1}
\end{figure}
\begin{figure*}[t]
\centering
\includegraphics{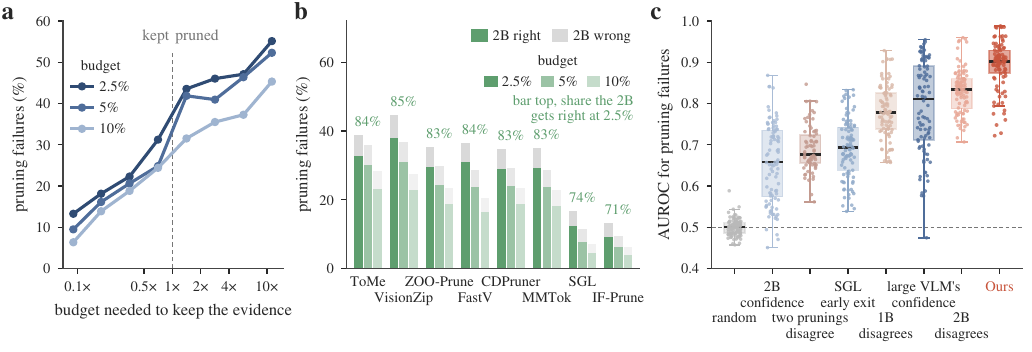}
\caption{\textbf{Pruning failures and their detection.} Failures on InternVL3-38B and 8B, (a) of IF-Prune on 6 text-rich benchmarks and (b) of each existing method on 16 benchmarks. (c) AUROC of eight signals on PRUDENT's first pass, one point per setting (Appendix~\ref{app:findings}).}
\label{fig:finding}
\end{figure*}
\begin{figure*}[t]
\centering
\includegraphics{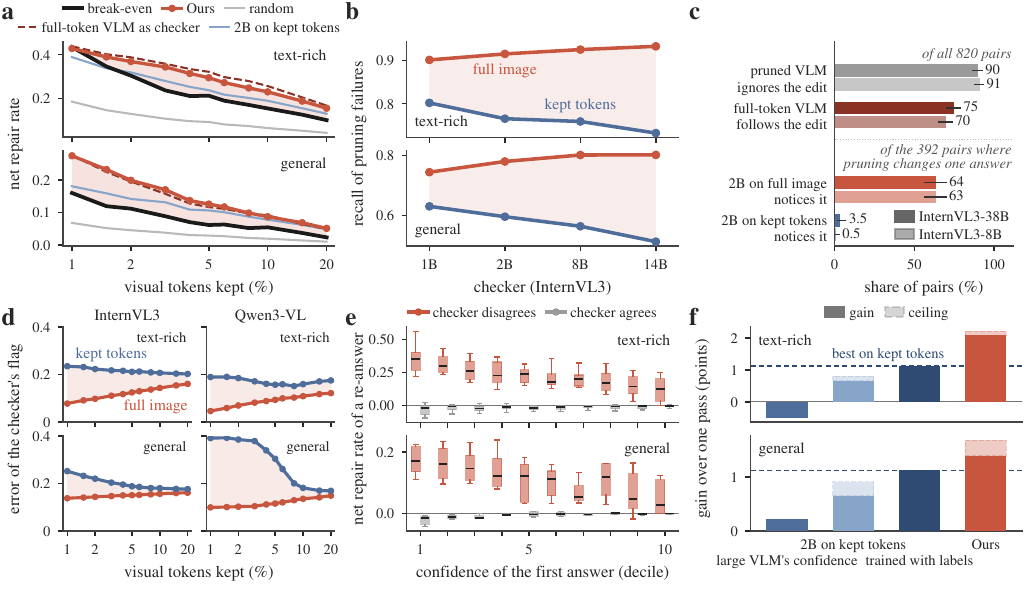}
\caption{\textbf{Measurements for Section~\ref{sec:theory}.} Means over InternVL3-38B, 8B and each group's benchmarks. (a) Net repair rates and the threshold of Proposition~\ref{prop:threshold} (break-even). (b) Recall at 2.5\%. (c) Twins made by editing only the pruned region (Appendix~\ref{app:twins}). (d) Error of the checker's flag. (e) Net repair rate by confidence. (f) Gain and ceiling at 2.5\% by the information an order uses (Appendices~\ref{app:grid} and~\ref{app:theory-more}).}
\label{fig:theory}
\end{figure*}
\input{sections/tab_main_38b}
\input{sections/tab_main_8b}

%% file: sections/tab_main_38b.tex
\begin{table*}[t]
\centering\scriptsize\setlength{\tabcolsep}{1.42pt}%
\renewcommand{\arraystretch}{0.9}
\caption{\textbf{PRUDENT (Ours) and existing methods on InternVL3-38B.} GQA and VQAv2 omit questions where full tokens give a stock phrase, and PRUDENT also leads with them (Appendix~\ref{app:setup}). \textbf{Best} in bold, \protect\underline{second}{} underlined.}
\label{tab:main-38b}
\begin{tabular}{@{}lccccccccccccccccccc@{}}
\toprule
 & \multicolumn{6}{c}{Text-rich} & \multicolumn{10}{c}{General} & \multicolumn{2}{c}{Grounding} & \\
\cmidrule(lr){2-7}\cmidrule(lr){8-17}\cmidrule(lr){18-19}
Method & ChartQA & DocVQA & InfoVQA & TextVQA & OCRB. & WebSRC & GQA & VQAv2 & OK-VQA & MMB & MMB-CN & MME & POPE & SEED & AI2D & SQA-I & RefC-A & RefC-B & Avg. \\
\midrule
\rowcolor{gray!12} Full tokens & 88.9 & 93.4 & 77.4 & 83.7 & 85.4 & 94.3 & 67.5 & 83.4 & 58.0 & 89.2 & 88.8 & 2466 & 90.8 & 79.2 & 87.8 & 99.4 & 94.6 & 88.5 & 100.0 \\
\midrule
\multicolumn{20}{@{}l}{\textit{Retain 2.5\% of the visual tokens}} \\
ToMe & 19.3 & 14.8 & 29.4 & 24.9 & 5.5 & 24.3 & 55.4 & 62.0 & 28.5 & 70.7 & 70.4 & 1780 & 82.7 & 62.2 & 69.1 & 82.1 & 14.0 & 11.1 & 51.8 \\
VisionZip & 17.7 & 12.9 & 29.0 & 15.4 & 4.8 & 19.7 & 48.2 & 51.7 & 16.5 & 53.2 & 52.7 & 1561 & 73.3 & 56.8 & 68.3 & 78.4 & 12.7 & 8.7 & 44.0 \\
ZOO-Prune & 25.5 & 17.2 & 27.9 & 24.2 & 5.8 & 25.7 & 55.7 & 63.8 & 29.3 & 70.6 & 68.9 & 1812 & 85.0 & 64.9 & 70.0 & 84.6 & 13.6 & 11.6 & 53.0 \\
FastV & 22.5 & 19.5 & 30.0 & 41.2 & 17.0 & 28.9 & 51.8 & 57.8 & 31.0 & 67.8 & 68.0 & 1857 & 76.3 & 56.7 & 71.6 & 82.4 & 15.3 & 11.6 & 53.4 \\
CDPruner & 24.7 & 16.4 & 28.0 & 28.2 & 16.4 & 28.8 & 52.5 & 58.2 & 35.2 & 68.3 & 67.3 & 1798 & 78.7 & 57.5 & 69.1 & 80.8 & 13.3 & 11.2 & 52.5 \\
MMTok & 22.2 & 15.8 & 29.4 & 36.8 & 15.6 & 24.2 & 58.4 & 67.4 & 29.1 & 71.6 & 70.1 & 1931 & 86.2 & 66.6 & 70.8 & 80.0 & 17.7 & 12.4 & 55.3 \\
SGL & \underline{60.9} & \underline{81.5} & \underline{54.4} & \underline{77.2} & 48.1 & 68.5 & 60.9 & 70.4 & 42.7 & 75.9 & 74.0 & 1888 & 83.0 & 70.0 & 74.3 & 84.4 & 65.3 & \underline{57.2} & 79.1 \\
IF-Prune & 52.2 & 77.2 & 52.5 & 75.6 & \underline{65.4} & \underline{71.7} & \underline{63.9} & \underline{74.2} & \underline{44.8} & \underline{84.0} & \underline{82.7} & \underline{2293} & \underline{89.5} & \underline{75.3} & \underline{76.0} & \underline{90.1} & \underline{68.9} & 56.6 & \underline{83.4} \\
\rowcolor{oursbg} \textbf{Ours} & \textbf{71.9} & \textbf{85.5} & \textbf{57.6} & \textbf{80.8} & \textbf{66.3} & \textbf{78.5} & \textbf{65.9} & \textbf{77.7} & \textbf{47.5} & \textbf{85.3} & \textbf{84.1} & \textbf{2383} & \textbf{89.9} & \textbf{76.3} & \textbf{79.3} & \textbf{92.5} & \textbf{72.6} & \textbf{64.3} & \textbf{88.4} \\
\midrule
\multicolumn{20}{@{}l}{\textit{Retain 5\% of the visual tokens}} \\
ToMe & 22.1 & 17.8 & 30.5 & 31.5 & 6.2 & 27.4 & 57.4 & 64.6 & 35.4 & 73.4 & 72.1 & 1850 & 84.0 & 65.1 & 69.8 & 83.3 & 17.1 & 14.5 & 55.1 \\
VisionZip & 25.5 & 18.3 & 30.0 & 24.9 & 7.5 & 26.3 & 55.6 & 61.8 & 35.5 & 70.2 & 69.7 & 1842 & 82.6 & 64.8 & 70.9 & 84.8 & 18.3 & 12.7 & 54.2 \\
ZOO-Prune & 37.8 & 23.9 & 30.3 & 33.0 & 9.6 & 35.4 & 59.7 & 65.3 & 39.5 & 77.0 & 77.4 & 1990 & 88.1 & 69.1 & 72.5 & 88.4 & 20.6 & 18.2 & 60.1 \\
FastV & 28.0 & 26.9 & 31.3 & 55.2 & 31.2 & 36.6 & 56.1 & 65.1 & 44.2 & 75.8 & 76.5 & 1984 & 82.0 & 62.2 & 74.1 & 86.3 & 29.2 & 21.6 & 62.6 \\
CDPruner & 32.2 & 22.4 & 29.7 & 38.7 & 28.9 & 36.2 & 55.9 & 62.8 & 42.5 & 77.0 & 74.2 & 1956 & 83.0 & 62.0 & 70.6 & 84.5 & 18.7 & 15.3 & 59.4 \\
MMTok & 27.0 & 23.9 & 32.8 & 46.4 & 27.9 & 35.1 & 62.9 & 70.2 & 42.0 & 80.0 & 77.2 & 2088 & 88.7 & 70.9 & 73.6 & 85.4 & 26.7 & 21.6 & 63.6 \\
SGL & \underline{68.3} & \underline{85.0} & \underline{60.0} & \underline{79.7} & 60.7 & 76.8 & 63.7 & 74.8 & 50.5 & 81.9 & 80.9 & 2090 & 87.9 & 73.5 & 76.8 & 89.7 & 75.0 & 68.7 & 86.4 \\
IF-Prune & 63.4 & 82.7 & 59.2 & 79.0 & \underline{74.5} & \underline{79.5} & \underline{65.5} & \underline{76.1} & \underline{51.0} & \underline{85.6} & \underline{84.0} & \underline{2399} & \underline{89.8} & \underline{76.4} & \underline{79.0} & \underline{92.4} & \underline{80.0} & \underline{72.0} & \underline{89.5} \\
\rowcolor{oursbg} \textbf{Ours} & \textbf{78.2} & \textbf{88.4} & \textbf{63.9} & \textbf{82.1} & \textbf{76.8} & \textbf{84.0} & \textbf{66.8} & \textbf{79.6} & \textbf{52.7} & \textbf{86.1} & \textbf{85.1} & \textbf{2456} & \textbf{90.2} & \textbf{77.0} & \textbf{81.4} & \textbf{95.0} & \textbf{82.7} & \textbf{75.9} & \textbf{93.2} \\
\midrule
\multicolumn{20}{@{}l}{\textit{Retain 10\% of the visual tokens}} \\
ToMe & 30.0 & 26.2 & 33.2 & 48.9 & 21.9 & 37.4 & 61.8 & 70.2 & 47.0 & 80.2 & 79.7 & 2072 & 88.0 & 70.2 & 71.9 & 88.5 & 26.9 & 25.6 & 64.6 \\
VisionZip & 42.8 & 32.1 & 32.3 & 41.4 & 16.5 & 38.1 & 61.0 & 70.1 & 48.8 & 80.1 & 79.1 & 2020 & 87.3 & 70.9 & 72.9 & 89.2 & 33.3 & 23.2 & 65.1 \\
ZOO-Prune & 55.1 & 36.2 & 32.7 & 44.0 & 20.5 & 49.7 & 63.6 & 66.9 & 47.0 & 83.2 & 82.4 & 2097 & 89.3 & 73.3 & 75.1 & 92.4 & 35.6 & 30.4 & 68.9 \\
FastV & 38.9 & 38.8 & 35.2 & 68.2 & 44.8 & 47.7 & 60.7 & 72.5 & 52.9 & 82.2 & 81.6 & 2206 & 85.8 & 68.0 & 77.5 & 91.2 & 55.1 & 43.3 & 73.6 \\
CDPruner & 43.1 & 32.8 & 33.0 & 50.5 & 41.3 & 46.9 & 59.8 & 67.6 & 48.7 & 81.8 & 80.0 & 2084 & 85.8 & 66.2 & 72.8 & 90.0 & 29.3 & 25.5 & 67.5 \\
MMTok & 38.2 & 35.2 & 35.4 & 57.4 & 41.5 & 45.0 & 64.9 & 73.3 & 49.5 & 83.8 & 81.9 & 2212 & 89.8 & 73.7 & 77.1 & 91.3 & 43.0 & 34.5 & 71.8 \\
SGL & 74.2 & \underline{87.9} & 64.9 & 81.4 & 73.4 & 82.4 & 66.1 & \underline{78.8} & 55.0 & 85.8 & 85.3 & 2201 & 89.6 & 76.4 & 80.4 & 94.9 & 85.6 & \underline{79.9} & 92.4 \\
IF-Prune & \underline{76.5} & 87.1 & \underline{65.5} & \underline{81.5} & \underline{81.2} & \underline{85.2} & \underline{66.4} & \underline{78.8} & \underline{55.3} & \underline{86.2} & \underline{85.4} & \underline{2458} & \underline{90.2} & \underline{77.0} & \underline{82.3} & \underline{95.0} & \underline{87.4} & 78.4 & \underline{94.1} \\
\rowcolor{oursbg} \textbf{Ours} & \textbf{83.8} & \textbf{91.1} & \textbf{68.2} & \textbf{82.9} & \textbf{82.5} & \textbf{87.8} & \textbf{67.0} & \textbf{81.2} & \textbf{56.2} & \textbf{87.0} & \textbf{86.4} & \textbf{2477} & \textbf{90.3} & \textbf{77.7} & \textbf{83.9} & \textbf{96.5} & \textbf{89.7} & \textbf{82.3} & \textbf{96.4} \\
\bottomrule
\end{tabular}
\end{table*}

%% file: sections/tab_main_8b.tex
\begin{table*}[t]
\centering\scriptsize\setlength{\tabcolsep}{1.42pt}%
\renewcommand{\arraystretch}{0.9}
\caption{\textbf{PRUDENT (Ours) and existing methods on InternVL3-8B.} Columns and marks as in Table~\ref{tab:main-38b}.}
\label{tab:main-8b}
\begin{tabular}{@{}lccccccccccccccccccc@{}}
\toprule
 & \multicolumn{6}{c}{Text-rich} & \multicolumn{10}{c}{General} & \multicolumn{2}{c}{Grounding} & \\
\cmidrule(lr){2-7}\cmidrule(lr){8-17}\cmidrule(lr){18-19}
Method & ChartQA & DocVQA & InfoVQA & TextVQA & OCRB. & WebSRC & GQA & VQAv2 & OK-VQA & MMB & MMB-CN & MME & POPE & SEED & AI2D & SQA-I & RefC-A & RefC-B & Avg. \\
\midrule
\rowcolor{gray!12} Full tokens & 86.6 & 91.7 & 71.2 & 80.9 & 84.2 & 90.8 & 62.9 & 79.7 & 49.0 & 85.9 & 84.4 & 2384 & 91.0 & 78.1 & 84.1 & 98.4 & 94.3 & 86.5 & 100.0 \\
\midrule
\multicolumn{20}{@{}l}{\textit{Retain 2.5\% of the visual tokens}} \\
ToMe & 16.5 & 10.2 & 24.1 & 18.2 & 4.4 & 18.2 & 46.6 & 54.8 & 21.5 & 59.1 & 56.7 & 1534 & 77.2 & 56.5 & 66.5 & 77.5 & 11.3 & 9.6 & 46.5 \\
VisionZip & 14.0 & 9.5 & 24.6 & 11.5 & 3.0 & 17.4 & 41.6 & 47.0 & 10.2 & 44.2 & 43.6 & 1343 & 68.5 & 52.0 & 66.4 & 74.3 & 9.7 & 8.7 & 40.0 \\
ZOO-Prune & 27.4 & 14.7 & 23.1 & 25.1 & 9.1 & 24.6 & 50.7 & 62.0 & 29.3 & 66.9 & 66.2 & 1728 & 86.0 & 63.9 & 67.6 & 86.1 & 41.6 & 39.6 & 57.2 \\
FastV & 15.7 & 12.4 & 23.9 & 31.3 & 16.9 & 19.7 & 47.8 & 59.4 & 31.4 & 60.7 & 58.8 & 1715 & 77.4 & 53.4 & 67.8 & 81.5 & 20.3 & 12.1 & 51.4 \\
CDPruner & 26.0 & 14.6 & 24.4 & 30.2 & 21.4 & 26.7 & 52.2 & 62.9 & 37.9 & 70.4 & 67.4 & 1833 & 82.4 & 60.9 & 67.4 & 84.6 & 30.3 & 29.0 & 58.4 \\
MMTok & 20.2 & 13.1 & 23.7 & 25.6 & 12.1 & 20.7 & 53.2 & 65.4 & 31.4 & 65.1 & 60.9 & 1793 & 84.5 & 63.1 & 66.7 & 79.6 & 15.4 & 14.6 & 53.4 \\
SGL & \underline{61.2} & \underline{80.4} & \underline{51.8} & \underline{74.6} & 48.0 & 66.7 & 56.9 & 70.9 & 38.1 & 71.7 & 68.6 & 1804 & 83.6 & 70.4 & 72.3 & 85.6 & 81.6 & 69.7 & 81.9 \\
IF-Prune & 60.0 & 72.5 & 45.8 & 73.3 & \underline{65.8} & \underline{68.0} & \underline{60.4} & \underline{74.9} & \underline{46.0} & \underline{80.8} & \underline{78.4} & \underline{2242} & \underline{90.1} & \underline{74.7} & \underline{74.0} & \underline{90.0} & \underline{86.6} & \underline{78.4} & \underline{87.7} \\
\rowcolor{oursbg} \textbf{Ours} & \textbf{76.1} & \textbf{82.9} & \textbf{53.1} & \textbf{78.4} & \textbf{67.6} & \textbf{76.1} & \textbf{62.6} & \textbf{77.2} & \textbf{46.2} & \textbf{82.3} & \textbf{79.4} & \textbf{2304} & \textbf{90.6} & \textbf{76.0} & \textbf{77.0} & \textbf{92.9} & \textbf{89.1} & \textbf{80.4} & \textbf{92.3} \\
\midrule
\multicolumn{20}{@{}l}{\textit{Retain 5\% of the visual tokens}} \\
ToMe & 17.9 & 12.8 & 26.2 & 23.6 & 4.6 & 21.7 & 50.2 & 59.6 & 27.3 & 63.8 & 62.0 & 1649 & 80.6 & 60.2 & 66.7 & 79.8 & 14.0 & 12.8 & 50.7 \\
VisionZip & 18.4 & 12.1 & 26.8 & 18.9 & 4.9 & 20.6 & 50.7 & 59.3 & 30.0 & 64.3 & 61.9 & 1656 & 76.7 & 60.2 & 67.6 & 76.9 & 18.6 & 11.6 & 50.6 \\
ZOO-Prune & 36.2 & 22.1 & 25.5 & 33.4 & 18.3 & 31.2 & 54.2 & 67.3 & 34.6 & 74.2 & 72.1 & 1901 & \underline{88.8} & 68.8 & 69.5 & 89.7 & 62.2 & 54.9 & 65.6 \\
FastV & 27.6 & 19.8 & 24.1 & 54.4 & 30.4 & 26.6 & 54.4 & 69.3 & 41.5 & 72.7 & 71.8 & 1969 & 85.7 & 63.1 & 71.2 & 86.8 & 52.4 & 36.6 & 65.3 \\
CDPruner & 34.9 & 20.8 & 25.9 & 40.6 & 28.3 & 32.9 & 56.0 & 68.4 & 42.7 & 75.3 & 73.4 & 1964 & 86.3 & 65.2 & 68.1 & 87.6 & 45.8 & 40.6 & 65.7 \\
MMTok & 23.8 & 19.5 & 27.0 & 38.0 & 21.5 & 27.8 & 58.0 & 71.0 & 39.8 & 73.1 & 70.3 & 1969 & 88.7 & 68.4 & 69.3 & 83.1 & 27.7 & 25.5 & 61.9 \\
SGL & 69.5 & \underline{83.4} & \underline{56.9} & \underline{78.1} & 60.0 & 74.7 & 60.3 & 74.7 & 44.1 & 78.4 & 74.8 & 2042 & 88.2 & 73.3 & 75.1 & 90.1 & 85.8 & 76.2 & 88.7 \\
IF-Prune & \underline{70.2} & 77.5 & 53.2 & 76.5 & \underline{74.7} & \underline{74.9} & \underline{61.4} & \underline{76.6} & \underline{46.9} & \underline{82.6} & \underline{79.6} & \underline{2328} & \textbf{90.6} & \underline{75.8} & \underline{76.7} & \underline{92.6} & \underline{90.6} & \underline{82.4} & \underline{92.0} \\
\rowcolor{oursbg} \textbf{Ours} & \textbf{78.5} & \textbf{85.5} & \textbf{58.4} & \textbf{79.9} & \textbf{76.3} & \textbf{81.3} & \textbf{63.0} & \textbf{78.5} & \textbf{47.8} & \textbf{83.0} & \textbf{80.8} & \textbf{2352} & \textbf{90.6} & \textbf{76.3} & \textbf{78.6} & \textbf{94.1} & \textbf{91.6} & \textbf{83.9} & \textbf{95.1} \\
\midrule
\multicolumn{20}{@{}l}{\textit{Retain 10\% of the visual tokens}} \\
ToMe & 25.0 & 18.8 & 28.6 & 39.6 & 22.5 & 30.3 & 55.9 & 69.0 & 38.2 & 73.8 & 72.6 & 1940 & 87.3 & 67.3 & 68.3 & 85.1 & 27.0 & 26.1 & 61.9 \\
VisionZip & 31.7 & 26.0 & 29.9 & 37.8 & 20.9 & 31.2 & 56.4 & 69.5 & 41.4 & 74.7 & 73.3 & 1921 & 84.0 & 67.7 & 68.8 & 84.4 & 34.4 & 23.6 & 63.3 \\
ZOO-Prune & 52.0 & 33.1 & 28.8 & 44.4 & 27.5 & 42.8 & 57.2 & 71.2 & 40.1 & 78.6 & 77.2 & 2045 & 89.6 & 72.6 & 73.3 & 91.9 & 77.9 & 68.2 & 74.2 \\
FastV & 51.2 & 34.9 & 28.1 & 69.6 & 44.7 & 38.2 & 59.1 & 75.8 & 45.7 & 81.1 & 78.5 & 2215 & 88.5 & 71.4 & 74.9 & 93.7 & 78.0 & 65.0 & 78.4 \\
CDPruner & 47.8 & 31.2 & 29.4 & 51.3 & 37.4 & 42.9 & 58.5 & 72.5 & 45.7 & 80.6 & 77.2 & 2092 & 88.8 & 68.9 & 69.8 & 90.5 & 62.4 & 55.0 & 73.7 \\
MMTok & 34.6 & 28.7 & 30.7 & 49.4 & 30.9 & 36.6 & 60.6 & 74.9 & 44.8 & 79.5 & 76.1 & 2097 & \underline{90.2} & 72.2 & 72.7 & 89.6 & 56.4 & 49.1 & 71.7 \\
SGL & 75.3 & \underline{85.9} & \underline{61.0} & \underline{80.4} & 70.7 & \underline{79.9} & 61.9 & 77.6 & 47.3 & 82.4 & 80.3 & 2185 & 90.1 & 75.7 & 77.8 & \underline{94.4} & 91.0 & 82.6 & 93.8 \\
IF-Prune & \underline{77.6} & 82.8 & 58.7 & 79.0 & \underline{79.8} & 79.3 & \underline{62.8} & \underline{77.9} & \underline{47.9} & \underline{82.6} & \underline{80.9} & \underline{2368} & \textbf{90.5} & \underline{76.0} & \underline{78.5} & 94.0 & \underline{92.0} & \underline{83.6} & \underline{94.9} \\
\rowcolor{oursbg} \textbf{Ours} & \textbf{83.3} & \textbf{88.4} & \textbf{64.3} & \textbf{81.2} & \textbf{81.8} & \textbf{84.9} & \textbf{63.2} & \textbf{78.9} & \textbf{48.6} & \textbf{84.2} & \textbf{81.2} & \textbf{2403} & \textbf{90.5} & \textbf{76.7} & \textbf{80.4} & \textbf{95.5} & \textbf{92.9} & \textbf{85.4} & \textbf{97.5} \\
\bottomrule
\end{tabular}
\end{table*}

%% file: sections/3_method.tex
\section{Method}
\label{sec:method}

\subsection{Preliminaries}
\label{sec:prelim}

To answer a question $q$ about an image, a large VLM encodes the image into $m$ visual tokens.
Visual token pruning orders them by a ranking $s_q$ and keeps the top fraction $k$, the kept tokens $S_q(k)$.
The large VLM then answers from them and the question alone.
We write its answer as $a_q(k)$ and its score as $y_q(k)$, with $k = 1$ for full tokens.
Existing pruning methods differ in how they build $s_q$ (Section~\ref{sec:related}).

The budget $K$ is the average fraction of visual tokens kept per question.
We call one model, one benchmark and one budget a setting.
Most existing methods use one-pass pruning, which answers every question once with the fraction $K$.
A question is a pruning failure at fraction $k$ when the large VLM answers correctly with full tokens and incorrectly with $S_q(k)$.
The confidence $\pi_q(k)$ is the length-normalized probability of the answer $a_q(k)$.

\subsection{PRUDENT}
\label{sec:prudent}

Many pruning failures are questions whose answer depends on a small detail that pruning has removed (Figure~\ref{fig:finding}a).
A small VLM that reads the full image still has this detail and often answers correctly.
On these questions, its answer differs from the pruned answer, and the difference is visible at test time without labels.
We call this small VLM the checker, write its answer as $c_q$ and define its disagreement with the large VLM at fraction $k$ as $\delta_q(k) = \mathds{1}[c_q \neq a_q(k)]$.
Two answers count as equal when they match under the answer format of the benchmark (Appendix~\ref{app:match}).
For each of the eight existing methods, the checker answers most of its pruning failures correctly (Figure~\ref{fig:finding}b).
Ordering questions by disagreement and then by confidence finds pruning failures better than confidence alone, with the highest AUROC on every benchmark (Figure~\ref{fig:finding}c and Appendix~\ref{app:findings}).

PRUDENT uses this signal to decide how many visual tokens each question receives (Figure~\ref{fig:b1} and Appendix~\ref{app:examples}).
In the first pass, the large VLM answers every question with the fraction $b$ and gives the first answer $a_q(b)$.
Questions on which the checker disagrees are then sorted by the confidence of their first answer, from low to high.
Up to a quota $\rho$, the large VLM re-answers the leading questions $R$ with the fraction $2K$, and the new answers replace the first ones.
Places left in the quota go to the agreeing questions with the lowest confidence.
Every final answer comes from the large VLM (Algorithm~\ref{alg:prudent} in Appendix~\ref{app:budget}).
Questions that arrive one at a time can be ranked among those seen so far, within the budget and at 0.1 to 0.3 points lower on average (Appendix~\ref{app:budget}).

The first pass spends the fraction $b$ on every question, and the re-answers spend $2K$ on a fraction $\rho$ of them, so that $b + 2\rho K = K$, the budget of one-pass pruning.
We set $b = K/2$ and $\rho = 1/4$, and $b = 1\%$ and $\rho = 0.3$ at 2.5\% (Appendix~\ref{app:budget}).
Counted in tokens, PRUDENT reads 98\% of the visual tokens of one-pass pruning on average (Appendix~\ref{app:budget}).
Section~\ref{sec:cost} measures the other costs of the checker and the re-answers.

Disagreement is a useful signal only when the answer of the checker is reliable.
The checker therefore abstains when the least probable token of its answer falls below a probability $\tau$, or when its answer is longer than $\ell$ words.
Long answers rarely match word for word, so their disagreement says little.
The large VLM then answers the question once with the fraction $K$.
We denote by $C$ the questions on which the checker does not abstain.
The quota $\rho$ is counted over $C$, so the average budget stays $K$.
PRUDENT differs from one-pass pruning only on $C$.
A re-answered question $q$ gains $u_q = y_q(2K) - y_q(b)$, and every question in $C$ gives up $h_q = y_q(K) - y_q(b)$ in the first pass.
Over the $n$ questions of a benchmark, PRUDENT gains
\begin{equation}
\Delta = \frac{1}{n}\Big[\sum_{q \in R} u_q - \sum_{q \in C} h_q\Big].
\label{eq:gain}
\end{equation}

The first pass and the re-answers use the same ranking $s_q$.
The ranking decides where the large VLM looks, and PRUDENT decides which questions it looks at again.
In our instance, a small VLM called the guide provides the ranking (Appendix~\ref{app:ranking}).
For open-ended questions, it averages the ranks given by its own attention while it answers the question~\cite{sgl} and by a learned estimate of token importance~\cite{ifprune}.
For multiple-choice and yes-or-no questions, where its attention says little about the evidence, it uses the learned estimate alone, as IF-Prune does.
PRUDENT also gains with other rankings (Section~\ref{sec:general}).

\subsection{Theoretical Analysis}
\label{sec:theory}

We show when re-answering pays, what a checker can see, and how to order re-answers (proofs in Appendix~\ref{app:proofs}).

\begin{proposition}[Threshold for re-answering]
\label{prop:threshold}
Let $P_R$ be the mean of $u_q$ over $R$, the net repair rate, and let $\bar{H}$ be the mean of $h_q$ over $C$.
For every $R \subseteq C$ with $|R| = \rho|C|$,
\begin{equation}
\Delta = \frac{|C|}{n}\,\rho\,\Big(P_R - \frac{\bar{H}}{\rho}\Big),
\label{eq:threshold}
\end{equation}
so re-answering gains exactly when $P_R$ exceeds the threshold $\bar{H}/\rho$.
When $R$ is drawn at random from $C$ and the mean one-pass score $S_C(k)$ over $C$ is concave and non-decreasing on $[b, 2K]$, the expected gain is at most $\frac{|C|}{n}\big[S_C(K - \rho b) - S_C(K)\big] \le 0$.
With scores in $\{0, 1\}$, every $R$ has $P_R \le \min(1, p_+/\rho)$.
Here $p_+$ is the share of $C$ answered wrongly at $b$ and correctly at $2K$.
An oracle that knows the labels attains this bound when $\rho \le 1 - p_-$, with $p_-$ the share of $C$ answered correctly at $b$ and wrongly at $2K$.
\end{proposition}

The first pass costs every checked question some score, which the re-answers must outweigh.
The threshold rises as the budget tightens and the one-pass curve steepens.
PRUDENT's net repair rate rises with it and exceeds it most near 1\% on the general benchmarks and 3\% on the text-rich ones (Figure~\ref{fig:theory}a).
The threshold and $p_+$ need only one-pass runs at $b$, $K$ and $2K$, so a labeled development set shows when re-answering pays.
A cut chosen on InternVL3 predicts it in 76 of 84 Qwen3-VL-8B settings (Appendix~\ref{app:grid}).

The flag $\delta^*_q = \mathds{1}[a_q(b) \neq a_q(1)]$ records whether pruning changed the answer of the large VLM, and $\gamma = P(c_q \neq a_q(1))$ is how often the checker disagrees with the full-token answer.
We write $V_q(k)$ for the kept tokens of fraction $k$ with their ranks and the question, and $V_q = V_q(b)$.

\begin{proposition}[Pruned and full views]
\label{prop:views}
A detector that depends only on $V_q$ gives the same output on two questions with the same $V_q$.
When pruning changes the answer to one of them and not to the other, the detector errs on one of the two.
Let a detector flag $c'_q \neq a_q(b)$ for a checker $c'$ that reads $V_q$.
Its recall and false alarm rate for $\delta^*$ are at most $P(c'_q \neq a_q(b))/P(\delta^*_q = 1)$ and $P(c'_q \neq a_q(b))/P(\delta^*_q = 0)$, respectively.
Both bounds shrink as $c'$ answers more like the large VLM.
If answer equality is an equivalence relation, a checker $c$ that reads the full image satisfies
\begin{equation}
P\big(\delta_q(b) \neq \delta^*_q\big) = \gamma - P(E_q) \le \gamma,
\label{eq:views}
\end{equation}
where $E_q$ is the event that $c_q$, $a_q(b)$ and $a_q(1)$ are pairwise distinct.
\end{proposition}

Answering like the large VLM moves a checker on the full image toward the full-token answer, but on the kept tokens, it repeats the large VLM's mistakes.
A checker should therefore read the full image and answer like the large VLM. Its size and family enter only through $\gamma$ (Figure~\ref{fig:theory}b and Appendix~\ref{app:theory-more}).

Rewriting only the pruned evidence rarely changes the pruned answer (Figure~\ref{fig:theory}c).
When pruning changes only one of the two answers, the checker separates the two images far more often on the full image than on the kept tokens.

\begin{corollary}[Nested views]
\label{cor:views}
For fractions $b < b'$, the input $V_q(b)$ is a function of $V_q(b')$.
Hence $I(Y; V_q(b)) \le I(Y; V_q(b'))$ for the mutual information with every quantity $Y$ of a question, such as its full-token answer.
The bound of Eq.~\eqref{eq:views} holds with the same $\gamma$ at every budget.
\end{corollary}

As the budget tightens, a detector on the kept tokens can only lose information, while the guarantee of a checker on the full image stays the same, so the full view's advantage should grow.
The share of failures with their evidence wholly pruned also grows, to about half at 1\% on text-rich benchmarks (Appendix~\ref{app:theory-more}).
The checker's flag on the full image errs less often at tighter budgets, but not on the kept tokens (Figure~\ref{fig:theory}d).
Re-answering by the full view then gains more over the kept tokens at tight budgets than at loose ones (Appendix~\ref{app:theory-more}).
Let $r(d, p) = \mathbb{E}[u_q \mid \delta_q(b) = d,\ \pi_q(b) = p]$ be the mean gain of a re-answer.

\begin{proposition}[Order and ceiling]
\label{prop:order}
Suppose that $r(1, p) \ge r(0, p')$ for every $p$ and $p'$, and that $r(d, p)$ does not increase in $p$ for either $d$.
Consider the rules that see only $(\delta, \pi)$ and re-answer the same number of questions.
Among them, the expected gain is maximal when the disagreeing questions come first and each group is sorted from low to high confidence.
For every information $I$, the re-answers of a rule that sees only $I$ repair in expectation at most the sum of the $\rho|C|$ largest values of $\mathbb{E}[u_q \mid I_q]$ over $C$.
We call this sum the ceiling of $I$.
\end{proposition}

A disagreeing question gains more from a re-answer than an agreeing one of the same confidence, and its gain falls as the confidence rises (Figure~\ref{fig:theory}e).
PRUDENT stays close to the ceiling of these two signals (Figure~\ref{fig:theory}f).
Even the full-token answer of the large VLM as the checker would raise this ceiling by only about half a point (Appendix~\ref{app:grid}).

Reading the full image raises this ceiling.
With a checker on the kept tokens, the ceiling of its disagreement and the confidence lies below PRUDENT's gain at tight budgets (Figure~\ref{fig:theory}f), and so does every rule on them (Proposition~\ref{prop:order}).
A classifier trained with labels on pruned-view signals falls further behind PRUDENT as budgets tighten, to half its gain on text-rich benchmarks at 2.5\% (Appendix~\ref{app:theory-more}).

\begin{figure*}[t]
\centering
\includegraphics{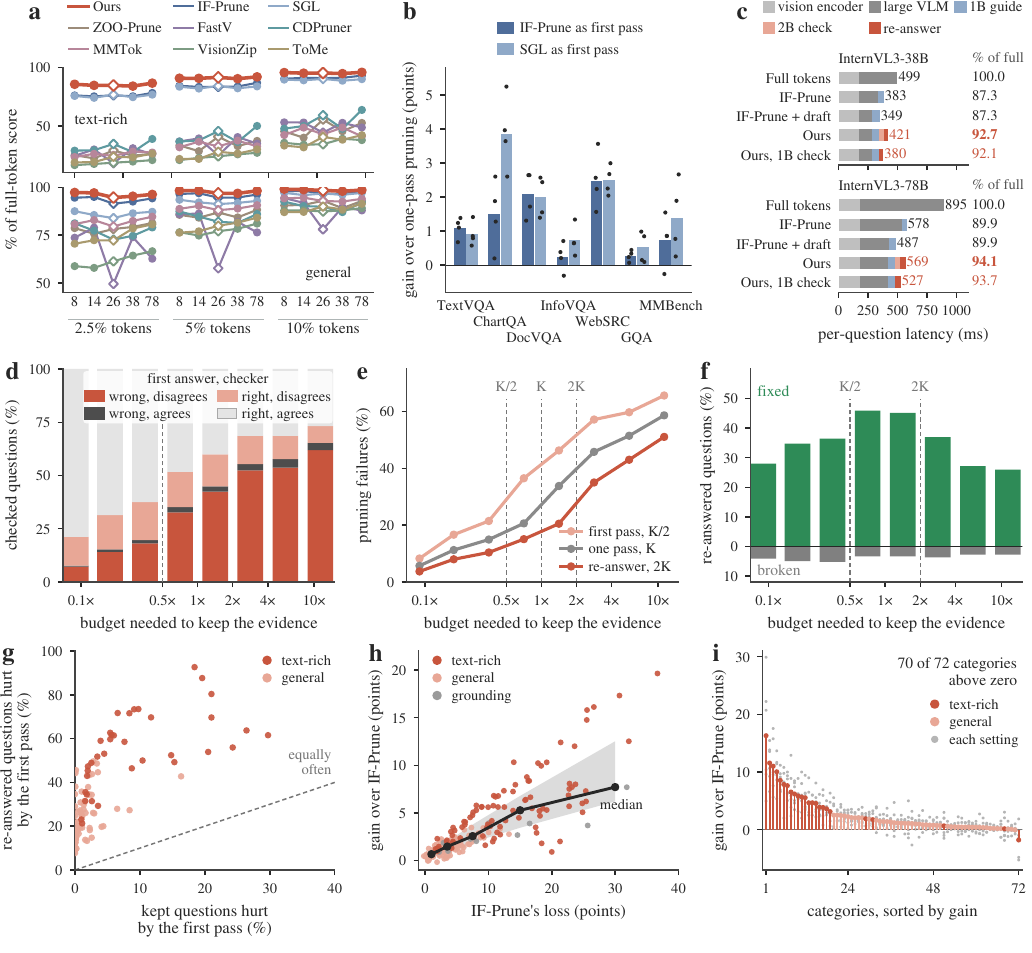}
\caption{\textbf{Generalization, cost and analysis.} (a) Five InternVL models on nine benchmarks, InternVL2.5-26B in open markers. (b) IF-Prune or SGL ranks in both passes. (c) Latency per question at 5\% on seven benchmarks. 1B check uses the guide as checker. (d--f) The evidence chain on text-rich benchmarks. (g, h) One point per setting. (i) Categories of at least 100 questions (Appendices~\ref{app:general}, \ref{app:cost} and~\ref{app:why}).}
\label{fig:analysis}
\end{figure*}
\begin{corollary}[Budget law]
\label{cor:budget}
With $b = K/2$ and $\rho = 1/4$, the threshold equals $4\big[S_C(K) - S_C(K/2)\big]$.
By Proposition~\ref{prop:threshold}, every $R$ also has $P_R \le \min(1, 4p_+)$, so the gain is at most $\rho|C|/n$ times the margin of $\min(1, 4p_+)$ over the threshold.
This margin vanishes as the budget loosens and few questions remain repairable.
If $b$ and the re-answer budget $b_2$ vary freely and $P_R$ changes little with $b$, the first-order condition sets the slope of the one-pass curve at $b$ equal to $P_R / b_2$.
\end{corollary}

%% file: sections/4_experiments.tex
\section{Experiments}
\input{sections/tab_extreme}
\label{sec:exp}

\subsection{Experimental Setup}
\label{sec:setup}

\paragraph{Models and benchmarks.}
The main experiments use InternVL3-38B and 8B as the large VLM~\cite{zhu2025internvl3}, from the family that IF-Prune and SGL use.
The checker is InternVL3-2B and the guide is InternVL2.5-1B~\cite{chen2024expanding}.
The 18 benchmarks (Appendix~\ref{app:setup}) are 6 text-rich ones, which require reading text in documents, charts, infographics, photos and web pages, 10 general ones on natural images and diagrams, and 2 grounding ones.

\paragraph{Baselines and protocol.}
We compare PRUDENT with ToMe~\cite{tome}, VisionZip~\cite{visionzip}, ZOO-Prune~\cite{zooprune}, FastV~\cite{fastv}, CDPruner~\cite{cdpruner}, MMTok~\cite{mmtok}, SGL~\cite{sgl} and IF-Prune~\cite{ifprune}, which cover every kind of ranking in Section~\ref{sec:related}.
All run in lmms-eval~\cite{zhang-etal-2025-lmms} on the same large VLM, each with its own ranking at the same budget as PRUDENT, SGL without its early exit, as in its released code.
Our run of IF-Prune's official code closely matches its paper (Appendix~\ref{app:setup}).
The main tables use budgets of 2.5\%, 5\% and 10\%, Table~\ref{tab:extreme} adds 1\% and Figure~\ref{fig:hero} extends to 40\%.
Each benchmark keeps its own metric, and we average the ratio of each score to the full-token score.
Decoding is greedy, and within each setting a paired bootstrap over questions compares methods.

\subsection{Main Results}
\input{sections/tab_family}
\input{sections/tab_ablation}
\label{sec:main}

\textbf{PRUDENT outperforms every existing method, and its lead grows at tighter budgets, where more evidence is pruned (Tables~\ref{tab:main-38b} and~\ref{tab:main-8b}).}
It scores highest on every benchmark at every budget on both models, with two ties on POPE.
Over these 108 settings, it is significantly better than each existing method in most and worse in none, also after a Holm correction (Appendix~\ref{app:main}).
It also leads from 2.5\% to 40\% in Figure~\ref{fig:hero}, and at 1\% on each text-rich benchmark at every model size (Table~\ref{tab:extreme} and Appendix~\ref{app:main}).
Text-rich questions gain the most, since their answers often rest on a few words or digits.
There, the lead over IF-Prune comes from both the guide's ranking and re-answering, which adds about two points from 2.5\% to 10\% with the same ranking (Table~\ref{tab:ablation}).
On multiple-choice and yes-or-no benchmarks, the rankings are the same, so the lead there comes from re-answering alone and is significant in most of these settings (Appendix~\ref{app:main}).
A low-resolution thumbnail alone takes more tokens than PRUDENT at 10\% and loses far more on text-rich benchmarks (Appendix~\ref{app:main}).

\subsection{Generalization and Cost}
\label{sec:general}
\label{sec:cost}

\textbf{PRUDENT's lead holds as the large VLM grows from 8B to 78B, and carries over to two other model families (Figure~\ref{fig:analysis}a and Table~\ref{tab:family}).}
On Qwen3-VL~\cite{bai2025qwen3} and Gemma~4~\cite{team2026gemma}, a small model of the same family is the guide and the checker, and PRUDENT shares SGL's ranking.
It beats SGL in most settings, so the gain comes from re-answering, and it has the highest average over each model's 10 benchmarks (Appendix~\ref{app:general}).

\textbf{Re-answering on disagreement also lifts the two strongest existing methods with their own rankings (Figure~\ref{fig:analysis}b).}
With the ranking of IF-Prune or SGL in both passes, PRUDENT beats the method alone significantly in most settings and loses significantly in none (Appendix~\ref{app:general}).

\textbf{PRUDENT adds a nearly fixed cost per question, mostly the checker's pass over the full image (Figure~\ref{fig:analysis}c).}
The large VLM decodes speculatively from the guide's answer~\cite{3618408.3619203}, as does IF-Prune + draft.
As for IF-Prune and SGL, whose small VLMs also read the full image, the budget counts the large VLM's visual tokens.
With the guide as checker, PRUDENT needs no model beyond IF-Prune + draft and about its FLOPs, and scores 4.8 and 3.8 points higher on 38B and 78B (Appendix~\ref{app:cost}).

\subsection{Ablations and Analysis}
\label{sec:ablation}
\label{sec:why}

\textbf{Re-answering pays when disagreement picks the questions, and each part of PRUDENT contributes (Table~\ref{tab:ablation}).}
Random re-answering loses to one-pass pruning, as Proposition~\ref{prop:threshold} predicts, and so does choosing by SGL's early-exit score, or by the large VLM's confidence on text-rich benchmarks.
Ordering the disagreeing questions by confidence adds a little (Proposition~\ref{prop:order}).
Changing $\tau$, $\ell$ or the checker moves group means by less than a point (Appendices~\ref{app:general} and~\ref{app:ablation}).

\textbf{The gain rests on evidence that the first pass prunes, the checker still sees, and the re-answer keeps (Figure~\ref{fig:analysis}d--f).}
We locate this evidence for text-rich questions whose answer appears in the image, and place each by the budget that keeps it, in units of $K$ (Appendix~\ref{app:why}).
Disagreement rises once the evidence is beyond the first pass (Figure~\ref{fig:analysis}d).
The re-answer pushes the onset of failures to $2K$, with far more fixes than breaks (Figure~\ref{fig:analysis}e--f).

\textbf{PRUDENT moves the budget toward the questions that the first pass hurts (Figure~\ref{fig:analysis}g).}
In every setting, first answers below the full-token answer are far more common among the re-answered questions than among the rest.

\textbf{PRUDENT gains more where one-pass pruning loses more, and in almost every category of questions (Figure~\ref{fig:analysis}h--i).}
It also loses fewer correct answers to pruning than every existing method (Appendix~\ref{app:why}).

%% file: sections/tab_extreme.tex
\begin{table}[tb]
\centering\scriptsize\setlength{\tabcolsep}{11.89pt}
\renewcommand{\arraystretch}{0.9}
\caption{\textbf{An extreme budget of 1\%.} Mean ratio to the full-token score on the six text-rich benchmarks (Appendix~\ref{app:main}).}
\label{tab:extreme}
\begin{tabular}{lccccc}
\toprule
Method & 8B & 14B & 26B & 38B & 78B \\
\midrule
ToMe & 15.5 & 16.1 & 19.1 & 18.5 & 21.2 \\
VisionZip & 14.9 & 15.5 & 16.5 & 15.5 & 17.1 \\
ZOO-Prune & 18.8 & 17.6 & 20.2 & 18.3 & 19.1 \\
FastV & 15.7 & 21.8 & 17.2 & 22.8 & 19.7 \\
CDPruner & 21.1 & 21.7 & 25.3 & 20.9 & 26.8 \\
MMTok & 17.6 & 16.9 & 19.1 & 20.6 & 20.8 \\
SGL & 61.0 & 58.8 & 62.4 & 60.2 & 63.3 \\
IF-Prune & \underline{64.4} & \underline{62.1} & \underline{64.0} & \underline{62.3} & \underline{65.3} \\
\rowcolor{oursbg} \textbf{Ours} & \textbf{73.6} & \textbf{71.0} & \textbf{72.5} & \textbf{71.1} & \textbf{73.2} \\
\bottomrule
\end{tabular}
\end{table}

%% file: sections/tab_family.tex
\begin{table*}[t]
\centering\scriptsize\setlength{\tabcolsep}{2.11pt}
\renewcommand{\arraystretch}{0.9}
\caption{\textbf{Other model families.} Each family uses the budget where SGL keeps about 90\% of the full-token score here. Avg.\ is over all 10 benchmarks (Appendix~\ref{app:general}). IF-Prune's estimator is for InternVL2.5-1B. ZOO-Prune is undefined on Gemma 4 (n/a).}
\label{tab:family}
\begin{tabular}{lcccccccccccccccccc}
\toprule
 & \multicolumn{6}{c}{Qwen3-VL-8B, 10\%} & \multicolumn{6}{c}{Qwen3-VL-32B, 10\%} & \multicolumn{6}{c}{Gemma 4 31B, 20\%} \\
\cmidrule(lr){2-7}\cmidrule(lr){8-13}\cmidrule(lr){14-19}
Method & TextVQA & ChartQA & DocVQA & GQA & MMB & Avg. & TextVQA & ChartQA & DocVQA & GQA & MMB & Avg. & TextVQA & ChartQA & DocVQA & GQA & MMB & Avg. \\
\midrule
\rowcolor{gray!12} Full tokens & 82.1 & 85.2 & 95.5 & 61.5 & 84.6 & 100.0 & 83.5 & 87.3 & 96.2 & 62.5 & 88.1 & 100.0 & 75.8 & 68.0 & 93.0 & 47.3 & 84.1 & 100.0 \\
ToMe & 36.2 & 25.2 & 28.3 & 51.5 & 64.9 & 57.2 & 39.5 & 21.2 & 30.2 & 54.4 & 73.7 & 59.5 & 60.8 & 39.5 & 40.3 & 44.7 & 82.7 & 76.1 \\
VisionZip & 49.1 & 35.1 & 44.1 & 49.5 & 65.6 & 60.3 & 50.5 & 33.1 & 40.1 & 53.2 & 77.1 & 63.1 & 62.6 & \textbf{65.4} & 60.0 & 44.4 & 83.6 & 86.6 \\
ZOO-Prune & 44.9 & 37.9 & 45.5 & 54.1 & 68.8 & 65.2 & 56.7 & 35.2 & 55.8 & 56.1 & 77.1 & 69.5 & n/a & n/a & n/a & n/a & n/a & n/a \\
FastV & 35.5 & 23.1 & 29.0 & 44.7 & 57.8 & 48.2 & 60.3 & 22.4 & 33.4 & 46.7 & 71.0 & 58.5 & 49.6 & 36.6 & 23.1 & 40.8 & 77.3 & 66.6 \\
CDPruner & 51.1 & 31.4 & 40.5 & 50.7 & 70.3 & 65.0 & 47.2 & 22.2 & 34.8 & 49.7 & 74.7 & 57.9 & 68.2 & 63.5 & 61.4 & 44.9 & \underline{83.7} & 85.9 \\
MMTok & 60.8 & 43.5 & 59.7 & 55.4 & 68.7 & 70.0 & 62.7 & 41.0 & 60.9 & 57.5 & 77.7 & 72.7 & \underline{74.0} & \underline{65.3} & 73.4 & \textbf{45.1} & \textbf{84.6} & \underline{91.4} \\
SGL & \underline{80.0} & \underline{74.9} & \underline{93.0} & \underline{59.7} & \underline{73.5} & \underline{90.3} & \underline{80.8} & \underline{60.0} & \underline{93.2} & \underline{59.9} & \underline{78.5} & \underline{87.8} & 73.5 & 52.0 & \underline{85.5} & 44.8 & 80.1 & 91.1 \\
\rowcolor{oursbg} \textbf{Ours} & \textbf{81.3} & \textbf{77.7} & \textbf{94.1} & \textbf{60.9} & \textbf{78.9} & \textbf{91.4} & \textbf{82.2} & \textbf{62.0} & \textbf{94.5} & \textbf{61.0} & \textbf{82.0} & \textbf{89.0} & \textbf{75.3} & 56.0 & \textbf{89.4} & \underline{45.0} & 82.4 & \textbf{92.5} \\
\bottomrule
\end{tabular}
\end{table*}

%% file: sections/tab_ablation.tex
\begin{table}[t]
\centering\scriptsize\setlength{\tabcolsep}{3.18pt}%
\renewcommand{\arraystretch}{0.9}
\caption{\textbf{Ablations of PRUDENT (Ours).} Means over InternVL3-38B and 8B and each group's benchmarks, scored as in Table~\ref{tab:main-38b}. The first row is the ratio to the full-token score. The others change one part, in points of this ratio (Appendices~\ref{app:general} and~\ref{app:ablation}).}
\label{tab:ablation}
\begin{tabular}{lcccccc}
\toprule
 & \multicolumn{3}{c}{Text-rich} & \multicolumn{3}{c}{General} \\
\cmidrule(lr){2-4}\cmidrule(lr){5-7}
 & 2.5\% & 5\% & 10\% & 2.5\% & 5\% & 10\% \\
\midrule
\rowcolor{oursbg} \textbf{PRUDENT (Ours)} & \textbf{84.8} & \textbf{90.6} & \textbf{95.2} & \textbf{94.9} & \textbf{96.8} & \textbf{98.1} \\
\midrule
\multicolumn{7}{l}{\textit{Change one part}} \\
IF-Prune's ranking in both passes & $-$7.8 & $-$5.4 & $-$2.9 & $-$0.5 & $-$0.3 & $-$0.2 \\
No checker, one pass at $K$ & $-$2.4 & $-$1.9 & $-$1.8 & $-$1.8 & $-$1.2 & $-$1.0 \\
The checker never abstains & $-$1.1 & $-$1.2 & $-$1.0 & $-$0.4 & $-$0.3 & $-$0.2 \\
IF-Prune's ranking in the re-answers & $-$1.4 & $-$0.7 & $-$0.3 & $-$0.1 & $-$0.1 & $-$0.1 \\
\midrule
\multicolumn{7}{l}{\textit{Choose the re-answered questions by}} \\
Random choice & $-$6.4 & $-$4.7 & $-$3.8 & $-$4.9 & $-$2.7 & $-$1.8 \\
SGL's early-exit score & $-$4.4 & $-$3.3 & $-$2.8 & $-$3.4 & $-$1.8 & $-$1.3 \\
Large VLM's confidence & $-$3.0 & $-$2.4 & $-$2.1 & $-$1.7 & $-$0.9 & $-$0.6 \\
Disagreement, in random order & $-$0.8 & $-$0.6 & $-$0.4 & $-$0.2 & $-$0.1 & $-$0.1 \\
\midrule
\multicolumn{7}{l}{\textit{Change when the checker abstains, default probability 0.5 and 10 words}} \\
Below probability 0.3 & $-$0.5 & $-$0.3 & $-$0.3 & $-$0.2 & $-$0.2 & $-$0.1 \\
Below probability 0.8 & $-$0.1 & 0.0 & $-$0.5 & $-$0.5 & $-$0.3 & $-$0.4 \\
Above 5 words & +0.2 & +0.1 & +0.1 & 0.0 & 0.0 & 0.0 \\
Above 20 words & 0.0 & $-$0.1 & $-$0.1 & 0.0 & 0.0 & 0.0 \\
\midrule
\multicolumn{7}{l}{\textit{Change the checker, default InternVL3-2B}} \\
The guide, InternVL2.5-1B & 0.0 & $-$0.3 & $-$0.5 & $-$0.5 & $-$0.3 & $-$0.3 \\
InternVL3-8B, on 38B only & +0.1 & +0.5 & +0.6 & +0.3 & +0.1 & +0.2 \\
\bottomrule
\end{tabular}
\end{table}

%% file: sections/5_conclusion.tex
\section{Conclusion}
\label{sec:conclusion}

Looking at the answers that visual token pruning produces, we showed that, with strong rankings, its damage falls on a few questions.
A small VLM reading the full image flags them by disagreeing.
Detectors on the kept tokens confuse two images that share them.
PRUDENT re-answers these questions with twice the tokens at the same average budget and outperforms eight existing methods on 18 benchmarks across model sizes and families.
As rankings improve, deciding which answers to believe may matter as much as deciding which tokens to keep.

%% file: sections/6_llm_usage.tex
\section*{LLM Usage Statement}

We used Claude, GPT, Gemini, and DeepSeek to assist with literature searches, running experimental code, preparing figures and tables, and editing the manuscript. DeepSeek-V4.1-Flash scored the open-ended MM-Vet responses (Appendix~\ref{app:setup}). The authors directed the work and take full responsibility for the paper.

%% file: sections/9_appendix.tex
\appendix
\makeatletter\let\Hy@backout\@gobble\makeatother%
\renewcommand{\arraystretch}{0.9}%
\setcounter{topnumber}{3}\setcounter{bottomnumber}{2}\setcounter{totalnumber}{6}
\renewcommand{\topfraction}{0.9}\renewcommand{\bottomfraction}{0.6}\renewcommand{\textfraction}{0.07}
\renewcommand{\floatpagefraction}{0.7}
\makeatletter\setlength{\@fptop}{0pt}\setlength{\@fpbot}{0pt plus 1fil}
\let\appx@xfloat\@xfloat\def\@xfloat#1[#2]{\appx@xfloat#1[htbp]}%
\makeatother

Appendix~\ref{app:examples} shows PRUDENT on seven questions, and the rest of this supplementary material follows the order of the paper.
Appendices~\ref{app:match} to~\ref{app:theory-more} give details of the method, the proofs and further evidence for Section~\ref{sec:theory}.
Appendices~\ref{app:setup} to~\ref{app:why} give the experimental setup and further results in the order of Section~\ref{sec:exp}.
Appendix~\ref{app:limits} discusses the limitations of this line of work.

\FloatBarrier
\section{Examples}
\label{app:examples}

\begin{figure*}[t]
\centering
\includegraphics{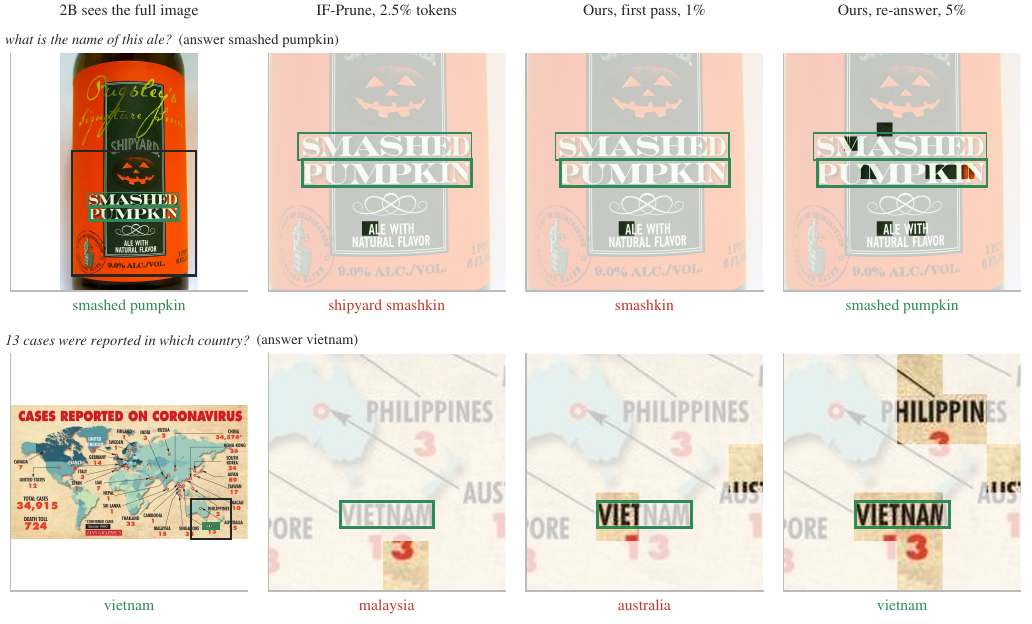}
\caption{\textbf{The two questions of Figure~\ref{fig:b1} in full.} InternVL3-38B at an average budget of 2.5\%, where PRUDENT (Ours) answers first with 1\% and re-answers with 5\%. The first column is the full image that the checker reads, with the region that the other columns enlarge. Tokens outside the kept set are veiled, the green box marks the evidence of the answer, and green and red answers are right and wrong.}
\label{fig:examples}
\end{figure*}

\begin{figure*}[t]
\centering
\includegraphics{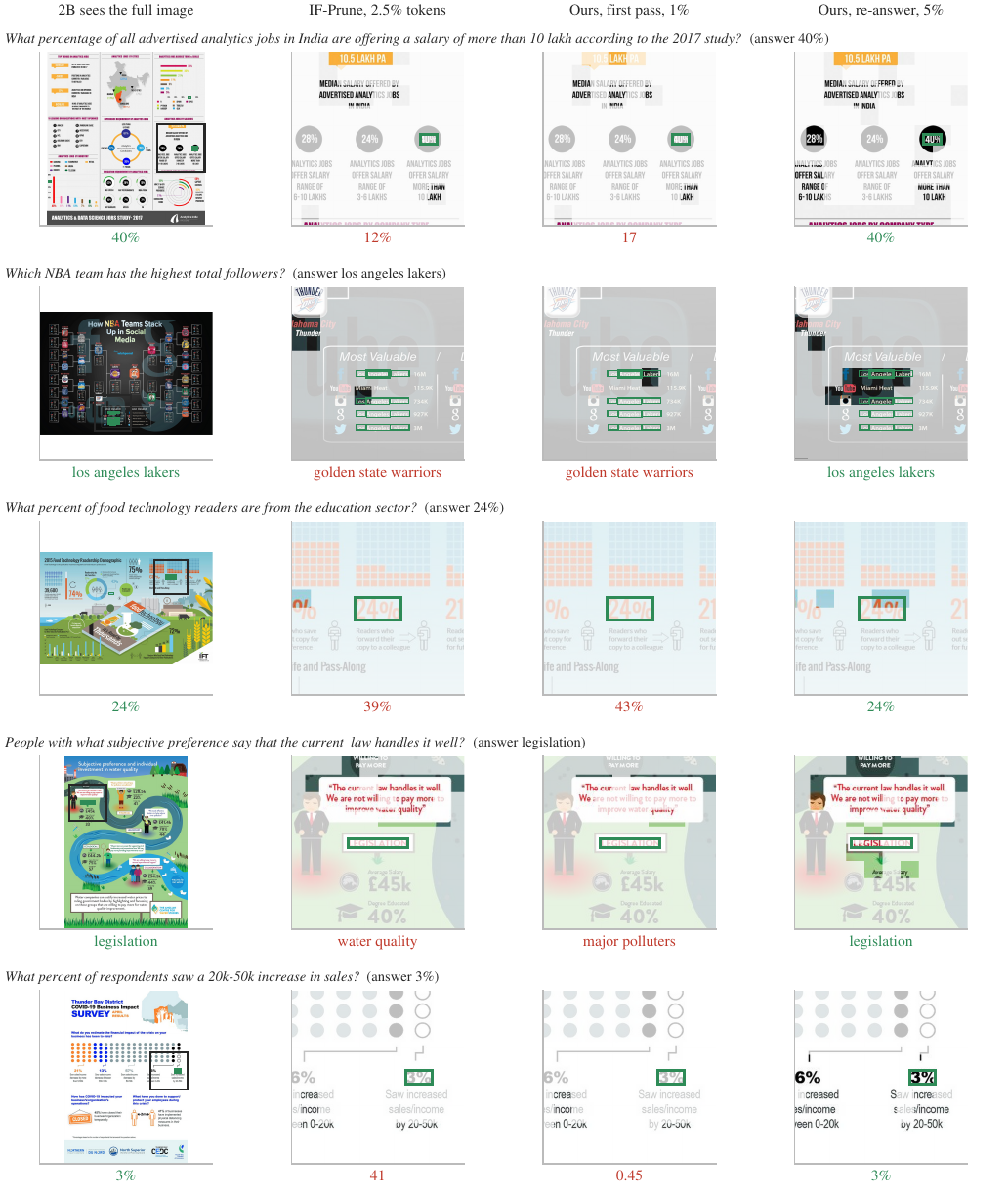}
\caption{\textbf{Five more questions on InternVL3-8B.} They are drawn as in Figure~\ref{fig:examples}. In each, IF-Prune keeps no token of the evidence and answers wrongly, the checker answers correctly from the full image, and the re-answer keeps the evidence and answers correctly. Green answers are correct and red answers wrong under the scoring of the benchmark.}
\label{fig:app-examples}
\end{figure*}

Figures~\ref{fig:examples} and~\ref{fig:app-examples} show seven questions on which the first pass loses the evidence and the re-answer recovers it, starting with the two of Figure~\ref{fig:b1}.
In each, IF-Prune and the first pass of PRUDENT keep too little of the text that answers the question, and both answer wrongly.
The checker reads this text in the full image and disagrees, and the re-answer keeps more of it and answers correctly.

\FloatBarrier

\section{Answer Matching}
\label{app:match}

Two answers count as equal when they match under the answer format of the benchmark.
For multiple-choice questions we compare option letters, so \texttt{B}, \texttt{b.}, \texttt{(b) cat} and \texttt{B: cat} all count as \texttt{B}.
When either answer does not start with an option letter, we compare the two normalized answers instead.
Yes-or-no questions compare the normalized answers.
On RefCOCO two boxes count as equal when their intersection over union is at least 0.5.
For open-ended questions we compare the core of each answer.
The core is the text inside the first pair of double quotes when the answer quotes any, and the whole answer otherwise.
Quotes matter on OCRBench, whose answers often wrap the word in a sentence template.
Normalization lowercases the text, collapses white space and drops trailing punctuation.
The comparison reads only the two answers and not the ground truth, so it is available at test time.
Proposition~\ref{prop:views} uses that it is an equivalence relation (Appendix~\ref{app:proofs}).

\FloatBarrier
\section{Algorithm and Hyperparameters}
\label{app:budget}

Algorithm~\ref{alg:prudent} gives PRUDENT on the questions of one benchmark.

\begin{algorithm}[t]
\caption{\textbf{PRUDENT on the questions of a benchmark.}}
\label{alg:prudent}
\small
\begin{algorithmic}[1]
\Require questions $Q$ with their images, budget $K$, first-pass fraction $b$ and quota $\rho$ with $b + 2\rho K = K$, abstention parameters $\tau$ and $\ell$
\Ensure the final answer $\hat{a}_q$ of every $q \in Q$
\State $C \gets \emptyset$
\For{$q \in Q$}
  \State $s_q \gets$ ranking of the visual tokens by the guide
  \State $c_q \gets$ answer of the checker from the full image
  \If{the least token probability of $c_q$ is below $\tau$, or $c_q$ has more than $\ell$ words}
    \State $\hat{a}_q \gets a_q(K)$ \Comment{the checker abstains}
  \Else
    \State $a_q(b), \pi_q(b) \gets$ the large VLM on $S_q(b)$ \Comment{first pass}
    \State $\delta_q \gets \mathds{1}[c_q \neq a_q(b)]$; \ $C \gets C \cup \{q\}$
  \EndIf
\EndFor
\State sort $C$ with $\delta_q = 1$ first, each group by $\pi_q(b)$ from low to high
\State $R \gets$ the first $\lfloor \rho |C| \rfloor$ questions of $C$
\For{$q \in C$}
  \State $\hat{a}_q \gets a_q(2K)$ if $q \in R$, else $\hat{a}_q \gets a_q(b)$ \Comment{re-answers}
\EndFor
\State \Return $\{\hat{a}_q\}_{q \in Q}$
\end{algorithmic}
\end{algorithm}

By default the first pass keeps the fraction $b = K/2$, the re-answers keep $2K$ and the quota is $\rho = 1/4$.
We use these values at the budgets of 1\%, 5\% and 10\%.
At the budget of 2.5\%, the first pass keeps 1\% and the re-answers keep 5\%, so $b + 2\rho K = K$ gives a quota of 0.3.
The budget averages a fraction per question, and the quota is rounded down to whole questions, so this average stays at most $K$.
Images differ in their number of visual tokens, so counted in tokens the large VLM of PRUDENT reads 98\% of the visual tokens of one-pass pruning on average over the 108 settings of Tables~\ref{tab:main-38b} and~\ref{tab:main-8b}, and at most 1.3\% more in any of them.
Re-answering only the disagreeing questions and leaving the rest of the quota unspent uses about 90\% of the budget in these settings and changes the mean ratio to the full-token score by less than 0.1 points.
The checker abstains when the least probable token of its answer has a probability below $\tau = 0.5$, or when its answer is longer than $\ell = 10$ words.
We use these values on every model and benchmark, and the hyperparameter study shows how the results change with them (Appendix~\ref{app:ablation}).
We chose $\tau$ and $\ell$ on InternVL3-8B and 38B with ChartQA, DocVQA, InfoVQA, TextVQA, OCRBench, WebSRC, GQA, MMBench, MME and POPE, and with MMStar, RealWorldQA and MM-Vet of Appendix~\ref{app:main}.
The other eight benchmarks of Tables~\ref{tab:main-38b} and~\ref{tab:main-8b} use them unchanged, and PRUDENT scores highest on each of them at every budget of both models.
With these values, the checker abstains on 0.3\% to 9.4\% of the questions of the multiple-choice and yes-or-no benchmarks.
On the other benchmarks it abstains on 13\% to 23\%, except OCRBench and OK-VQA with 44\% and 42\%.
On RefCOCO, the least probable token of its box coordinates falls below $\tau$ on almost every question.

\paragraph{Questions in a stream.}
Algorithm~\ref{alg:prudent} sorts the checked questions of a benchmark before it re-answers any of them.
Questions can also arrive one at a time.
PRUDENT then places each checked question in the order of Algorithm~\ref{alg:prudent} among the $m$ checked questions seen so far.
It re-answers the question when it falls among the first $\rho m$ and the re-answers so far stay below $\rho m$.
The average budget then never exceeds $K$, and no labels or held-out questions are needed.
We replay the 108 settings of Tables~\ref{tab:main-38b} and~\ref{tab:main-8b} with five random orders of arrival each.
This version keeps 98\% to 100\% of the budget and lowers the mean ratio to the full-token score by 0.1 to 0.3 points, and by at most 0.8 points on any benchmark.
It stays above IF-Prune in every setting where Algorithm~\ref{alg:prudent} is above it.

\FloatBarrier
\section{Rankings}
\label{app:ranking}

The guide is InternVL2.5-1B, and it gives two maps of the visual tokens.
The first is its own attention to the visual tokens while it reads the question and answers it, as in SGL.
The second is the importance that IF-Prune learns with a LoRA adapter on the same 1B model.
On open-ended benchmarks the ranking averages the ranks of the two maps.
Multiple-choice and yes-or-no benchmarks on InternVL are the exception.
There the guide answers with a single letter, a yes or a no, and its attention hardly points to the evidence.
On these benchmarks we use the learned importance alone.
On the other model families, the guide and the checker are a small model of the same family, Qwen3-VL-2B for Qwen3-VL and Gemma~4 E4B for Gemma~4.
The ranking there is the SGL map of that model.

\FloatBarrier
\section{Proofs}
\label{app:proofs}

\paragraph{Setting.}
A benchmark has the questions $Q$ with $|Q| = n$.
The checker answers the questions in $C \subseteq Q$, and PRUDENT re-answers the questions in $R \subseteq C$ with $|R| = \rho|C|$.
The questions outside $C$ are answered once with the fraction $K$.
Decoding is greedy, so every answer $a_q(k)$ and its score $y_q(k) \in [0, 1]$ are fixed functions of the input.
In practice the quota is rounded down to whole questions, and this only lowers the average budget.
The proofs treat $\rho|C|$ as an integer.
We write $u_q = y_q(2K) - y_q(b)$ and $h_q = y_q(K) - y_q(b)$, the one-pass curve $S_C(k) = \frac{1}{|C|}\sum_{q \in C} y_q(k)$, the net repair rate $P_R = \frac{1}{|R|}\sum_{q \in R} u_q$ and $\bar{H} = \frac{1}{|C|}\sum_{q \in C} h_q = S_C(K) - S_C(b)$.
With scores in $\{0, 1\}$, every $u_q$ lies in $\{-1, 0, 1\}$, and $p_+$ and $p_-$ are the shares of $C$ with $u_q = 1$ and $u_q = -1$.
Probabilities are shares of the questions in $C$, and answer equality is an equivalence relation (Appendix~\ref{app:match}).
Boxes of RefCOCO are the exception, since IoU $\ge 0.5$ is not transitive.
The checker abstains on almost all of them.

\paragraph{Proof of Proposition~\ref{prop:threshold}.}
The total score of PRUDENT is $\sum_{q \notin C} y_q(K) + \sum_{q \in C \setminus R} y_q(b) + \sum_{q \in R} y_q(2K)$, and one-pass pruning scores $\sum_{q \in Q} y_q(K)$.
In the difference the questions outside $C$ cancel.
A question in $C \setminus R$ contributes $y_q(b) - y_q(K) = -h_q$, and a question in $R$ contributes $y_q(2K) - y_q(K) = u_q - h_q$.
The sum is Eq.~\eqref{eq:gain},
\begin{equation*}
n\Delta = \sum_{q \in R} u_q - \sum_{q \in C} h_q = \rho|C|\,P_R - |C|\,\bar{H},
\end{equation*}
which is Eq.~\eqref{eq:threshold}, and $\Delta > 0$ exactly when $P_R > \bar{H}/\rho$.

Let $R$ be drawn uniformly from the subsets of $C$ of size $\rho|C|$.
Every question then enters $R$ with probability $\rho$, so
\begin{equation*}
\mathbb{E}[n\Delta] = |C|\big[\rho\,S_C(2K) + (1-\rho)\,S_C(b) - S_C(K)\big].
\end{equation*}
Since $b + 2\rho K = K$, we have $(1-\rho)\,b + \rho \cdot 2K = K - \rho b$.
If $S_C$ is concave on $[b, 2K]$, Jensen's inequality gives $\rho\,S_C(2K) + (1-\rho)\,S_C(b) \le S_C(K - \rho b)$.
If $S_C$ is also non-decreasing, $S_C(K - \rho b) \le S_C(K)$, and the expected gain is at most $\frac{|C|}{n}[S_C(K - \rho b) - S_C(K)] \le 0$.
Our experiments check the three-point condition $\rho\,S_C(2K) + (1-\rho)\,S_C(b) < S_C(K)$ directly (Appendix~\ref{app:grid}), and concavity is a sufficient condition for it.

For the oracle, sort the questions of $C$ by $u_q$.
With scores in $\{0, 1\}$, a share $p_+$ of them has $u_q = 1$ and a share $p_-$ has $u_q = -1$.
If $p_+ \ge \rho$, the top $\rho|C|$ questions all have $u_q = 1$ and $P_R = 1$.
Otherwise they include every question with $u_q = 1$, and the remaining places take questions with $u_q = 0$, which suffice when $\rho \le 1 - p_-$.
Then $P_R = p_+/\rho$.
No subset of size $\rho|C|$ has a larger sum of $u_q$ than the top $\rho|C|$ values, so $P_R \le \min(1, p_+/\rho)$ for every $R$.
With partial scores, the same argument with $u^+_q = \max(u_q, 0)$ bounds $P_R$ by $\min\big(1, \frac{1}{\rho|C|}\sum_{q \in C} u^+_q\big)$.
\hfill$\square$

\paragraph{Proof of Proposition~\ref{prop:views}.}
Let a detector $g$ of $\delta^*$ be a function of $V_q$.
Two questions $q$ and $q'$ with $V_q = V_{q'}$ receive the same output $g(V_q) = g(V_{q'})$.
When $\delta^*_q \neq \delta^*_{q'}$, this single output matches only one of the two labels.

Now let $g$ flag $\delta'_q = \mathds{1}[c'_q \neq a_q(b)]$ for a checker $c'$ that reads $V_q$.
Its recall satisfies
\begin{equation*}
\begin{aligned}
P(\delta'_q = 1 \mid \delta^*_q = 1) &= \frac{P(c'_q \neq a_q(b),\ \delta^*_q = 1)}{P(\delta^*_q = 1)} \\
&\le \frac{P(c'_q \neq a_q(b))}{P(\delta^*_q = 1)},
\end{aligned}
\end{equation*}
and the false alarm rate satisfies the same bound with $\delta^*_q = 0$.
Both bounds vanish as $P(c'_q \neq a_q(b))$ goes to zero.
This happens as $c'$ comes to answer like the large VLM on the kept tokens.

For a checker $c$ that reads the full image, consider the three answers $c_q$, $a_q(b)$ and $a_q(1)$ of a question.
Since answer equality is an equivalence relation, they follow exactly one of five patterns.
\begin{itemize}
\item All three are equal. Then $\delta_q(b) = 0$, $\delta^*_q = 0$ and $c_q = a_q(1)$.
\item $c_q = a_q(b) \neq a_q(1)$. Then $\delta_q(b) = 0$, $\delta^*_q = 1$ and $c_q \neq a_q(1)$. The checker misses this failure.
\item $c_q = a_q(1) \neq a_q(b)$. Then $\delta_q(b) = 1$, $\delta^*_q = 1$ and $c_q = a_q(1)$.
\item $a_q(b) = a_q(1) \neq c_q$. Then $\delta_q(b) = 1$, $\delta^*_q = 0$ and $c_q \neq a_q(1)$. The checker raises a false alarm.
\item The three are pairwise distinct. Then $\delta_q(b) = 1$, $\delta^*_q = 1$ and $c_q \neq a_q(1)$.
\end{itemize}
Transitivity rules out the remaining combinations, such as $c_q = a_q(b)$ and $a_q(b) = a_q(1)$ with $c_q \neq a_q(1)$.
The flag $\delta_q(b)$ differs from $\delta^*_q$ exactly in the second and the fourth pattern.
The event $c_q \neq a_q(1)$ holds exactly in the second, the fourth and the fifth, and the fifth is the event $E_q$.
Hence $P(\delta_q(b) \neq \delta^*_q) = \gamma - P(E_q)$, which is Eq.~\eqref{eq:views}.
\hfill$\square$

\paragraph{Remark.}
$V_q$ holds the embeddings of the kept tokens, their ranks and the question.
The large VLM answers from part of it, so $a_q(b)$ is a function of $V_q$.
The vision encoder reads every image tile as a whole before pruning.
Two images that agree on the pixels of the kept tokens can therefore differ in $V_q$, and Appendix~\ref{app:twins} measures how close such images come.

\paragraph{Proof of Corollary~\ref{cor:views}.}
The ranking $s_q$ does not depend on the budget, so $S_q(b) \subseteq S_q(b')$ for $b < b'$.
Pruning acts after the vision encoder, so a kept token has the same embedding at every budget.
Dropping from $V_q(b')$ the tokens ranked beyond the fraction $b$ gives $V_q(b)$, which is thus a function of $V_q(b')$.
For any quantity $Y$, the chain $Y \to V_q(b') \to V_q(b)$ is then a Markov chain, and the data processing inequality gives $I(Y; V_q(b)) \le I(Y; V_q(b'))$.
The share $\gamma = P(c_q \neq a_q(1))$ involves neither $b$ nor the kept tokens, since $c_q$ and $a_q(1)$ both read the full image, and the checked questions $C$ do not depend on $b$.
The proof of Proposition~\ref{prop:views} holds at every $b$, so Eq.~\eqref{eq:views} holds with the same $\gamma$.
\hfill$\square$

\paragraph{Proof of Proposition~\ref{prop:order}.}
Group the questions of $C$ by the value of $I$, and let $\mu_i$ be the mean of $u_q$ in group $i$.
A rule that sees only $I$ cannot tell the questions of a group apart, so we treat them as exchangeable.
If the rule re-answers $m_i$ questions of group $i$, the expected sum of $u_q$ over $R$ is $\sum_i m_i \mu_i$ with $\sum_i m_i = \rho|C|$.
Give every question the mean $\mu_i$ of its group. The sum is then at most the sum of the $\rho|C|$ largest of these values over $C$, the ceiling of $I$.
Now take $I = (\delta, \pi)$, so that the mean of a question is $r(\delta_q, \pi_q)$.
The two conditions state that $r(1, p) \ge r(0, p')$ for every $p$ and $p'$ and that $r(d, p)$ does not increase in $p$.
The order that takes the disagreeing questions first and sorts each group from low to high confidence therefore sorts $C$ by $r(\delta_q, \pi_q)$ from high to low.
Its first $\rho|C|$ questions attain the ceiling of $(\delta, \pi)$.
In our experiments the group means are estimated on the same questions with the confidence binned into deciles.
This makes the ceiling optimistic, as an upper bound should be.
\hfill$\square$

\paragraph{Proof of Corollary~\ref{cor:budget}.}
With $b = K/2$ and $\rho = 1/4$, the threshold is $\bar{H}/\rho = 4[S_C(K) - S_C(K/2)]$, and Proposition~\ref{prop:threshold} bounds $P_R$ by $\min(1, 4p_+)$.
Eq.~\eqref{eq:threshold} then bounds the gain of every $R$ by $\rho|C|/n$ times $\min(1, 4p_+) - \bar{H}/\rho$.
When the budget grows, the one-pass curve flattens and fewer questions remain repairable, so both the threshold and $P_R$ shrink and the gain vanishes.
For the split, let the first pass keep $b$ and the re-answers keep $b_2$, so that $\rho = (K - b)/b_2$ and
\begin{equation*}
\frac{n\Delta}{|C|} = \frac{K - b}{b_2}\,P_R - \big[S_C(K) - S_C(b)\big].
\end{equation*}
If $P_R$ changes little with $b$, the derivative in $b$ is $S_C'(b) - P_R/b_2$, which vanishes when the slope of the one-pass curve at $b$ equals $P_R/b_2$.
A concave $S_C$ makes the gain concave in $b$, so this $b$ maximizes it.
Which share of $K$ it takes depends on the shape of the one-pass curve.
At 1\% and 1.5\%, the tightest budgets of the grid, the best splits keep a larger share (Appendix~\ref{app:grid}).
\hfill$\square$
\FloatBarrier
\section{Twins Made by Editing the Pruned Region}
\label{app:twins}

Proposition~\ref{prop:views} concerns two questions with the same first-pass input whose truth differs in the pruned region.
We approximate such pairs from real pruning failures by editing only the pixels of the pruned evidence, so that both images keep the same token positions.
We then measure how much the edit changes the pruned answer.

\paragraph{Selection.}
We fixed the selection before any run.
Candidates are checked questions at the 2.5\% budget on the six text-rich benchmarks whose answers we can locate in the image.
At this budget the first pass keeps 1\%.
Each candidate is a pruning failure on which the checker disagrees, and every evidence token of it ranks beyond the first pass.
We keep a candidate under four conditions.
Its first answer has at most two words or is a number.
Its evidence spans at most three lines and 2\% of the image, and no kept token touches it.
The first answer fits the evidence box at a font size of at least 8 pixels.
Of the 1{,}520 candidates of the two models, these rules drop 99, 219, 287 and 95 questions in this order and leave 820 pairs (Table~\ref{tab:twins}).

\input{sections/tab_twins}

\paragraph{Editing.}
The edit changes only the pixels inside the evidence boxes.
It fills each box with the median color of its border and writes the wrong first answer of the large VLM in the color farthest from that fill.
The vision encoder of InternVL reads each image tile and a thumbnail of the whole image separately.
When no kept token lies in a tile that the edit changes, the first-pass input of the two images is bit-identical.
This holds for few pairs.
In all others, the edit can reach the kept tokens through the attention of the vision encoder.

\paragraph{Runs.}
Each pair yields eight answers.
The large VLM answers both images with the first pass and with full tokens, and the checker answers them from the full image and from the kept tokens alone.
All runs share one inference configuration, and the ranking of the first pass always comes from the original image, so both images keep the same token positions.

\begin{figure*}[t]
\centering
\includegraphics{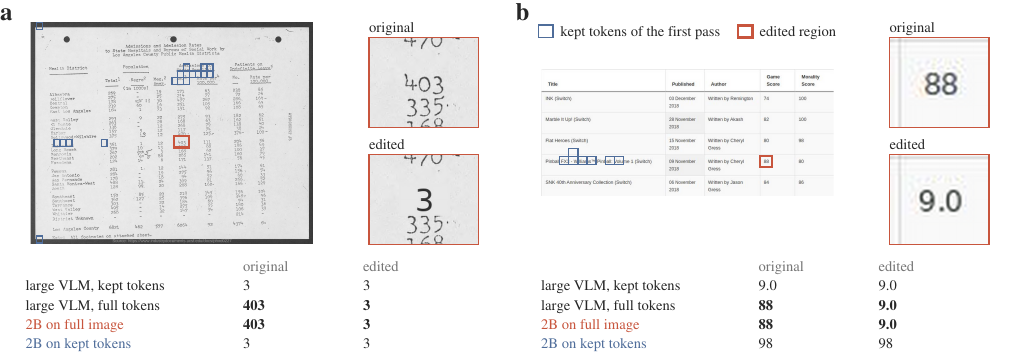}
\caption{\textbf{Two twins of InternVL3-38B at the 2.5\% budget.} Panel (a) comes from DocVQA and (b) from WebSRC. The edit writes the wrong first answer into the region of the evidence. The large VLM keeps the same token positions in both images, and its answer from them stays the same. Its full-token answer and the answer of the checker on the full image follow the edit, while the checker on the kept tokens cannot tell the images apart. Bold answers differ between the two images.}
\label{fig:twins}
\end{figure*}

\paragraph{Results.}
The pruned answer stays the same on about 9 in 10 pairs, although the edit changes the full-token answer on most pairs (Table~\ref{tab:twins}).
Its confidence barely moves.
The pairs with a bit-identical first-pass input keep the pruned answer in every case, which checks the pipeline.
Valid twins are the 392 pairs on which pruning changed the answer for the original image and not for the edited one (Figure~\ref{fig:theory}c).
On these twins, the checker that reads the full image separates the two images in most cases.
A checker that reads only the kept tokens almost never does.
It can separate a valid twin only if its answer changes between the two images.
On all pairs, its answer changes on only 12\% and 7\% for InternVL3-38B and 8B.
Figure~\ref{fig:twins} shows an example.

\FloatBarrier
\section{The Budget Grid}
\label{app:grid}

The budget grid compares PRUDENT with one-pass pruning at budgets from 1\% to 40\% on InternVL3-38B and InternVL3-8B.
It covers the six text-rich benchmarks together with GQA and VQAv2.
The budgets of 30\% and 40\% cover ChartQA, DocVQA, TextVQA and GQA, so the tables below stop at 20\%.
Every budget uses $b = K/2$ and a re-answer budget of $2K$, except that the 2.5\% budget keeps the values of Appendix~\ref{app:budget}.
The curves over the budget in Figure~\ref{fig:theory}a and~d leave out the 2.5\% budget for this reason, and the tables below include it.
All orders share the abstention and the quota of PRUDENT and differ only in which questions they re-answer first.

\paragraph{Threshold and net repair rate.}
Table~\ref{tab:app-law} gives the threshold of Proposition~\ref{prop:threshold} and the net repair rates on the grid.
On the text-rich benchmarks the threshold rises steadily as the budget tightens, since the one-pass curve grows steeper.
The net repair rate of PRUDENT rises more slowly and stays close to that of $\delta^*$.
The gain is therefore largest between 2.5\% and 4\% on these benchmarks.
At 1\%, the threshold rises to the net repair rate even of the order that puts $\delta^*$ first, so re-answering about breaks even there.
On GQA and VQAv2 the one-pass curve stays flat down to the tightest budgets, the threshold stays low, and the gain grows as the budget tightens.

\input{sections/tab_app_law}

\paragraph{Random selection.}
When $R$ is drawn at random, the expected gain equals $\frac{|C|}{n}\big[\rho\,S_C(2K) + (1-\rho)\,S_C(b) - S_C(K)\big]$ (Appendix~\ref{app:proofs}).
On the settings where we also draw $R$ at random twenty times, the mean of the draws stays within 0.15 points of this expression.
The three-point condition of Appendix~\ref{app:proofs} holds in almost every setting of the grid, so random selection loses there (Table~\ref{tab:app-ceil}).

\input{sections/tab_app_ceil}

\paragraph{Ceilings.}
Table~\ref{tab:app-ceil} adds the ceilings of three kinds of information (Proposition~\ref{prop:order}).
PRUDENT stays within about half a point of the ceiling of its own information at every budget.
Replacing the disagreement by $\delta^*$ raises the ceiling by much less than the distance that remains to the oracle.
The ceiling of the pruned view lies well below the gain of PRUDENT at tight budgets and approaches it as the budget loosens (Corollary~\ref{cor:views}).
Figure~\ref{fig:theory}f averages all 6 text-rich and 10 general benchmarks at 2.5\%.
There the ceiling of the pruned view is 0.8 points against a gain of 2.1 on the text-rich benchmarks, and 0.9 against 1.4 on the general ones.

\paragraph{Split of the budget.}
Table~\ref{tab:app-split} varies the first-pass fraction $b$ and the re-answer budget $b_2$ at the same average budget.
For each benchmark we pick in hindsight the split with the highest score on the grid.
At the tightest budgets, 1\% and 1.5\%, the best splits keep a larger share of the budget in the first pass, as Corollary~\ref{cor:budget} suggests, and they re-answer more questions with a smaller budget each.

\input{sections/tab_app_split}

\paragraph{Another model family.}
Table~\ref{tab:app-qwen} repeats the threshold and the net repair rates on Qwen3-VL-8B, with Qwen3-VL-2B as the guide and the checker (Appendix~\ref{app:ranking}).
Its one-pass curve rises steeply between 1.5\% and 5\%, for example from 37 to 73 points on the checked questions of ChartQA.
The threshold therefore peaks in this range, at budgets where the threshold of InternVL3 is already low.
On InternVL3, a cut at 0.43 on the threshold best separates the settings where re-answering gains from the others.
The same cut predicts this outcome on 90\% of the 84 settings of Qwen3-VL-8B, 7 benchmarks at 12 budgets.
Always predicting a gain is right on 68\% of them.
Ranked by the threshold alone, the settings of Qwen3-VL-8B where re-answering gains come first with an AUROC of 0.94.

\input{sections/tab_app_qwen}
\FloatBarrier
\section{Further Evidence for Section~\ref{sec:theory}}
\label{app:theory-more}

\paragraph{The ladder on every benchmark.}
Tables~\ref{tab:app-ladder-38b} and~\ref{tab:app-ladder-8b} give the ladder of Figure~\ref{fig:theory}f on each of the 16 benchmarks at three budgets.
On most benchmarks and budgets, PRUDENT gains more than the orders by confidence and by the pruned view, and the oracle stays far above every order.

\input{sections/tab_app_ladder_38b}

\input{sections/tab_app_ladder_8b}

\input{sections/tab_app_duality}

\paragraph{Checkers in both views.}
Table~\ref{tab:app-duality} lists every checker in the full and the pruned view.
The InternVL3 checkers run through the same pipeline in both views, so each pair differs only in what the checker reads.
In the full view, a checker that disagrees less often with the full-token answer raises fewer false alarms, and its recall stays high.
In the pruned view, recall falls in every setting as the checker answers more like the large VLM on the kept tokens.
Checkers of another family, Qwen3-VL, err about as often in the full view as the InternVL3 checkers with a similar $\gamma$.
Eq.~\eqref{eq:views} holds exactly in every setting, which checks the computation.
Figure~\ref{fig:theory}b leaves out InternVL3-8B as the checker of InternVL3-8B.
That checker is the large VLM itself, which gives its own full-token answer on the full image and reads its own first-pass input on the kept tokens.

\paragraph{The full view at tighter budgets.}
Table~\ref{tab:app-views} follows InternVL3-2B in both views across the budget grid, on the questions that PRUDENT checks, and Table~\ref{tab:app-views-qwen} follows Qwen3-VL-2B on Qwen3-VL-8B.
As the budget tightens, the disagreement of the full view differs from $\delta^*$ less often, while that of the pruned view does not improve.
Among the pruning failures with located evidence on the text-rich benchmarks, the share whose evidence lies wholly in the pruned region grows from about one in ten at 20\% to about half at 1\%.
The gain of the full view over the pruned view grows with it, from a few tenths of a point or less at 20\% to more than a point at 2.5\% (Table~\ref{tab:app-ceil}).
On Qwen3-VL-8B, Qwen3-VL-2B gains more from the full image than from the kept tokens by one to four points at 2.5\% and by less than a point at 10\%.

\input{sections/tab_app_views}

\input{sections/tab_app_views_qwen}

\begin{figure*}[t]
\centering
\includegraphics{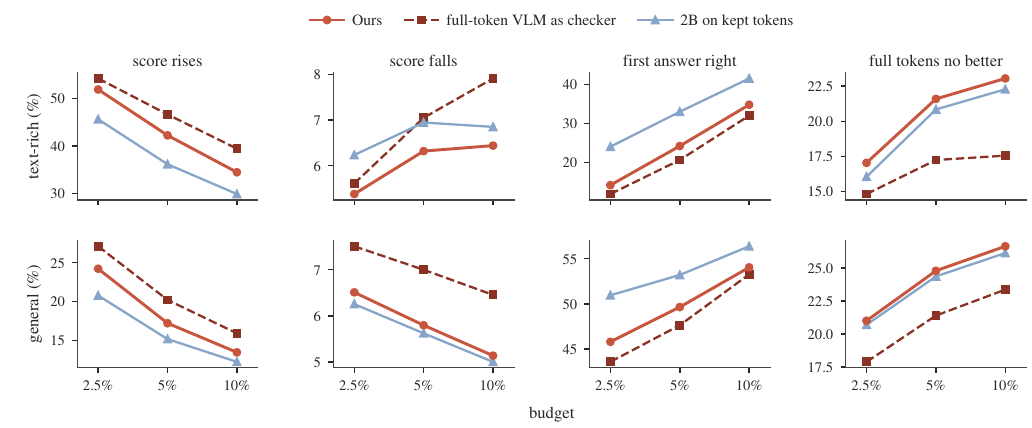}
\caption{\textbf{Outcomes of the re-answers.} Shares of the re-answered questions, means over InternVL3-38B and 8B. The score rises or falls from the first pass to the re-answer, the first answer was already right, or full tokens would not answer better than the first pass. The full-token VLM as checker orders by $\delta^*$. The three orders re-answer the same number of questions.}
\label{fig:app-quota}
\end{figure*}

\paragraph{Outcomes of the re-answers.}
Figure~\ref{fig:app-quota} sorts the re-answered questions by what the re-answer does.
At 2.5\% on the text-rich benchmarks, about half of them gain from the re-answer and few lose.
Ordering by $\delta^*$ buys almost the same mix, so the distance to the oracle lies mainly in which failures more tokens can repair.
Some of the others were already right in the first pass, and some stay wrong even with full tokens.
On GQA and VQAv2 several answers can score as correct, so the checker can disagree with a first answer that also scores as correct.

\paragraph{The conditions of Proposition~\ref{prop:order}.}
Table~\ref{tab:app-mono} checks the two conditions on the grid.
Within a confidence decile, re-answering a disagreeing question gains more than re-answering an agreeing one in almost every case.
The first condition of Proposition~\ref{prop:order} compares every decile of disagreeing questions with every decile of agreeing ones.
It holds on each group with the settings pooled (Figure~\ref{fig:theory}e), and in 40 of the 48 settings on the deciles with at least 20 questions.
Re-answering an agreeing question gains nothing on average, and among the disagreeing questions the gain falls as the confidence rises in most settings.

\begin{table}[htbp]\centering
\begingroup
\renewenvironment{table}[1][]{\begin{minipage}[t]{0.485\textwidth}\vspace{0pt}\centering}{\end{minipage}}%
\renewenvironment{table*}[1][]{\begin{minipage}[t]{0.485\textwidth}\vspace{0pt}\centering}{\end{minipage}}%
\input{sections/tab_app_mono}\hfill\input{sections/tab_app_pvgain}\endgroup
\end{table}

\paragraph{Pruned-view signals combined with labels.}
Table~\ref{tab:app-pvgain} trains a classifier with labels on the pruned-view signals in its caption, including three pruned-view checkers.
At 2.5\% on the text-rich benchmarks it reaches about half the gain of PRUDENT, whose order needs no labels.
PRUDENT gains more in most settings.

%% file: sections/tab_twins.tex
\begin{table*}[t]
\centering
\caption{\textbf{Twins made by editing the pruned region.} The two images of a pair keep the same token positions and differ in the truth of the pruned region. Intervals are Wilson 95\% intervals.}
\label{tab:twins}
\small
\begin{tabular}{@{}lrr@{}}
\toprule
 & InternVL3-38B & InternVL3-8B \\
\midrule
Candidate questions & 771 & 749 \\
Pairs built & 411 & 409 \\
\quad with a bit-identical first-pass input & 14 & 16 \\
Pruned answer unchanged (\%) & 90.0 (86.7 to 92.6) & 91.0 (87.8 to 93.4) \\
Full-token answer follows the edit (\%) & 74.9 (70.5 to 78.9) & 69.7 (65.1 to 73.9) \\
Valid twins & 198 & 194 \\
\quad split by the checker on the full image (\%) & 63.6 & 63.4 \\
\quad split by the checker on the kept tokens (\%) & 3.5 & 0.5 \\
Change in first-pass confidence (median) & 0.017 & 0.017 \\
\bottomrule
\end{tabular}
\end{table*}

%% file: sections/tab_app_law.tex
\begin{table*}[t]
\centering\footnotesize\setlength{\tabcolsep}{6.58pt}
\caption{\textbf{Threshold and net repair rate on the budget grid.} Thr.\ is the threshold $\bar{H}/\rho$, and Ours and $\delta^*$ are the net repair rates $P_R$ of PRUDENT and of the order that puts $\delta^*$ first. Means over the six text-rich benchmarks, and over GQA and VQAv2.}
\label{tab:app-law}
\begin{tabular}{@{}lcccccccccccc@{}}
\toprule
 & \multicolumn{3}{c}{Text-rich, InternVL3-38B} & \multicolumn{3}{c}{Text-rich, InternVL3-8B} & \multicolumn{3}{c}{General, InternVL3-38B} & \multicolumn{3}{c}{General, InternVL3-8B} \\
\cmidrule(lr){2-4}\cmidrule(lr){5-7}\cmidrule(lr){8-10}\cmidrule(lr){11-13}
Budget & Thr. & Ours & $\delta^*$ & Thr. & Ours & $\delta^*$ & Thr. & Ours & $\delta^*$ & Thr. & Ours & $\delta^*$ \\
\midrule
1\% & 0.45 & 0.43 & 0.44 & 0.42 & 0.43 & 0.44 & 0.17 & 0.29 & 0.27 & 0.15 & 0.25 & 0.28 \\
1.5\% & 0.37 & 0.40 & 0.42 & 0.33 & 0.38 & 0.38 & 0.13 & 0.25 & 0.22 & 0.11 & 0.22 & 0.23 \\
2\% & 0.33 & 0.38 & 0.41 & 0.28 & 0.35 & 0.37 & 0.13 & 0.22 & 0.20 & 0.10 & 0.18 & 0.19 \\
2.5\% & 0.32 & 0.40 & 0.41 & 0.28 & 0.38 & 0.39 & 0.13 & 0.21 & 0.20 & 0.10 & 0.16 & 0.16 \\
3\% & 0.25 & 0.35 & 0.37 & 0.23 & 0.34 & 0.35 & 0.09 & 0.19 & 0.17 & 0.08 & 0.15 & 0.15 \\
4\% & 0.21 & 0.32 & 0.35 & 0.21 & 0.31 & 0.32 & 0.07 & 0.16 & 0.14 & 0.07 & 0.11 & 0.12 \\
5\% & 0.21 & 0.30 & 0.34 & 0.22 & 0.28 & 0.31 & 0.06 & 0.15 & 0.13 & 0.06 & 0.10 & 0.10 \\
6\% & 0.19 & 0.28 & 0.30 & 0.19 & 0.27 & 0.29 & 0.07 & 0.15 & 0.14 & 0.05 & 0.08 & 0.08 \\
8\% & 0.18 & 0.26 & 0.30 & 0.16 & 0.23 & 0.26 & 0.06 & 0.14 & 0.14 & 0.04 & 0.06 & 0.06 \\
10\% & 0.17 & 0.24 & 0.28 & 0.13 & 0.22 & 0.23 & 0.07 & 0.12 & 0.13 & 0.04 & 0.05 & 0.05 \\
15\% & 0.13 & 0.19 & 0.21 & 0.12 & 0.18 & 0.19 & 0.06 & 0.10 & 0.10 & 0.01 & 0.04 & 0.03 \\
20\% & 0.10 & 0.15 & 0.17 & 0.10 & 0.15 & 0.16 & 0.04 & 0.08 & 0.08 & 0.01 & 0.02 & 0.02 \\
\bottomrule
\end{tabular}
\end{table*}

%% file: sections/tab_app_ceil.tex
\begin{table*}[t]
\centering\footnotesize\setlength{\tabcolsep}{6.00pt}
\caption{\textbf{Gains over one-pass pruning on the budget grid.} Values are in points of the benchmark score. Each order re-answers the same number of questions. The ceiling of an order is the ceiling of its information, its flag together with the confidence of the large VLM in deciles (Proposition~\ref{prop:order}). Orders sort by the exact confidence and can lie slightly above their ceiling.}
\label{tab:app-ceil}
\begin{tabular}{@{}llccccccccc@{}}
\toprule
 & & & & \multicolumn{2}{c}{Pruned view} & \multicolumn{2}{c}{Full view (Ours)} & \multicolumn{2}{c}{$\delta^*$} & \\
\cmidrule(lr){5-6}\cmidrule(lr){7-8}\cmidrule(lr){9-10}
 & Budget & Random & Confidence & Order & Ceiling & Order & Ceiling & Order & Ceiling & Oracle \\
\midrule
\multirow{5}{*}{\shortstack[l]{Text-rich\\InternVL3-38B}} & 1\% & $-$4.8 & $-$1.8 & $-$0.8 & $-$0.6 & $-$0.3 & 0.2 & 0.0 & 0.5 & 7.1 \\
 & 2.5\% & $-$3.9 & $-$0.6 & 0.5 & 0.7 & 1.8 & 1.9 & 2.3 & 2.5 & 6.5 \\
 & 5\% & $-$2.0 & 0.0 & 1.0 & 1.1 & 2.0 & 2.2 & 2.7 & 3.0 & 5.5 \\
 & 10\% & $-$2.0 & $-$0.2 & 0.8 & 0.8 & 1.5 & 1.6 & 2.2 & 2.3 & 3.8 \\
 & 20\% & $-$1.2 & 0.2 & 0.7 & 0.8 & 1.1 & 1.2 & 1.5 & 1.6 & 2.6 \\
\midrule
\multirow{5}{*}{\shortstack[l]{Text-rich\\InternVL3-8B}} & 1\% & $-$4.8 & $-$1.4 & $-$0.9 & $-$0.6 & 0.1 & 0.4 & 0.3 & 0.6 & 6.8 \\
 & 2.5\% & $-$3.2 & $-$0.4 & 0.8 & 0.9 & 2.4 & 2.5 & 2.7 & 2.8 & 7.1 \\
 & 5\% & $-$2.4 & $-$0.7 & 0.2 & 0.3 & 1.4 & 1.5 & 1.9 & 2.2 & 5.1 \\
 & 10\% & $-$1.4 & $-$0.1 & 0.8 & 0.9 & 1.7 & 1.9 & 2.0 & 2.3 & 4.1 \\
 & 20\% & $-$1.2 & $-$0.2 & 0.5 & 0.6 & 1.1 & 1.3 & 1.3 & 1.5 & 3.0 \\
\midrule
\multirow{5}{*}{\shortstack[l]{General\\InternVL3-38B}} & 1\% & $-$1.9 & $-$1.2 & 0.5 & 0.7 & 2.4 & 2.6 & 1.9 & 3.4 & 6.2 \\
 & 2.5\% & $-$1.8 & $-$1.0 & 0.8 & 1.0 & 2.0 & 2.0 & 1.6 & 2.4 & 4.4 \\
 & 5\% & $-$0.6 & $-$0.5 & 1.3 & 1.4 & 1.7 & 1.7 & 1.3 & 2.0 & 3.6 \\
 & 10\% & $-$0.9 & $-$0.7 & 0.8 & 0.9 & 1.0 & 1.0 & 1.2 & 1.3 & 2.7 \\
 & 20\% & $-$0.5 & $-$0.3 & 0.7 & 0.7 & 0.7 & 0.7 & 0.7 & 0.8 & 2.2 \\
\midrule
\multirow{5}{*}{\shortstack[l]{General\\InternVL3-8B}} & 1\% & $-$1.7 & $-$0.7 & 0.3 & 0.2 & 2.0 & 2.0 & 2.6 & 2.6 & 5.0 \\
 & 2.5\% & $-$1.3 & $-$0.3 & 0.4 & 0.5 & 1.5 & 1.5 & 1.4 & 1.6 & 3.9 \\
 & 5\% & $-$0.7 & 0.0 & 0.5 & 0.5 & 0.8 & 0.9 & 0.9 & 0.9 & 2.6 \\
 & 10\% & $-$0.5 & $-$0.2 & 0.1 & 0.3 & 0.3 & 0.4 & 0.2 & 0.4 & 1.8 \\
 & 20\% & $-$0.1 & 0.1 & 0.3 & 0.4 & 0.4 & 0.6 & 0.3 & 0.4 & 1.8 \\
\bottomrule
\end{tabular}
\end{table*}

%% file: sections/tab_app_split.tex
\begin{table*}[t]
\centering\footnotesize\setlength{\tabcolsep}{5.37pt}
\caption{\textbf{The best split of the budget in hindsight.} For every benchmark we pick, among the splits on the grid, the first-pass fraction $b$ and the re-answer budget $b_2$ with the highest score at the same average budget. We report their medians as multiples of $K$ and the mean gain of this pick over the default split, in points.}
\label{tab:app-split}
\begin{tabular}{@{}lccccccccccccc@{}}
\toprule
 & & \multicolumn{3}{c}{Text-rich, InternVL3-38B} & \multicolumn{3}{c}{Text-rich, InternVL3-8B} & \multicolumn{3}{c}{General, InternVL3-38B} & \multicolumn{3}{c}{General, InternVL3-8B} \\
\cmidrule(lr){3-5}\cmidrule(lr){6-8}\cmidrule(lr){9-11}\cmidrule(lr){12-14}
Budget & Splits & $b/K$ & $b_2/K$ & Gap & $b/K$ & $b_2/K$ & Gap & $b/K$ & $b_2/K$ & Gap & $b/K$ & $b_2/K$ & Gap \\
\midrule
1\% & 34 & 0.50 & 1.5 & 1.3 & 0.50 & 1.5 & 1.1 & 0.50 & 1.5 & 0.3 & 0.50 & 1.8 & 0.1 \\
1.5\% & 48 & 0.50 & 1.3 & 1.5 & 0.42 & 1.3 & 1.2 & 0.42 & 1.8 & 0.6 & 0.42 & 2.0 & 0.1 \\
2\% & 60 & 0.38 & 1.5 & 1.5 & 0.38 & 1.8 & 1.3 & 0.31 & 2.0 & 0.5 & 0.25 & 2.2 & 0.2 \\
2.5\% & 70 & 0.30 & 1.6 & 1.3 & 0.30 & 1.8 & 0.9 & 0.25 & 2.4 & 0.3 & 0.30 & 2.4 & 0.2 \\
3\% & 78 & 0.29 & 1.7 & 0.7 & 0.25 & 1.7 & 1.0 & 0.25 & 2.5 & 0.5 & 0.25 & 2.6 & 0.2 \\
4\% & 84 & 0.38 & 1.9 & 0.9 & 0.31 & 1.7 & 1.2 & 0.28 & 2.5 & 0.5 & 0.38 & 2.5 & 0.3 \\
5\% & 88 & 0.30 & 1.8 & 1.3 & 0.30 & 1.8 & 1.4 & 0.35 & 2.4 & 0.3 & 0.25 & 3.1 & 0.2 \\
6\% & 90 & 0.29 & 2.0 & 1.2 & 0.21 & 2.0 & 1.0 & 0.29 & 2.6 & 0.5 & 0.25 & 2.9 & 0.2 \\
8\% & 88 & 0.25 & 2.0 & 0.8 & 0.31 & 1.9 & 0.9 & 0.31 & 2.5 & 0.1 & 0.50 & 2.8 & 0.4 \\
10\% & 84 & 0.30 & 2.0 & 0.6 & 0.22 & 2.0 & 0.4 & 0.12 & 3.0 & 0.3 & 0.40 & 3.5 & 0.4 \\
\bottomrule
\end{tabular}
\end{table*}

%% file: sections/tab_app_qwen.tex
\begin{table}[t]
\centering\footnotesize\setlength{\tabcolsep}{5.28pt}
\caption{\textbf{Threshold and net repair rate on Qwen3-VL-8B.} The checker is Qwen3-VL-2B. Columns as in Table~\ref{tab:app-law}, and Bound is $\min(1, p_+/\rho)$ of Proposition~\ref{prop:threshold}. Means over ChartQA, DocVQA, InfoVQA, TextVQA and OCRBench, and over GQA and VQAv2.}
\label{tab:app-qwen}
\begin{tabular}{@{}lcccccccc@{}}
\toprule
 & \multicolumn{4}{c}{Text-rich} & \multicolumn{4}{c}{General} \\
\cmidrule(lr){2-5}\cmidrule(lr){6-9}
Budget & Thr. & Ours & $\delta^*$ & Bound & Thr. & Ours & $\delta^*$ & Bound \\
\midrule
1\% & 0.51 & 0.42 & 0.42 & 0.92 & 0.08 & 0.24 & 0.25 & 0.62 \\
1.5\% & 0.57 & 0.44 & 0.44 & 0.92 & 0.10 & 0.36 & 0.38 & 0.87 \\
2\% & 0.52 & 0.42 & 0.43 & 0.86 & 0.23 & 0.45 & 0.46 & 0.95 \\
2.5\% & 0.54 & 0.45 & 0.46 & 0.81 & 0.34 & 0.49 & 0.49 & 0.92 \\
3\% & 0.42 & 0.38 & 0.40 & 0.79 & 0.50 & 0.50 & 0.51 & 1.00 \\
4\% & 0.35 & 0.38 & 0.39 & 0.72 & 0.55 & 0.49 & 0.50 & 0.96 \\
5\% & 0.31 & 0.37 & 0.38 & 0.71 & 0.51 & 0.45 & 0.48 & 0.90 \\
6\% & 0.28 & 0.35 & 0.36 & 0.68 & 0.41 & 0.40 & 0.44 & 0.84 \\
8\% & 0.25 & 0.34 & 0.35 & 0.59 & 0.28 & 0.31 & 0.37 & 0.59 \\
10\% & 0.25 & 0.32 & 0.34 & 0.54 & 0.19 & 0.25 & 0.27 & 0.44 \\
15\% & 0.26 & 0.30 & 0.32 & 0.46 & 0.11 & 0.14 & 0.14 & 0.29 \\
20\% & 0.25 & 0.28 & 0.30 & 0.43 & 0.06 & 0.09 & 0.09 & 0.21 \\
\bottomrule
\end{tabular}
\end{table}

%% file: sections/tab_app_ladder_38b.tex
\begin{table*}[t]
\centering\footnotesize\setlength{\tabcolsep}{5.24pt}
\caption{\textbf{The ladder per benchmark on InternVL3-38B.} Gains are in points of the benchmark score over one-pass pruning. Every order re-answers the same number of questions with the same abstention. Conf.\ orders by the confidence of the large VLM, Pruned by the disagreement of InternVL3-2B on the kept tokens, Ours by the disagreement of the checker on the full image, and $\delta^*$ by the full-token answer of the large VLM itself.}
\label{tab:app-ladder-38b}
\begin{tabular}{@{}lccccccccccccccc@{}}
\toprule
 & \multicolumn{5}{c}{Retain 2.5\% of the visual tokens} & \multicolumn{5}{c}{Retain 5\% of the visual tokens} & \multicolumn{5}{c}{Retain 10\% of the visual tokens} \\
\cmidrule(lr){2-6}\cmidrule(lr){7-11}\cmidrule(lr){12-16}
Benchmark & Conf. & Pruned & Ours & $\delta^*$ & Oracle & Conf. & Pruned & Ours & $\delta^*$ & Oracle & Conf. & Pruned & Ours & $\delta^*$ & Oracle \\
\midrule
ChartQA & $-$0.2 & 1.8 & 3.1 & 4.1 & 9.0 & 1.6 & 3.4 & 4.3 & 5.2 & 8.6 & 0.7 & 2.5 & 3.7 & 4.8 & 6.1 \\
DocVQA & 0.7 & 2.1 & 3.5 & 4.4 & 6.1 & $-$0.1 & 1.0 & 2.4 & 3.3 & 4.4 & $-$0.1 & 1.2 & 2.0 & 2.3 & 3.0 \\
InfoVQA & $-$1.6 & $-$1.5 & $-$0.4 & 0.3 & 7.7 & $-$0.1 & 0.3 & 1.4 & 1.9 & 7.6 & $-$0.7 & 0.0 & 0.4 & 1.0 & 5.8 \\
TextVQA & $-$0.5 & 1.7 & 2.7 & 2.8 & 4.4 & $-$0.4 & 0.8 & 1.3 & 1.4 & 2.9 & $-$0.1 & 0.4 & 0.8 & 0.6 & 2.0 \\
OCRBench & $-$2.0 & $-$1.0 & 0.2 & 0.1 & 5.9 & $-$2.1 & $-$0.7 & 0.4 & 0.8 & 4.2 & $-$1.8 & $-$0.4 & 0.1 & 1.0 & 1.6 \\
WebSRC & 0.0 & 0.1 & 1.7 & 2.1 & 6.2 & 0.9 & 1.2 & 2.5 & 3.6 & 5.3 & 0.6 & 0.9 & 2.0 & 3.6 & 4.3 \\
GQA & $-$0.7 & $-$0.1 & 0.8 & 1.3 & 4.0 & $-$0.1 & 0.7 & 0.9 & 1.1 & 3.4 & $-$0.2 & 0.2 & 0.4 & 0.6 & 2.6 \\
VQAv2 & $-$1.3 & 1.7 & 3.1 & 1.9 & 4.8 & $-$0.8 & 1.8 & 2.5 & 1.4 & 3.8 & $-$1.2 & 1.4 & 1.6 & 1.8 & 2.9 \\
OK-VQA & $-$4.2 & $-$2.0 & $-$0.8 & 1.2 & 3.6 & $-$2.0 & $-$0.8 & $-$0.4 & 1.6 & 2.5 & $-$1.1 & $-$0.3 & $-$0.1 & 0.6 & 1.7 \\
MMB & 0.9 & 0.8 & 1.1 & 1.5 & 2.9 & 0.5 & 0.6 & 0.6 & 0.6 & 1.9 & 0.5 & 0.5 & 0.5 & 0.5 & 1.6 \\
MMB-CN & 0.8 & 0.8 & 1.4 & 1.6 & 3.2 & 0.8 & 0.7 & 0.9 & 0.9 & 2.0 & 0.7 & 0.7 & 0.8 & 0.8 & 1.8 \\
MME & 1.1 & 1.3 & 1.4 & 1.6 & 3.7 & 0.5 & 0.4 & 0.5 & 0.6 & 1.9 & 0.0 & $-$0.2 & 0.0 & 0.1 & 1.2 \\
POPE & 0.5 & 0.5 & 0.4 & 0.4 & 2.9 & 0.4 & 0.4 & 0.4 & 0.4 & 1.9 & 0.1 & 0.1 & 0.1 & 0.1 & 1.0 \\
SEED & $-$0.1 & 0.2 & 0.9 & 1.1 & 4.1 & 0.3 & 0.4 & 0.6 & 0.6 & 2.8 & 0.4 & 0.5 & 0.6 & 0.6 & 2.4 \\
AI2D & 2.0 & 1.1 & 3.3 & 3.6 & 6.3 & 1.7 & 0.8 & 2.4 & 3.7 & 5.3 & 1.7 & 0.9 & 1.6 & 2.8 & 4.4 \\
SQA-I & 1.7 & 2.1 & 2.4 & 2.4 & 3.9 & 2.2 & 2.4 & 2.6 & 2.6 & 3.5 & 1.3 & 1.4 & 1.5 & 1.5 & 2.2 \\
\bottomrule
\end{tabular}
\end{table*}

%% file: sections/tab_app_ladder_8b.tex
\begin{table*}[t]
\centering\footnotesize\setlength{\tabcolsep}{5.24pt}
\caption{\textbf{The ladder per benchmark on InternVL3-8B.} Gains are in points of the benchmark score over one-pass pruning. Every order re-answers the same number of questions with the same abstention. Conf.\ orders by the confidence of the large VLM, Pruned by the disagreement of InternVL3-2B on the kept tokens, Ours by the disagreement of the checker on the full image, and $\delta^*$ by the full-token answer of the large VLM itself.}
\label{tab:app-ladder-8b}
\begin{tabular}{@{}lccccccccccccccc@{}}
\toprule
 & \multicolumn{5}{c}{Retain 2.5\% of the visual tokens} & \multicolumn{5}{c}{Retain 5\% of the visual tokens} & \multicolumn{5}{c}{Retain 10\% of the visual tokens} \\
\cmidrule(lr){2-6}\cmidrule(lr){7-11}\cmidrule(lr){12-16}
Benchmark & Conf. & Pruned & Ours & $\delta^*$ & Oracle & Conf. & Pruned & Ours & $\delta^*$ & Oracle & Conf. & Pruned & Ours & $\delta^*$ & Oracle \\
\midrule
ChartQA & 2.1 & 3.4 & 5.5 & 5.8 & 8.7 & $-$0.6 & 0.6 & 1.5 & 2.4 & 4.8 & 0.8 & 1.6 & 2.8 & 2.7 & 4.5 \\
DocVQA & 0.8 & 1.4 & 3.3 & 3.6 & 6.6 & 0.0 & 1.3 & 2.5 & 3.5 & 5.3 & 0.1 & 1.3 & 2.3 & 2.9 & 4.1 \\
InfoVQA & $-$1.0 & $-$0.1 & 1.1 & 1.3 & 10.2 & $-$1.6 & $-$1.3 & $-$0.1 & 0.2 & 8.1 & $-$0.7 & $-$0.2 & 0.7 & 1.4 & 6.5 \\
TextVQA & $-$0.2 & 1.6 & 3.0 & 3.1 & 4.8 & $-$0.1 & 1.1 & 1.7 & 1.5 & 3.2 & 0.2 & 0.9 & 1.5 & 1.1 & 2.6 \\
OCRBench & $-$2.8 & $-$1.0 & 0.0 & 0.3 & 6.0 & $-$2.1 & $-$1.0 & 0.4 & 0.4 & 3.9 & $-$1.6 & $-$0.2 & 0.3 & 0.2 & 1.9 \\
WebSRC & $-$1.2 & $-$0.6 & 1.4 & 1.9 & 6.3 & 0.2 & 0.8 & 2.4 & 3.1 & 5.5 & 0.5 & 1.4 & 2.5 & 3.8 & 4.8 \\
GQA & $-$0.1 & 0.5 & 1.5 & 1.4 & 4.8 & 0.1 & 0.4 & 0.7 & 0.8 & 2.9 & $-$0.1 & 0.0 & 0.1 & 0.1 & 2.1 \\
VQAv2 & $-$0.6 & 0.3 & 1.4 & 1.5 & 2.9 & $-$0.1 & 0.6 & 1.0 & 1.0 & 2.2 & $-$0.3 & 0.2 & 0.5 & 0.4 & 1.6 \\
OK-VQA & $-$1.7 & $-$0.4 & 0.2 & 1.1 & 2.7 & $-$1.0 & $-$0.1 & 0.2 & 0.5 & 1.9 & $-$0.6 & 0.2 & 0.2 & 0.2 & 1.4 \\
MMB & 0.9 & 0.9 & 1.0 & 1.1 & 2.5 & 0.0 & 0.3 & 0.3 & 0.3 & 1.6 & 0.8 & 1.0 & 1.0 & 1.0 & 2.1 \\
MMB-CN & 0.6 & 0.6 & 0.8 & 1.0 & 3.0 & 0.3 & 0.4 & 0.5 & 0.5 & 1.9 & 0.6 & 0.6 & 0.6 & 0.7 & 1.9 \\
MME & 1.0 & 1.1 & 1.1 & 1.2 & 3.2 & 0.5 & 0.4 & 0.5 & 0.6 & 1.8 & 0.3 & 0.3 & 0.3 & 0.3 & 1.3 \\
POPE & 0.4 & 0.4 & 0.5 & 0.4 & 1.7 & 0.0 & 0.0 & 0.0 & $-$0.1 & 1.0 & 0.0 & 0.0 & 0.0 & 0.0 & 1.0 \\
SEED & 0.4 & 0.6 & 1.3 & 1.4 & 4.5 & $-$0.1 & 0.1 & 0.5 & 0.4 & 2.6 & 0.5 & 0.5 & 0.7 & 0.6 & 2.6 \\
AI2D & 0.7 & 0.3 & 2.9 & 3.5 & 6.1 & 0.7 & 0.1 & 1.9 & 2.5 & 5.1 & 1.3 & 0.4 & 2.0 & 2.1 & 4.1 \\
SQA-I & 2.0 & 2.2 & 2.9 & 2.9 & 4.2 & 1.0 & 1.4 & 1.5 & 1.5 & 2.4 & 1.4 & 1.3 & 1.5 & 1.5 & 2.7 \\
\bottomrule
\end{tabular}
\end{table*}

%% file: sections/tab_app_duality.tex
\begin{table*}[t]
\centering\footnotesize\setlength{\tabcolsep}{6.00pt}
\caption{\textbf{Checkers in the full and the pruned view.} Values are in percent. $\gamma$ is how often the checker disagrees with the full-token answer, recall and false alarm refer to $\delta^*$, and Differs is how often its flag differs from $\delta^*$. All questions of the 16 benchmarks, with InternVL3-38B and InternVL3-8B as the large VLM. InternVL3-1B is a model of its own, not the guide InternVL2.5-1B.}
\label{tab:app-duality}
\begin{tabular}{@{}llcccccccc@{}}
\toprule
 & & \multicolumn{4}{c}{Text-rich} & \multicolumn{4}{c}{General} \\
\cmidrule(lr){3-6}\cmidrule(lr){7-10}
View & Checker & $\gamma$ & Recall & False alarm & Differs & $\gamma$ & Recall & False alarm & Differs \\
\midrule
\multicolumn{10}{@{}l}{\textit{First pass keeps 1\% of the visual tokens}} \\
Full & InternVL3-1B & 39.0 & 90.1 & 21.6 & 14.9 & 26.0 & 74.4 & 20.6 & 19.7 \\
 & InternVL3-2B & 30.5 & 91.5 & 13.6 & 10.5 & 19.7 & 78.0 & 14.4 & 14.7 \\
 & InternVL3-8B & 25.5 & 92.5 & 10.5 & 8.6 & 14.4 & 80.2 & 9.2 & 10.6 \\
 & InternVL3-14B & 23.1 & 93.2 & 9.3 & 7.6 & 13.8 & 80.2 & 8.5 & 10.3 \\
 & Qwen3-VL-2B & 35.9 & 92.2 & 18.8 & 12.1 & 26.3 & 74.7 & 19.2 & 19.1 \\
 & Qwen3-VL-4B & 32.0 & 93.1 & 16.1 & 10.5 & 22.5 & 78.2 & 15.9 & 15.8 \\
 & Qwen3-VL-8B & 31.5 & 93.2 & 15.5 & 10.1 & 22.1 & 78.0 & 15.5 & 15.6 \\
Pruned & InternVL3-1B & 56.8 & 80.2 & 31.1 & 24.7 & 32.5 & 63.1 & 25.4 & 25.5 \\
 & InternVL3-2B & 52.3 & 76.5 & 23.5 & 22.7 & 26.7 & 59.6 & 17.8 & 20.9 \\
 & InternVL3-8B & 50.4 & 75.9 & 21.7 & 22.0 & 22.5 & 56.4 & 12.6 & 17.7 \\
 & InternVL3-14B & 49.7 & 73.2 & 19.9 & 22.2 & 22.9 & 51.3 & 12.2 & 18.5 \\
\midrule
\multicolumn{10}{@{}l}{\textit{First pass keeps 2.5\% of the visual tokens}} \\
Full & InternVL3-1B & 39.0 & 86.8 & 24.1 & 19.0 & 26.0 & 71.1 & 21.4 & 21.1 \\
 & InternVL3-2B & 30.5 & 88.3 & 15.7 & 13.5 & 19.7 & 73.6 & 15.1 & 15.7 \\
 & InternVL3-8B & 25.5 & 88.5 & 12.2 & 11.4 & 14.4 & 74.9 & 9.9 & 11.4 \\
 & InternVL3-14B & 23.1 & 90.3 & 10.8 & 9.8 & 13.8 & 75.5 & 9.3 & 10.9 \\
 & Qwen3-VL-2B & 35.9 & 89.7 & 20.7 & 15.5 & 26.3 & 71.8 & 20.5 & 20.5 \\
 & Qwen3-VL-4B & 32.0 & 90.6 & 17.8 & 13.5 & 22.5 & 74.6 & 16.9 & 17.2 \\
 & Qwen3-VL-8B & 31.5 & 90.9 & 17.3 & 13.0 & 22.1 & 73.5 & 16.5 & 17.0 \\
Pruned & InternVL3-1B & 49.0 & 79.7 & 29.8 & 25.3 & 29.5 & 65.2 & 24.3 & 24.3 \\
 & InternVL3-2B & 44.2 & 77.1 & 23.2 & 22.2 & 23.7 & 61.6 & 17.3 & 19.3 \\
 & InternVL3-8B & 41.9 & 74.4 & 20.3 & 21.5 & 19.4 & 57.9 & 12.3 & 15.7 \\
 & InternVL3-14B & 39.9 & 72.2 & 16.7 & 20.0 & 19.5 & 52.8 & 11.8 & 16.0 \\
\midrule
\multicolumn{10}{@{}l}{\textit{First pass keeps 5\% of the visual tokens}} \\
Full & InternVL3-1B & 39.0 & 85.6 & 26.5 & 22.1 & 26.0 & 69.7 & 22.3 & 22.1 \\
 & InternVL3-2B & 30.5 & 85.8 & 17.8 & 16.1 & 19.7 & 71.9 & 15.9 & 16.5 \\
 & InternVL3-8B & 25.5 & 85.9 & 14.0 & 13.6 & 14.4 & 72.0 & 10.6 & 12.0 \\
 & InternVL3-14B & 23.1 & 87.4 & 12.0 & 11.7 & 13.8 & 72.9 & 10.0 & 11.4 \\
 & Qwen3-VL-2B & 35.9 & 87.7 & 22.9 & 18.8 & 26.3 & 71.6 & 21.6 & 21.5 \\
 & Qwen3-VL-4B & 32.0 & 88.2 & 19.9 & 16.6 & 22.5 & 74.4 & 17.8 & 18.1 \\
 & Qwen3-VL-8B & 31.5 & 88.5 & 19.4 & 16.2 & 22.1 & 73.4 & 17.6 & 17.9 \\
Pruned & InternVL3-1B & 45.8 & 80.7 & 31.4 & 26.9 & 28.3 & 64.5 & 24.1 & 24.1 \\
 & InternVL3-2B & 39.7 & 77.2 & 23.3 & 22.6 & 22.3 & 61.5 & 17.4 & 18.8 \\
 & InternVL3-8B & 36.3 & 75.9 & 20.0 & 20.6 & 17.9 & 57.0 & 12.4 & 15.0 \\
 & InternVL3-14B & 34.3 & 73.0 & 16.3 & 19.0 & 17.6 & 54.6 & 11.7 & 14.8 \\
\bottomrule
\end{tabular}
\end{table*}

%% file: sections/tab_app_views.tex
\begin{table*}[t]
\centering\footnotesize\setlength{\tabcolsep}{5.19pt}
\caption{\textbf{The full view at tighter budgets.} Each pair gives how often the disagreement of InternVL3-2B differs from $\delta^*$ (\%) when it reads the full image or the kept tokens of the large VLM, on the same checked questions. Evidence pruned is the share of pruning failures whose located evidence lies wholly in the pruned region (\%, text-rich benchmarks). The budgets of 2\% and 2.5\% share the first pass of 1\%, so their rows are the same.}
\label{tab:app-views}
\begin{tabular}{@{}lcccccccccc@{}}
\toprule
 & \multicolumn{2}{c}{Text-rich, InternVL3-38B} & \multicolumn{2}{c}{Text-rich, InternVL3-8B} & \multicolumn{2}{c}{General, InternVL3-38B} & \multicolumn{2}{c}{General, InternVL3-8B} & \multicolumn{2}{c}{Evidence pruned} \\
\cmidrule(lr){2-3}\cmidrule(lr){4-5}\cmidrule(lr){6-7}\cmidrule(lr){8-9}\cmidrule(lr){10-11}
Budget & Full & Pruned & Full & Pruned & Full & Pruned & Full & Pruned & 38B & 8B \\
\midrule
1\% & 8.2 & 23.1 & 7.3 & 23.8 & 13.3 & 24.8 & 14.3 & 25.6 & 49.2 & 47.2 \\
1.5\% & 9.4 & 22.7 & 8.8 & 23.6 & 13.6 & 23.0 & 14.6 & 23.5 & 44.2 & 41.5 \\
2\% & 10.1 & 22.1 & 9.4 & 22.5 & 13.9 & 21.7 & 14.8 & 22.4 & 38.9 & 36.8 \\
2.5\% & 10.1 & 22.1 & 9.4 & 22.5 & 13.9 & 21.7 & 14.8 & 22.4 & 38.9 & 36.8 \\
3\% & 11.5 & 21.9 & 10.6 & 21.5 & 14.3 & 20.1 & 15.1 & 20.7 & 34.1 & 31.6 \\
4\% & 12.2 & 21.9 & 11.2 & 21.2 & 14.5 & 19.2 & 15.4 & 19.9 & 30.9 & 27.8 \\
5\% & 13.0 & 21.4 & 11.9 & 20.8 & 14.6 & 18.5 & 15.5 & 19.2 & 26.8 & 24.4 \\
6\% & 13.4 & 21.4 & 12.5 & 21.1 & 14.7 & 18.3 & 15.6 & 18.8 & 23.4 & 21.7 \\
8\% & 14.2 & 21.2 & 13.2 & 20.5 & 15.0 & 17.7 & 15.9 & 18.5 & 19.5 & 17.9 \\
10\% & 14.7 & 20.8 & 13.9 & 20.5 & 15.2 & 17.8 & 16.1 & 18.3 & 17.8 & 15.9 \\
15\% & 15.9 & 20.8 & 14.8 & 19.9 & 15.5 & 17.4 & 16.5 & 18.0 & 13.8 & 12.0 \\
20\% & 16.4 & 20.4 & 15.7 & 20.1 & 15.7 & 17.4 & 16.6 & 17.9 & 12.3 & 8.9 \\
\bottomrule
\end{tabular}
\end{table*}

%% file: sections/tab_app_views_qwen.tex
\begin{table}[t]
\centering\footnotesize\setlength{\tabcolsep}{6.00pt}
\caption{\textbf{The full view at tighter budgets on Qwen3-VL-8B.} Columns as in Table~\ref{tab:app-views}, with Qwen3-VL-2B as the checker. The text-rich group has ChartQA, DocVQA, InfoVQA, TextVQA and OCRBench, the general group GQA and VQAv2. The budgets of 2\% and 2.5\% share the first pass of 1\%, so their rows are the same.}
\label{tab:app-views-qwen}
\begin{tabular}{@{}lcccc@{}}
\toprule
 & \multicolumn{2}{c}{Text-rich} & \multicolumn{2}{c}{General} \\
\cmidrule(lr){2-3}\cmidrule(lr){4-5}
Budget & Full & Pruned & Full & Pruned \\
\midrule
1\% & 4.6 & 18.9 & 9.9 & 39.1 \\
1.5\% & 5.9 & 19.0 & 10.1 & 39.3 \\
2\% & 6.9 & 18.5 & 10.2 & 38.5 \\
2.5\% & 6.9 & 18.5 & 10.2 & 38.5 \\
3\% & 8.2 & 17.1 & 10.4 & 37.9 \\
4\% & 8.8 & 16.1 & 11.2 & 34.0 \\
5\% & 9.3 & 15.7 & 11.6 & 30.4 \\
6\% & 9.9 & 15.9 & 12.1 & 26.1 \\
8\% & 10.3 & 15.1 & 13.0 & 20.1 \\
10\% & 10.8 & 15.9 & 13.6 & 18.2 \\
15\% & 11.7 & 17.1 & 14.2 & 17.0 \\
20\% & 12.1 & 17.5 & 14.9 & 16.9 \\
\bottomrule
\end{tabular}
\end{table}

%% file: sections/tab_app_mono.tex
\begin{table}[t]
\centering\footnotesize\setlength{\tabcolsep}{6.00pt}
\caption{\textbf{The conditions of Proposition~\ref{prop:order} on the budget grid.} The grid has the budgets of 2\%, 5\% and 10\%, on InternVL3-38B and -8B. Above is how often $r(1, p) \ge r(0, p)$ within a confidence decile. Falls is how often $r(1, p)$ falls as the decile rises, measured by a negative rank correlation, and Corr.\ is its median. Agreeing is the mean gain $r(0, p)$ of re-answering an agreeing question. On InternVL3-38B, GQA and VQAv2 leave out the questions whose first or full-token answer is the stock phrase of Appendix~\ref{app:setup}.}
\label{tab:app-mono}
\begin{tabular}{@{}lcccc@{}}
\toprule
 & Above (\%) & Falls (\%) & Corr. & Agreeing \\
\midrule
Text-rich 38B & 100.0 & 100.0 & $-$0.71 & $-$0.01 \\
Text-rich 8B & 98.3 & 100.0 & $-$0.76 & $-$0.02 \\
General 38B & 96.7 & 100.0 & $-$0.88 & $-$0.01 \\
General 8B & 86.7 & 100.0 & $-$0.72 & $-$0.01 \\
\bottomrule
\end{tabular}
\end{table}

%% file: sections/tab_app_pvgain.tex
\begin{table}[t]
\centering\footnotesize\setlength{\tabcolsep}{5.29pt}
\caption{\textbf{Pruned-view signals combined with labels.} A logistic regression, trained by five-fold cross-validation on each benchmark, combines the confidence of the large VLM, whether two pruned views give it the same answer, and the disagreement and token probability of InternVL3-2B, -8B and -14B on its kept tokens. It orders the questions by the predicted chance of a pruning failure or of a repair. Gains in points of the benchmark score over one-pass pruning, with the quota and abstention of PRUDENT.}
\label{tab:app-pvgain}
\begin{tabular}{@{}llccc@{}}
\toprule
 & & Ours & \multicolumn{2}{c}{Trained on the pruned view} \\
\cmidrule(lr){4-5}
 & Budget & (no labels) & Failures & Repairable \\
\midrule
\multirow{3}{*}{\shortstack[l]{Text-rich\\InternVL3-38B}} & 2.5\% & 1.8 & 0.9 & 0.8 \\
 & 5\% & 2.0 & 1.6 & 1.6 \\
 & 10\% & 1.5 & 1.2 & 1.2 \\
\midrule
\multirow{3}{*}{\shortstack[l]{Text-rich\\InternVL3-8B}} & 2.5\% & 2.4 & 1.3 & 1.2 \\
 & 5\% & 1.4 & 0.9 & 1.0 \\
 & 10\% & 1.7 & 1.6 & 1.5 \\
\midrule
\multirow{3}{*}{\shortstack[l]{General\\InternVL3-38B}} & 2.5\% & 1.4 & 1.1 & 1.1 \\
 & 5\% & 1.1 & 1.1 & 1.1 \\
 & 10\% & 0.7 & 0.7 & 0.7 \\
\midrule
\multirow{3}{*}{\shortstack[l]{General\\InternVL3-8B}} & 2.5\% & 1.4 & 1.1 & 1.1 \\
 & 5\% & 0.7 & 0.7 & 0.6 \\
 & 10\% & 0.7 & 0.6 & 0.6 \\
\bottomrule
\end{tabular}
\end{table}

%% file: sections/9b_appendix_exp.tex
\extrafloats{100}
\FloatBarrier
\section{Experimental Setup}
\label{app:setup}
\paragraph{Models.}
The large VLMs are InternVL3-8B, 14B, 38B and 78B, InternVL2.5-26B, Qwen3-VL-8B and 32B, and Gemma~4 31B.
Section~\ref{sec:setup} and Appendix~\ref{app:ranking} give the guide and the checker of each family, and all InternVL models share the same two.
InternVL splits each image into tiles of 448 pixels and adds a thumbnail.
All runs use H200 GPUs, and InternVL3-78B runs on two of them.

\paragraph{Benchmarks.}
Table~\ref{tab:app-benchmarks} lists the 18 benchmarks of the main tables and the 4 more of Appendix~\ref{app:main}.
They run in lmms-eval~\cite{zhang-etal-2025-lmms} with its default question prompts and metrics.
VQAv2 uses the first 12,612 questions of its validation split, and WebSRC uses every tenth question of its development split.
DeepSeek-V4.1-Flash scores the answers on MM-Vet for every method, full tokens included.
In the tables of the other two families, GQA and VQAv2 use a quarter of their questions and POPE a third.

\input{sections/appx/tab_app_benchmarks}

\paragraph{Existing methods.}
Every method selects its visual tokens before the first LLM layer of the large VLM, and it keeps the fraction $K$ of the visual tokens of each question.
FastV ranks the tokens by the attention that the last prompt token pays to them at layer 2 of a pass over all tokens, and the answer then reads only the kept tokens from the first layer.
VisionZip keeps the tokens that the [CLS] token of the vision encoder attends to most, and it merges the others into a few contextual tokens.
ToMe merges similar tokens by bipartite matching.
CDPruner, MMTok and ZOO-Prune follow their official code on the visual features of the large VLM.
SGL ranks the tokens by the attention of InternVL2.5-1B, the same model as our guide.
We run it without its early exit, as its released code does, so the answers of every method come from the large VLM.
IF-Prune runs its official code with its importance scores in 32-bit floating point, and its scores stay within two points of those in its paper (Table~\ref{tab:app-repro}).
For CDPruner on InternVL3-8B, our port keeps more of the full-token score than the run that HAWK~\cite{zhu2026hawk} reports (Table~\ref{tab:app-cdpruner}).
At a loose budget on the other two families, the ports keep 89\% to 99\% of the full-token score on average (Table~\ref{tab:app-ports}).

\begin{table}[htbp]\centering
\begingroup
\renewenvironment{table}[1][]{\begin{minipage}[t]{0.485\textwidth}\vspace{0pt}\centering}{\end{minipage}}%
\renewenvironment{table*}[1][]{\begin{minipage}[t]{0.485\textwidth}\vspace{0pt}\centering}{\end{minipage}}%
\input{sections/appx/tab_app_repro}\hfill\input{sections/appx/tab_app_cdpruner}\endgroup
\end{table}

\input{sections/appx/tab_app_ports}

\paragraph{Statistics.}
Decoding is greedy, and each setting runs once.
Two methods are compared by a paired bootstrap over the questions of a setting, and a difference is significant when its 95\% interval excludes zero.
MME resamples images within each of its categories and recomputes its total score.
MMBench resamples each question together with its rotated copies.

\paragraph{Two configurations of InternVL3.}
All runs of InternVL models as the large VLM use the literal system prompt \texttt{None}, as the code of IF-Prune does.
On a random sample of each benchmark, the official default prompt changes the score little, and every 95\% interval of the change includes zero (Table~\ref{tab:app-sysmsg}).
Every method and budget uses the same prompt.
With full tokens, InternVL3-38B answers 11\% to 22\% of the questions of GQA and VQAv2 with a stock phrase.
The phrase says that answering does not require reading text in the image.
With the official default prompt, 75\% of 150 such questions still get the phrase, so the phrase comes from the model itself.
Pruned runs can therefore score above full tokens on these two benchmarks.
Table~\ref{tab:app-phrase} scores every method again on the other questions.
With these questions kept, PRUDENT still scores highest at every budget, by 0.5 to 1.8 points on GQA and by 2.5 to 3.7 points on VQAv2.
Every score of InternVL3-38B on these two benchmarks, and every ratio to its full-token score, leaves out these questions, in the main text and in this supplement.
Only the counts over single questions in Figures~\ref{fig:finding}b--c and~\ref{fig:analysis}g and in Appendices~\ref{app:findings} and~\ref{app:why} keep them, and there a question with the stock phrase is never a pruning failure, since its full-token answer is wrong.

\begin{table}[htbp]\centering
\begingroup
\renewenvironment{table}[1][]{\begin{minipage}[t]{0.485\textwidth}\vspace{0pt}\centering}{\end{minipage}}%
\renewenvironment{table*}[1][]{\begin{minipage}[t]{0.485\textwidth}\vspace{0pt}\centering}{\end{minipage}}%
\input{sections/appx/tab_app_phrase}\hfill\input{sections/appx/tab_app_sysmsg}\endgroup
\end{table}
\FloatBarrier
\section{Pruning Failures and the Checker}
\label{app:findings}
Figure~\ref{fig:finding}a counts the pruning failures of IF-Prune among the questions of the six text-rich benchmarks whose evidence we locate in the image (Appendix~\ref{app:why}) and that the full-token run answers correctly.
IF-Prune has the highest average of the existing methods at every budget of Tables~\ref{tab:main-38b} and~\ref{tab:main-8b}, and the fewest pruning failures.

Table~\ref{tab:app-fail-methods} gives the values of Figure~\ref{fig:finding}b for each existing method, with two more columns.
For every method and budget, the checker answers most of the pruning failures correctly and disagrees with the pruned answer on most of them in every setting.
On the questions that the checker answers, ordering them by the disagreement and then by the confidence finds the failures better than the confidence alone in almost every setting.

\input{sections/appx/tab_app_fail_methods}

Table~\ref{tab:app-complement} gives the pruning failures of IF-Prune for each model, budget and benchmark.
On most benchmarks the checker is right on more than half of them, on both models and at every budget.

\input{sections/appx/tab_app_complement}

Table~\ref{tab:app-auroc} gives the AUROC of the eight signals of Figure~\ref{fig:finding}c on each benchmark.
Ordering the questions by the disagreement of the checker and then by the confidence of the large VLM gives the highest AUROC on every benchmark of both models.
Two of the signals read only the kept tokens, the confidence of the large VLM and the agreement of two prunings.
On the text-rich benchmarks, both fall below the disagreement of the 2B checker, which reads the full image.

\input{sections/appx/tab_app_auroc_wide}
\FloatBarrier
\section{Main Results}
\label{app:main}
Table~\ref{tab:app-signif} gives the difference between PRUDENT and IF-Prune or SGL in every setting of Tables~\ref{tab:main-38b} and~\ref{tab:main-8b}, with its significance.
Table~\ref{tab:app-signif-counts} counts the settings in which PRUDENT is significantly better than each existing method, and PRUDENT is significantly worse in none.
A Holm correction over all 864 of these comparisons keeps PRUDENT significantly better in at least 70 of the 108 settings against each method, and significantly worse in none.
On the seven multiple-choice and yes-or-no benchmarks, the guide ranks the tokens as IF-Prune does, so the differences from IF-Prune there come from re-answering alone.
There PRUDENT is significantly better in 29 of these 42 settings and worse in none.

\input{sections/appx/tab_app_signif}

\input{sections/appx/tab_app_signif_counts_wide}

\begin{figure*}[t]
\centering
\includegraphics{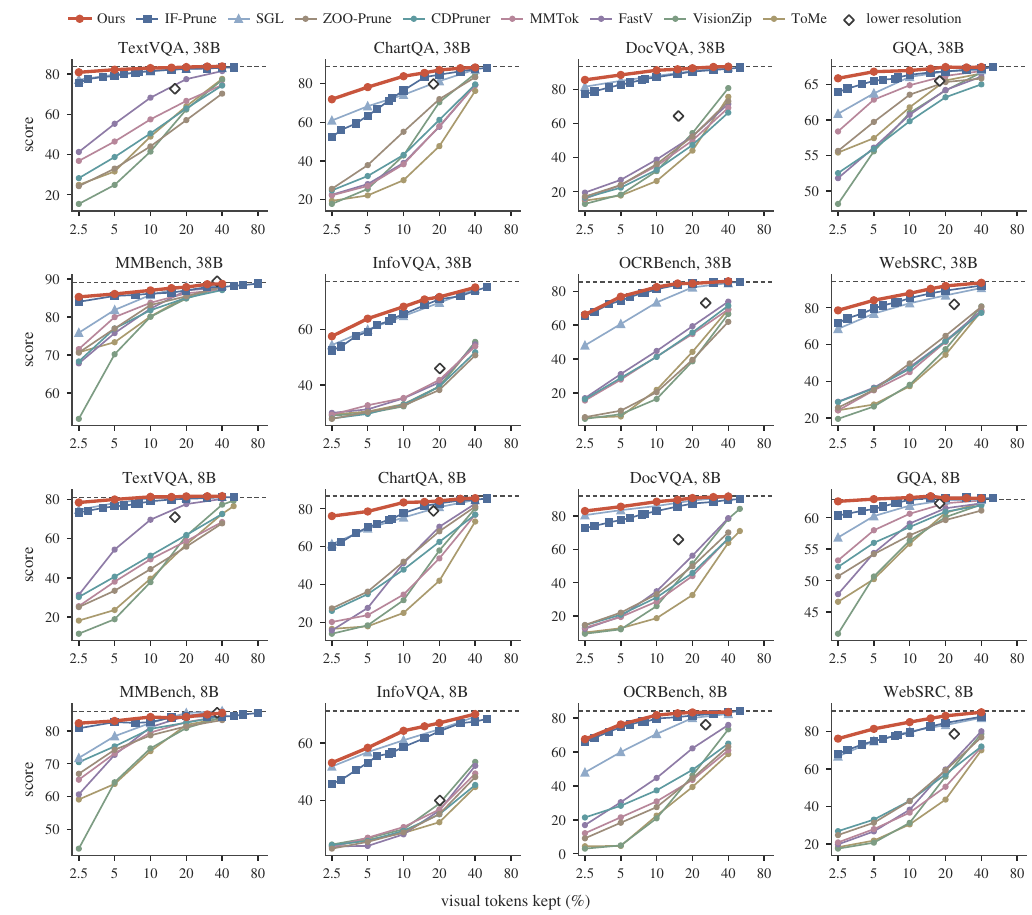}
\caption{\textbf{The frontier of Figure~\ref{fig:hero} on each benchmark.} Scores at every budget that we ran, on eight benchmarks, with InternVL3-38B in the top two rows and 8B in the bottom two. Figure~\ref{fig:hero} averages the first five over the five budgets that every method has. The dashed line is the full-token score, and the diamond feeds only the thumbnail of the image (Table~\ref{tab:app-lowres}).}
\label{fig:app-frontier}
\end{figure*}

In Figure~\ref{fig:hero}, the band is the lead of PRUDENT over the best existing method, and the dotted line ends at the budget at which IF-Prune or SGL first matches the score of PRUDENT at 2.5\%.
Figure~\ref{fig:app-frontier} draws the curves of Figure~\ref{fig:hero} on each of eight benchmarks, on a linear axis and with every budget that we ran.
Table~\ref{tab:app-frontier} gives the two budgets of Figure~\ref{fig:hero} beyond those of the main tables.
PRUDENT has the highest average of the eight benchmarks at both budgets and on both models.

\input{sections/appx/tab_app_frontier}

\begin{figure*}[t]
\centering
\includegraphics{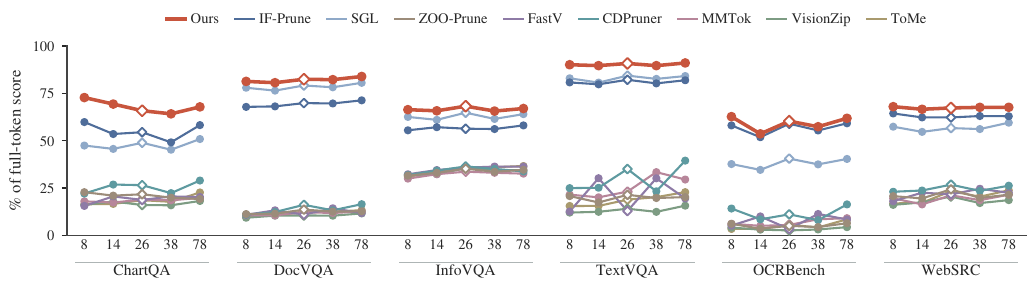}
\caption{\textbf{The 1\% budget per benchmark.} The values behind Table~\ref{tab:extreme}, as the share of the full-token score on the six text-rich benchmarks and five model sizes, with InternVL2.5-26B in open markers. PRUDENT (Ours) answers first with 0.5\% and re-answers with 2\%.}
\label{fig:app-extreme}
\end{figure*}

Figure~\ref{fig:app-extreme} gives Table~\ref{tab:extreme} on each text-rich benchmark.
At this budget, PRUDENT has the top score on each of the six benchmarks at every model size.
Against every existing method, it is significantly better in almost every setting of Table~\ref{tab:extreme} and significantly worse in none.

Table~\ref{tab:app-bench} adds MMMU, MMStar, RealWorldQA and MM-Vet.
PRUDENT scores highest in most settings of these four benchmarks.

\begin{table}[htbp]\centering
\begingroup
\renewenvironment{table}[1][]{\begin{minipage}[t]{0.485\textwidth}\vspace{0pt}\centering}{\end{minipage}}%
\renewenvironment{table*}[1][]{\begin{minipage}[t]{0.485\textwidth}\vspace{0pt}\centering}{\end{minipage}}%
\input{sections/appx/tab_app_more_bench}\hfill\input{sections/appx/tab_app_lowres}\endgroup
\end{table}

Lowering the resolution also spends fewer visual tokens (Table~\ref{tab:app-lowres}).
Feeding only the thumbnail of 256 tokens per image still uses at least 1.5 times the tokens of PRUDENT at 10\%.
It loses much of the score on the text-rich benchmarks, whose answers rest on small text.

\FloatBarrier
\section{Generalization}
\label{app:general}
Tables~\ref{tab:app-size-14b} to~\ref{tab:app-size-78b} give the nine benchmarks of Figure~\ref{fig:analysis}a on InternVL3-14B, InternVL2.5-26B and InternVL3-78B, in the layout of Table~\ref{tab:main-38b}.
PRUDENT has the highest average at every budget on each of these sizes.

\input{sections/appx/tab_app_size_14b}

\input{sections/appx/tab_app_size_26b}

\input{sections/appx/tab_app_size_78b}

Tables~\ref{tab:app-family-q8b} to~\ref{tab:app-family-g31b} give all 10 benchmarks of the other two families at the budget of Table~\ref{tab:family}.
PRUDENT has the highest average over the 10 benchmarks on each of the three models.
IF-Prune is left out of these tables and of Table~\ref{tab:family}, since its released importance estimator is trained for InternVL2.5-1B.
Each family uses the budget at which SGL keeps about 90\% of the full-token score, as IF-Prune does at 5\% on InternVL3.

\input{sections/appx/tab_app_family_q8b}

\input{sections/appx/tab_app_family_q32b}

\input{sections/appx/tab_app_family_g31b}

The bars of Figure~\ref{fig:analysis}b average InternVL3-38B and 8B at 10\% and 20\%, and the dots are the settings.
Table~\ref{tab:app-plugin} gives Figure~\ref{fig:analysis}b on each benchmark and adds the budget of 5\%.
With the ranking of IF-Prune or SGL in both passes, PRUDENT is significantly better than the method alone in most settings, and no setting drops significantly.

\input{sections/appx/tab_app_plugin}

\input{sections/appx/tab_app_checker_means}

Table~\ref{tab:app-checker} gives the last block of Table~\ref{tab:ablation} for each large VLM, with IF-Prune for reference.
The checker is always smaller than the large VLM, so the 8B checker runs on InternVL3-38B only.
These tables leave out RefCOCO, on which the checker abstains on almost every question.
Tables~\ref{tab:app-checker-38b} and~\ref{tab:app-checker-8b} give Table~\ref{tab:app-checker} on each benchmark.
Table~\ref{tab:app-checker-auroc} gives the AUROC of each checker on InternVL3-38B.
Its first passes of 1\%, 2.5\% and 5\% belong to the budgets of 2.5\%, 5\% and 10\%.
The median AUROC rises with the size of the checker at every budget.
On Qwen3-VL, the 2B and 4B checkers lift PRUDENT above SGL by about the same margin (Table~\ref{tab:app-checker-qwen}).

\input{sections/appx/tab_app_checker_38b}

\input{sections/appx/tab_app_checker_8b}

\begin{table}[htbp]\centering
\begingroup
\renewenvironment{table}[1][]{\begin{minipage}[t]{0.485\textwidth}\vspace{0pt}\centering}{\end{minipage}}%
\renewenvironment{table*}[1][]{\begin{minipage}[t]{0.485\textwidth}\vspace{0pt}\centering}{\end{minipage}}%
\input{sections/appx/tab_app_checker_auroc}\hfill\input{sections/appx/tab_app_checker_qwen}\endgroup
\end{table}
\FloatBarrier
\section{Cost}
\label{app:cost}
Table~\ref{tab:app-cost} times every method at a 5\% budget on ChartQA, DocVQA, GQA, InfoVQA, MMBench, OCRBench and TextVQA.
Every method answers the same questions.
Each question runs alone, on one H200 GPU for InternVL3-38B and on two for InternVL3-78B.
IF-Prune already loads the 1B model for its importance estimate.
The extra memory of PRUDENT over IF-Prune is therefore close to the 4.3 GB of the weights of the checker, on both models.
With the guide as the checker (Ours, 1B check), PRUDENT drops the pass of InternVL3-2B.
Its latency sums the measured stages of PRUDENT without that pass and takes the guide stage of IF-Prune + draft, since the guide then answers every question.
It needs the memory of IF-Prune + draft and about its FLOPs, 0.4 and 0.5 points more on InternVL3-38B and 78B, and it scores 4.8 and 3.8 points higher.

\input{sections/appx/tab_app_cost}
\FloatBarrier
\section{Ablations}
\label{app:ablation}
Tables~\ref{tab:app-ablate-38b} and~\ref{tab:app-ablate-8b} give the first two blocks of Table~\ref{tab:ablation} on each benchmark.
Every change lowers the mean of the text-rich benchmarks at every budget, and none raises the mean of the general ones.

\input{sections/appx/tab_app_ablate_38b}

\input{sections/appx/tab_app_ablate_8b}

\begin{figure*}[t]
\centering
\includegraphics{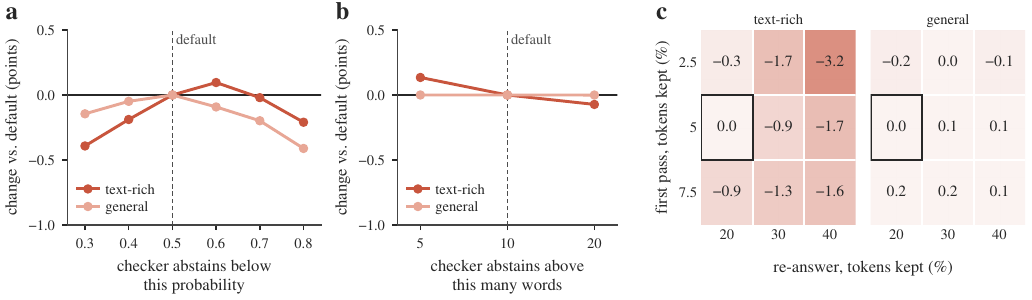}
\caption{\textbf{Abstention of the checker and the split of the budget.} (a) The least token probability below which the checker abstains and (b) the answer length above which it abstains, as the change from the default in points, averaged over both models, three budgets and the benchmarks of each group. (c) The split of a 10\% budget between the first pass and the re-answers. Tables~\ref{tab:app-abstain-38b}, \ref{tab:app-abstain-8b} and~\ref{tab:app-split10} give every benchmark.}
\label{fig:app-hparams}
\end{figure*}

Figure~\ref{fig:app-hparams}a--b and Tables~\ref{tab:app-abstain-38b} and~\ref{tab:app-abstain-8b} move the two points at which the checker abstains.
The probability $\tau$ ranges from 0.3 to 0.8, and the length $\ell$ is 5 or 20 words.
Each such move changes the mean of each group by less than one point.

\input{sections/appx/tab_app_abstain_38b}

\input{sections/appx/tab_app_abstain_8b}

Figure~\ref{fig:app-hparams}c and Table~\ref{tab:app-split10} split a 10\% budget in nine ways, with first passes of 2.5\%, 5\% and 7.5\% and re-answers of 20\%, 30\% and 40\%.
The default split keeps 5\% in the first pass and 20\% in the re-answers.
It stays within one point of the best split on every benchmark.

\input{sections/appx/tab_app_split10}
\FloatBarrier
\section{Analysis}
\label{app:why}
We locate the evidence of an answer by matching the answer to the words that OCR finds in the image.
InfoVQA and TextVQA provide these words, and RapidOCR reads them on DocVQA, ChartQA, WebSRC and OCRBench.
The evidence tokens are the visual tokens whose image regions overlap these words.
The budget needed to keep the evidence is the share of the visual tokens up to the first evidence token in a ranking.
Figures~\ref{fig:finding}a and~\ref{fig:analysis}d--f give this share in units of $K$, under the ranking of IF-Prune and of PRUDENT.
Figure~\ref{fig:analysis}d--f pools InternVL3-38B and 8B at three budgets.
Panel (d) uses the checked questions that the full-token run answers correctly, and at 2.5\% the first pass keeps 1\% (Appendix~\ref{app:budget}).

\begin{figure*}[t]
\centering
\includegraphics{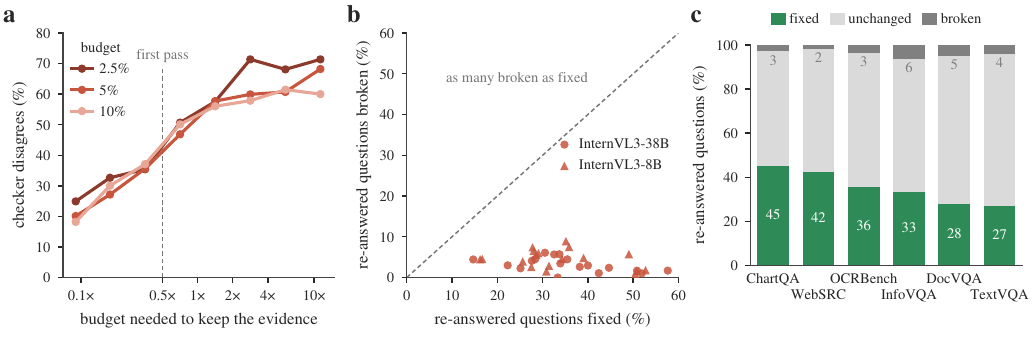}
\caption{\textbf{The evidence chain of Figure~\ref{fig:analysis}d--f in more detail.} All panels use the questions with located evidence on the six text-rich benchmarks. (a) Disagreement of the checker by the budget needed to keep the evidence, one line per budget. (b) Re-answered questions fixed against those broken, one point per model, benchmark and budget. (c) Outcomes of the re-answers per benchmark, both models and three budgets pooled.}
\label{fig:app-evidence}
\end{figure*}

In Figure~\ref{fig:analysis}d the checker rarely agrees with a wrong first answer, and in Figure~\ref{fig:analysis}f most fixes fall on questions whose evidence lies between the two budgets.
Figure~\ref{fig:app-evidence} splits Figure~\ref{fig:analysis}d--f further.
At every budget, the checker disagrees more often as the evidence lies farther beyond the first pass (Figure~\ref{fig:app-evidence}a).
In every setting, the re-answers fix at least three times as many questions as they break (Figure~\ref{fig:app-evidence}b).
Figure~\ref{fig:app-evidence}c gives their outcomes on each benchmark.

\begin{figure*}[t]
\centering
\includegraphics{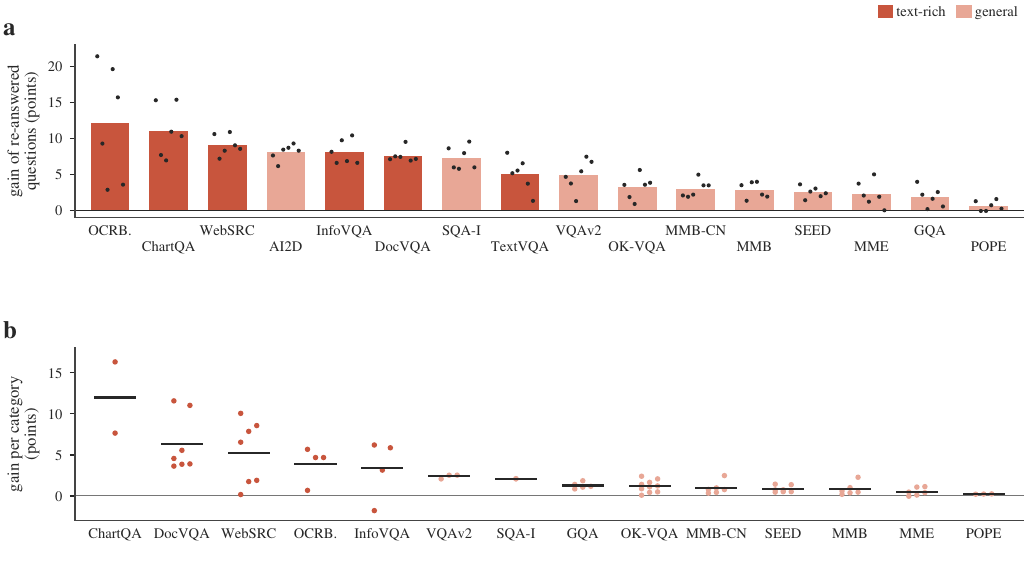}
\caption{\textbf{Per-benchmark views of Figure~\ref{fig:analysis}g--i.} (a) Gain of the re-answered questions over answering them once with $K$ tokens, the mean of the six settings of each benchmark as a bar and the settings as dots. (b) Gain of PRUDENT (Ours) over IF-Prune in each question category with at least 100 questions, the mean of its six settings as a dot and the mean of each benchmark as a dash.}
\label{fig:app-where}
\end{figure*}

In Figure~\ref{fig:analysis}g, the first pass hurts a question when its first answer scores below the full-token answer.
Table~\ref{tab:app-where} gives Figure~\ref{fig:analysis}g on each benchmark.
In every setting, first answers below the full-token answer are at least twice as common among the re-answered questions as among the kept ones.
The re-answered questions are 14\% to 30\% of all questions in a setting, with a median of 23\%.
With both of their passes, they take 35\% to 72\% of the visual tokens, with a median of 57\%.
The re-answered questions also gain over one pass at $K$ on every benchmark (Figure~\ref{fig:app-where}a).

\input{sections/appx/tab_app_where}

\begin{figure*}[t]
\centering
\includegraphics{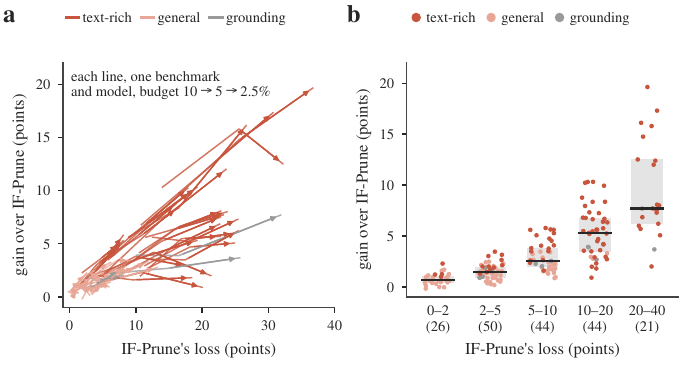}
\caption{\textbf{Gain over IF-Prune against its loss.} The settings are those of Figure~\ref{fig:analysis}h. (a) One line per benchmark and model, from 10\% to 5\% to 2.5\%. (b) The settings in five bins of the loss, with the median and the middle half of each bin, and the number of settings under each bin.}
\label{fig:app-room}
\end{figure*}

Figure~\ref{fig:analysis}h plots the gain of PRUDENT over IF-Prune against the loss of IF-Prune in every setting of five InternVL models.
Figure~\ref{fig:app-room} follows each benchmark and model across the three budgets.
As the budget tightens, the loss of IF-Prune and the gain of PRUDENT mostly grow together (Figure~\ref{fig:app-room}a).
Across bins of the loss, the median gain rises with the loss (Figure~\ref{fig:app-room}b).

Figure~\ref{fig:analysis}i averages each category over its six settings.
Figure~\ref{fig:app-where}b places each category of at least 100 questions under its benchmark, and the gain is positive in almost all of them.
Table~\ref{tab:app-lost} counts the correct answers that pruning costs each method, and PRUDENT loses the fewest at every budget of both models.
Table~\ref{tab:app-fixed} compares PRUDENT with IF-Prune question by question.
PRUDENT fixes more questions than it breaks in almost every setting, and it answers 21\% to 68\% of the pruning failures of IF-Prune correctly.

\input{sections/appx/tab_app_lost_wide}

\input{sections/appx/tab_app_fixed}

Appendix~\ref{app:examples} shows seven questions on which the first pass loses the evidence and the re-answer recovers it.

\FloatBarrier
\section{Limitations}
\label{app:limits}
Like other methods that compare the answers of two models, PRUDENT relies on answers that can be compared.
It compares short answers by their format (Appendix~\ref{app:match}), and its checker abstains on long ones, which PRUDENT answers once at the budget $K$.
Matching answers by their meaning could extend it to longer answers.
Proposition~\ref{prop:views} bounds the error of the checker by how often it departs from the full-token answer, so a checker that answers more like the large VLM gives a cleaner signal.

Our experiments cover questions about single images, in English and Chinese, on three model families.
Videos, inputs with several images and multi-turn dialogue are left to future work.

Like other work on visual token pruning, we state budgets in visual tokens of the large VLM.
The time that a saved token is worth depends on the serving system, since batching, caching and the attention kernel all change the cost of a token.
A re-answer reads the tokens of the first pass among its own, so it could reuse part of the computation of the first pass.

PRUDENT only changes which visual tokens the large VLM reads.
Its answers keep the errors and biases of the large VLM, and they need the same checks as any output of that model.

%% file: sections/appx/tab_app_benchmarks.tex
\begin{table}[t]
\centering\scriptsize\setlength{\tabcolsep}{3pt}
\caption{\textbf{Benchmarks.} Questions is the number of questions that every method answers. The last four appear in Table~\ref{tab:app-bench}.}
\label{tab:app-benchmarks}
\begin{tabular}{@{}llrl@{}}
\toprule
Benchmark & Split & Questions & Metric \\
\midrule
\multicolumn{4}{@{}l}{\textit{Text-rich}} \\
ChartQA~\cite{masry-etal-2022-chartqa} & test & 2,500 & relaxed accuracy \\
DocVQA~\cite{9423358} & validation & 5,349 & ANLS \\
InfoVQA~\cite{9706887} & validation & 2,801 & ANLS \\
TextVQA~\cite{8953586} & validation & 5,000 & VQA accuracy \\
OCRBench~\cite{Liu_2024} & test & 1,000 & accuracy \\
WebSRC~\cite{chen-etal-2021-websrc} & every 10th of dev & 5,283 & F1 \\
\midrule
\multicolumn{4}{@{}l}{\textit{General}} \\
GQA~\cite{Hudson_2019_CVPR} & test-dev & 12,578 & accuracy \\
VQAv2~\cite{10.1007/s11263-018-1116-0} & start of validation & 12,612 & VQA accuracy \\
OK-VQA~\cite{8953725} & validation & 5,046 & VQA accuracy \\
MMBench~\cite{10.1007/978-3-031-72658-3_13} & dev & 4,329 & circular accuracy \\
MMBench-CN & dev & 4,329 & circular accuracy \\
MME~\cite{fu2026mme} & -- & 2,374 & total of 2800 \\
POPE~\cite{li-etal-2023-evaluating} & test & 9,000 & F1 \\
SEED-Bench~\cite{Li_2024_CVPR} & image questions & 14,233 & accuracy \\
AI2D~\cite{kembhavi2016diagram} & test & 3,088 & accuracy \\
ScienceQA-IMG~\cite{lu2022learn} & test with image & 2,017 & accuracy \\
\midrule
\multicolumn{4}{@{}l}{\textit{Grounding}} \\
RefCOCO testA~\cite{yu2016modeling} & testA & 5,657 & accuracy at IoU 0.5 \\
RefCOCO testB & testB & 5,095 & accuracy at IoU 0.5 \\
\midrule
\multicolumn{4}{@{}l}{\textit{More benchmarks}} \\
MMMU~\cite{10656299} & validation, one image & 861 & accuracy \\
MMStar~\cite{chen2024are} & -- & 1,500 & accuracy \\
RealWorldQA & -- & 765 & accuracy \\
MM-Vet~\cite{3692070.3694451} & -- & 218 & score of a judge model \\
\bottomrule
\end{tabular}
\end{table}

%% file: sections/appx/tab_app_repro.tex
\begin{table}[t]
\centering\footnotesize\setlength{\tabcolsep}{4pt}
\caption{\textbf{Reproducing IF-Prune.} Both models of its paper prune at layer 0. Paper gives its Tables 1 and 2, bf16 our run of the official code, and fp32 the same run with the importance scores in fp32, the IF-Prune of all our tables.}
\label{tab:app-repro}
\begin{tabular}{@{}lrrrrrrrr@{}}
    \toprule
     & \multicolumn{2}{c}{Full tokens} & \multicolumn{3}{c}{Retain 20\%} & \multicolumn{3}{c}{Retain 5\%} \\
    \cmidrule(lr){2-3}\cmidrule(lr){4-6}\cmidrule(lr){7-9}
     & Paper & Our run & Paper & bf16 & fp32 & Paper & bf16 & fp32 \\
    \midrule
    \multicolumn{9}{@{}l}{\textit{InternVL2-26B}} \\
    TextVQA & 82.5 & 82.1 & 81.3 & 81.0 & 80.9 & 79.0 & 78.5 & 78.5 \\
    ChartQA & 84.9 & 84.5 & 82.4 & 81.7 & 82.0 & 71.0 & 71.7 & 71.9 \\
    GQA & 64.9 & 65.2 & 64.9 & 64.8 & 65.0 & 63.5 & 63.6 & 63.5 \\
    MMBench & 83.5 & 81.7 & 80.0 & 79.6 & 80.1 & 77.2 & 76.4 & 76.8 \\
    MMStar & 60.1 & 59.2 & 56.5 & 54.9 & 54.6 & 52.5 & 52.1 & 52.1 \\
    RWQA & 67.6 & 67.6 & 66.5 & 65.9 & 66.9 & 66.0 & 65.5 & 65.9 \\
    \multicolumn{9}{@{}l}{\textit{InternVL2-8B}} \\
    GQA & 62.7 & 62.7 & 62.5 & 61.3 & 62.2 & 58.5 & 58.5 & 59.0 \\
    MMBench & 81.9 & 82.0 & 79.6 & 77.9 & 80.3 & 76.5 & 75.2 & 76.7 \\
    MMStar & 59.1 & 58.3 & 56.9 & 53.9 & 55.8 & 53.3 & 50.7 & 51.7 \\
    RWQA & 65.1 & 64.6 & 63.1 & 62.7 & 63.5 & 62.5 & 60.4 & 62.1 \\
    \bottomrule
  \end{tabular}
\end{table}

%% file: sections/appx/tab_app_cdpruner.tex
\begin{table}[t]
\centering\footnotesize\setlength{\tabcolsep}{4pt}
\caption{\textbf{CDPruner on InternVL3-8B.} HAWK reports CDPruner with a fixed input of 1280 tokens in VLMEvalKit (its Table 2), and our port uses dynamic tiles in lmms-eval. Our port keeps a larger share of the full-token score at both budgets.}
\label{tab:app-cdpruner}
\begin{tabular}{@{}lrrrr@{}}
    \toprule
     & \multicolumn{2}{c}{TextVQA} & \multicolumn{2}{c}{ChartQA} \\
    \cmidrule(lr){2-3}\cmidrule(lr){4-5}
    InternVL3-8B & HAWK & Our port & HAWK & Our port \\
    \midrule
    Full tokens & 79.9 & 80.9 & 80.4 & 86.6 \\
    CDPruner, retain 40\% & 55.8 & 72.5 & 52.9 & 76.8 \\
    CDPruner, retain 20\% & 41.8 & 61.8 & 37.9 & 62.5 \\
    \bottomrule
  \end{tabular}
\end{table}

%% file: sections/appx/tab_app_ports.tex
\begin{table*}[t]
\centering\scriptsize\setlength{\tabcolsep}{9.21pt}
\caption{\textbf{Ports to the other two families.} The budgets are loose, 50\% of the visual tokens on Qwen3-VL and 60\% on Gemma 4, and every benchmark uses a quarter of its questions. Each value is the share of the full-token score on the same questions (\%). ZOO-Prune is not defined on Gemma 4.}
\label{tab:app-ports}
\begin{tabular}{lccccccccccc}
\toprule
Method & ChartQA & DocVQA & InfoVQA & TextVQA & OCRB. & GQA & VQAv2 & OK-VQA & MMB & POPE & Mean \\
\midrule
\multicolumn{12}{@{}l}{\textit{Qwen3-VL-8B, retain 50\%}} \\
ToMe & 79.9 & 89.8 & 79.5 & 94.7 & 76.8 & 99.4 & 96.9 & 94.6 & 98.4 & 98.1 & 90.8 \\
VisionZip & 93.5 & 97.8 & 90.8 & 98.9 & 88.2 & 97.2 & 98.0 & 97.3 & 98.7 & 100.0 & 96.0 \\
ZOO-Prune & 91.0 & 95.0 & 83.9 & 91.2 & 92.4 & 98.4 & 98.2 & 91.7 & 98.4 & 102.2 & 94.2 \\
CDPruner & 89.2 & 87.0 & 72.6 & 94.0 & 88.2 & 96.6 & 98.6 & 98.2 & 98.4 & 99.5 & 92.2 \\
MMTok & 93.1 & 95.2 & 85.7 & 96.6 & 86.3 & 98.4 & 98.1 & 98.1 & 98.8 & 100.0 & 95.0 \\
\midrule
\multicolumn{12}{@{}l}{\textit{Qwen3-VL-32B, retain 50\%}} \\
ToMe & 82.4 & 91.6 & 74.8 & 93.6 & 78.3 & 100.6 & 97.7 & 99.6 & 98.5 & 96.6 & 91.4 \\
VisionZip & 98.4 & 98.7 & 91.8 & 97.2 & 89.4 & 99.8 & 96.7 & 98.0 & 97.5 & 99.4 & 96.7 \\
ZOO-Prune & 90.7 & 96.3 & 85.9 & 95.0 & 89.4 & 100.2 & 96.1 & 98.5 & 98.7 & 98.4 & 94.9 \\
CDPruner & 82.2 & 86.8 & 66.5 & 92.4 & 77.4 & 97.8 & 93.3 & 95.9 & 97.2 & 96.2 & 88.6 \\
MMTok & 95.3 & 97.6 & 88.0 & 97.5 & 94.5 & 100.0 & 97.9 & 98.5 & 98.5 & 101.4 & 96.9 \\
\midrule
\multicolumn{12}{@{}l}{\textit{Gemma 4 31B, retain 60\%}} \\
ToMe & 93.5 & 91.2 & 76.7 & 99.2 & 87.8 & 98.4 & 98.7 & 97.1 & 101.8 & 99.1 & 94.3 \\
VisionZip & 98.6 & 97.4 & 94.6 & 99.7 & 94.6 & 99.5 & 98.7 & 100.0 & 100.9 & 100.9 & 98.5 \\
CDPruner & 99.3 & 94.0 & 88.1 & 98.6 & 95.1 & 98.2 & 98.2 & 99.1 & 101.3 & 99.1 & 97.1 \\
MMTok & 99.8 & 98.2 & 94.8 & 100.3 & 97.1 & 99.2 & 98.8 & 100.2 & 100.6 & 100.5 & 98.9 \\
\bottomrule
\end{tabular}
\end{table*}

%% file: sections/appx/tab_app_phrase.tex
\begin{table}[t]
\centering\scriptsize\setlength{\tabcolsep}{8.91pt}
\caption{\textbf{InternVL3-38B on the questions without the stock phrase.} On GQA and VQAv2, these are the 11,187 and 9,851 questions that its full-token run answers normally. On the other 11\% to 22\%, it answers with full tokens with the stock phrase of its TextVQA prompt, which says that answering does not require reading text in the image. This phrase pulls its full-token score below the pruned runs. Every method is scored again on the same subset. \textbf{Best} in bold, \protect\underline{second}{} underlined.}
\label{tab:app-phrase}
\begin{tabular}{lcccccc}
\toprule
 & \multicolumn{3}{c}{GQA} & \multicolumn{3}{c}{VQAv2} \\
\cmidrule(lr){2-4}\cmidrule(lr){5-7}
Method & 2.5\% & 5\% & 10\% & 2.5\% & 5\% & 10\% \\
\midrule
\rowcolor{gray!12} Full tokens & 67.5 & 67.5 & 67.5 & 83.4 & 83.4 & 83.4 \\
\midrule
ToMe & 55.4 & 57.4 & 61.8 & 62.0 & 64.6 & 70.2 \\
VisionZip & 48.2 & 55.6 & 61.0 & 51.7 & 61.8 & 70.1 \\
ZOO-Prune & 55.7 & 59.7 & 63.6 & 63.8 & 65.3 & 66.9 \\
FastV & 51.8 & 56.1 & 60.7 & 57.8 & 65.1 & 72.5 \\
CDPruner & 52.5 & 55.9 & 59.8 & 58.2 & 62.8 & 67.6 \\
MMTok & 58.4 & 62.9 & 64.9 & 67.4 & 70.2 & 73.3 \\
SGL & 60.9 & 63.7 & 66.1 & 70.4 & 74.8 & \underline{78.8} \\
IF-Prune & \underline{63.9} & \underline{65.5} & \underline{66.4} & \underline{74.2} & \underline{76.1} & \underline{78.8} \\
\rowcolor{oursbg} \textbf{Ours} & \textbf{65.9} & \textbf{66.8} & \textbf{67.0} & \textbf{77.7} & \textbf{79.6} & \textbf{81.2} \\
\bottomrule
\end{tabular}
\end{table}

%% file: sections/appx/tab_app_sysmsg.tex
\begin{table}[t]
\centering\scriptsize\setlength{\tabcolsep}{2.13pt}
\caption{\textbf{The system prompt of InternVL3.} Both prompts run with full tokens on a random sample of each benchmark. All our runs of the large VLM use the literal prompt \texttt{None}, as the IF-Prune code does. Diff.\ is the official default minus \texttt{None}, with its paired bootstrap 95\% interval, and every interval includes zero. Every method and budget uses the same prompt. MMBench is scored per rotated copy here, without the circular evaluation of Tables~\ref{tab:main-38b} and~\ref{tab:main-8b}.}
\label{tab:app-sysmsg}
\begin{tabular}{@{}llccccc@{}}
\toprule
Model & Benchmark & Questions & \texttt{None} & Official & Diff. & 95\% CI \\
\midrule
InternVL3-8B & ChartQA & 300 & 85.0 & 84.7 & $-0.33$ & [$-1.33$, $+0.67$] \\
 & DocVQA & 300 & 92.7 & 91.7 & $-1.07$ & [$-2.79$, $+0.59$] \\
 & TextVQA & 300 & 80.8 & 81.8 & $+1.07$ & [$-0.33$, $+2.50$] \\
 & MMB & 300 & 94.0 & 94.3 & $+0.33$ & [$0.00$, $+1.00$] \\
 & VQAv2 & 300 & 78.6 & 78.6 & $-0.03$ & [$-1.30$, $+1.13$] \\
 & GQA & 300 & 62.0 & 61.3 & $-0.67$ & [$-2.33$, $+1.00$] \\
 & POPE & 300 & 87.3 & 87.0 & $-0.33$ & [$-1.33$, $0.00$] \\
 & OCRBench & 300 & 83.7 & 83.3 & $-0.33$ & [$-2.33$, $+1.67$] \\
\midrule
InternVL3-38B & ChartQA & 200 & 92.5 & 91.5 & $-1.00$ & [$-2.50$, $0.00$] \\
 & DocVQA & 200 & 96.4 & 95.7 & $-0.70$ & [$-1.93$, $+0.05$] \\
 & TextVQA & 200 & 79.8 & 82.4 & $+2.65$ & [$0.00$, $+5.15$] \\
\bottomrule
\end{tabular}
\end{table}

%% file: sections/appx/tab_app_fail_methods.tex
\begin{table*}[t]
\centering\scriptsize\setlength{\tabcolsep}{5pt}
\renewcommand{\arraystretch}{0.9}
\caption{\textbf{Pruning failures of each existing method.} InternVL3-38B and 8B on the 16 benchmarks of Figure~\ref{fig:finding}b. Fail.\ is the share of all questions that the full-token run answers correctly and the method answers wrongly at the budget. Right and Disagr.\ are the shares of these failures that the checker answers correctly and that it disagrees with, as means over the benchmarks with the two models pooled. AUROC is for these failures on the questions that the checker answers, by the confidence of the large VLM (Conf.) or by the disagreement of the checker and then the confidence (Ours), as the median over the 96 settings of each method.}
\label{tab:app-fail-methods}
\begin{tabular}{@{}lccccccccccc@{}}
\toprule
 & \multicolumn{3}{c}{Fail.} & \multicolumn{3}{c}{Right} & \multicolumn{3}{c}{Disagr.} & \multicolumn{2}{c}{AUROC} \\
\cmidrule(lr){2-4}\cmidrule(lr){5-7}\cmidrule(lr){8-10}\cmidrule(lr){11-12}
Method & 2.5\% & 5\% & 10\% & 2.5\% & 5\% & 10\% & 2.5\% & 5\% & 10\% & Conf. & Ours \\
\midrule
ToMe & 38.9 & 36.0 & 28.2 & 84.3 & 83.8 & 82.0 & 95.9 & 95.8 & 95.1 & 0.71 & 0.87 \\
VisionZip & 44.6 & 36.7 & 27.6 & 85.3 & 84.2 & 82.4 & 96.4 & 96.0 & 95.5 & 0.70 & 0.86 \\
ZOO-Prune & 35.4 & 29.8 & 23.3 & 83.2 & 81.9 & 79.9 & 95.6 & 95.2 & 94.3 & 0.74 & 0.87 \\
FastV & 36.6 & 28.6 & 20.4 & 84.2 & 82.5 & 79.9 & 96.0 & 95.4 & 94.7 & 0.72 & 0.89 \\
CDPruner & 34.7 & 29.3 & 23.3 & 83.5 & 82.2 & 80.4 & 95.6 & 95.2 & 94.6 & 0.72 & 0.88 \\
MMTok & 35.0 & 28.7 & 22.7 & 83.2 & 82.2 & 80.4 & 95.6 & 95.4 & 94.8 & 0.71 & 0.89 \\
SGL & 16.7 & 11.4 & 7.2 & 73.9 & 68.6 & 63.2 & 90.4 & 88.4 & 85.9 & 0.82 & 0.91 \\
IF-Prune & 13.2 & 9.4 & 6.4 & 70.5 & 66.4 & 62.2 & 89.5 & 87.2 & 85.1 & 0.79 & 0.90 \\
\bottomrule
\end{tabular}
\end{table*}

%% file: sections/appx/tab_app_complement.tex
\begin{table*}[t]
\centering\scriptsize\setlength{\tabcolsep}{12.33pt}
\caption{\textbf{Pruning failures of IF-Prune and the checker, per benchmark.} Fail.\ is the share of all questions that the full-token run answers correctly and IF-Prune answers wrongly. Right is the share of these failures that InternVL3-2B answers correctly from the full image. Figure~\ref{fig:finding}b averages these values over the benchmarks with the two models pooled.}
\label{tab:app-complement}
\begin{tabular}{@{}lcccccccccccc@{}}
\toprule
 & \multicolumn{6}{c}{InternVL3-38B} & \multicolumn{6}{c}{InternVL3-8B} \\
\cmidrule(lr){2-7}\cmidrule(lr){8-13}
 & \multicolumn{2}{c}{2.5\%} & \multicolumn{2}{c}{5\%} & \multicolumn{2}{c}{10\%} & \multicolumn{2}{c}{2.5\%} & \multicolumn{2}{c}{5\%} & \multicolumn{2}{c}{10\%} \\
\cmidrule(lr){2-3}\cmidrule(lr){4-5}\cmidrule(lr){6-7}\cmidrule(lr){8-9}\cmidrule(lr){10-11}\cmidrule(lr){12-13}
Benchmark & Fail. & Right & Fail. & Right & Fail. & Right & Fail. & Right & Fail. & Right & Fail. & Right \\
\midrule
\multicolumn{13}{@{}l}{\textit{Text-rich}} \\
ChartQA & 38.3 & 79.7 & 27.4 & 76.4 & 14.4 & 67.2 & 28.7 & 78.3 & 18.4 & 69.8 & 11.0 & 61.5 \\
DocVQA & 15.5 & 74.6 & 10.7 & 67.5 & 6.7 & 64.1 & 18.5 & 79.0 & 14.1 & 75.3 & 9.4 & 72.6 \\
InfoVQA & 26.1 & 61.6 & 20.2 & 56.8 & 14.8 & 52.0 & 28.0 & 69.3 & 21.2 & 66.5 & 16.2 & 61.4 \\
TextVQA & 11.2 & 66.8 & 7.5 & 60.9 & 4.6 & 61.8 & 11.1 & 71.6 & 7.3 & 65.3 & 5.0 & 63.3 \\
OCRB. & 21.8 & 76.1 & 12.9 & 69.0 & 6.5 & 55.4 & 20.5 & 76.6 & 11.9 & 69.7 & 7.0 & 67.1 \\
WebSRC & 22.2 & 74.9 & 14.5 & 67.7 & 8.8 & 60.5 & 24.5 & 78.9 & 17.5 & 75.3 & 12.7 & 68.5 \\
\midrule
\multicolumn{13}{@{}l}{\textit{General}} \\
GQA & 7.0 & 55.0 & 5.2 & 53.4 & 4.0 & 52.5 & 7.4 & 58.5 & 6.0 & 56.1 & 4.3 & 54.5 \\
VQAv2 & 9.6 & 71.4 & 7.8 & 71.1 & 5.3 & 68.7 & 8.0 & 66.6 & 6.0 & 63.7 & 4.4 & 61.9 \\
OK-VQA & 17.4 & 46.9 & 10.7 & 45.1 & 6.2 & 45.7 & 7.9 & 48.6 & 6.2 & 49.4 & 4.4 & 51.6 \\
MMB & 5.5 & 62.3 & 3.9 & 69.2 & 3.2 & 65.7 & 5.3 & 70.6 & 3.8 & 70.6 & 3.3 & 66.0 \\
MMB-CN & 5.9 & 52.0 & 4.5 & 52.3 & 3.5 & 56.9 & 6.4 & 62.0 & 5.2 & 56.6 & 4.4 & 62.1 \\
MME & 4.7 & 67.6 & 2.6 & 69.4 & 1.6 & 65.8 & 4.0 & 63.5 & 2.5 & 62.7 & 1.8 & 66.7 \\
POPE & 3.3 & 72.3 & 2.7 & 71.2 & 2.2 & 74.0 & 2.5 & 62.2 & 1.9 & 63.4 & 1.8 & 68.6 \\
SEED & 7.2 & 64.3 & 5.6 & 59.8 & 4.6 & 59.5 & 7.2 & 66.6 & 5.7 & 65.9 & 5.1 & 65.0 \\
AI2D & 14.8 & 60.4 & 11.6 & 56.3 & 8.6 & 50.6 & 13.9 & 63.8 & 11.0 & 60.0 & 8.7 & 56.1 \\
SQA-I & 9.6 & 89.1 & 7.2 & 84.9 & 4.7 & 85.3 & 8.7 & 88.1 & 6.2 & 90.4 & 4.7 & 86.3 \\
\bottomrule
\end{tabular}
\end{table*}

%% file: sections/appx/tab_app_auroc_wide.tex
\begin{table*}[t]
\centering\scriptsize\setlength{\tabcolsep}{2.2pt}
\caption{\textbf{Detecting pruning failures per benchmark.} AUROC of the eight signals of Figure~\ref{fig:finding}c for the first pass, averaged over the three budgets, on the questions that the checker answers. Ours ranks the questions by the disagreement of the checker and then by the confidence of the large VLM. Two prun.\ compares the answers of the large VLM under the two rankings of Appendix~\ref{app:ranking}, and MMB has no run with the rank average (n/a). The two models side by side. \textbf{Best} in bold, \protect\underline{second}{} underlined, per row and model.}
\label{tab:app-auroc}
\begin{tabular}{@{}lcccccccccccccccc@{}}
\toprule
 & \multicolumn{8}{c}{InternVL3-38B} & \multicolumn{8}{c}{InternVL3-8B} \\
\cmidrule(lr){2-9}\cmidrule(lr){10-17}
Benchmark & \shortstack{Rand.} & \shortstack{2B\\conf.} & \shortstack{SGL\\exit} & \shortstack{Two\\\vphantom{f}prun.} & \shortstack{1B\\dis.} & \shortstack{VLM\\conf.} & \shortstack{2B\\dis.} & \shortstack{Ours} & \shortstack{Rand.} & \shortstack{2B\\conf.} & \shortstack{SGL\\exit} & \shortstack{Two\\\vphantom{f}prun.} & \shortstack{1B\\dis.} & \shortstack{VLM\\conf.} & \shortstack{2B\\dis.} & \shortstack{Ours} \\
\midrule
\multicolumn{17}{@{}l}{\textit{Text-rich}} \\
ChartQA & 0.50 & 0.65 & 0.70 & 0.66 & 0.85 & 0.82 & \underline{0.86} & \textbf{0.90} & 0.49 & 0.71 & 0.78 & 0.73 & 0.85 & 0.83 & \underline{0.86} & \textbf{0.91} \\
DocVQA & 0.49 & 0.56 & 0.68 & 0.77 & 0.84 & 0.73 & \underline{0.88} & \textbf{0.91} & 0.50 & 0.55 & 0.67 & 0.74 & 0.85 & 0.71 & \underline{0.88} & \textbf{0.91} \\
InfoVQA & 0.51 & 0.57 & 0.63 & 0.68 & 0.69 & 0.70 & \underline{0.77} & \textbf{0.81} & 0.51 & 0.54 & 0.60 & 0.66 & 0.69 & 0.64 & \underline{0.76} & \textbf{0.79} \\
TextVQA & 0.49 & 0.65 & 0.69 & 0.72 & 0.80 & 0.74 & \underline{0.84} & \textbf{0.89} & 0.49 & 0.65 & 0.68 & 0.70 & 0.82 & 0.76 & \underline{0.85} & \textbf{0.89} \\
OCRB. & 0.54 & 0.47 & 0.66 & 0.77 & 0.80 & 0.69 & \underline{0.84} & \textbf{0.88} & 0.52 & 0.52 & 0.61 & 0.81 & 0.78 & 0.63 & \underline{0.83} & \textbf{0.86} \\
WebSRC & 0.50 & 0.58 & 0.75 & 0.81 & 0.82 & 0.80 & \underline{0.87} & \textbf{0.90} & 0.51 & 0.56 & 0.75 & 0.79 & 0.85 & 0.73 & \underline{0.89} & \textbf{0.92} \\
\multicolumn{17}{@{}l}{\textit{General}} \\
GQA & 0.48 & 0.60 & 0.59 & 0.67 & 0.72 & 0.73 & \underline{0.75} & \textbf{0.82} & 0.49 & 0.63 & 0.62 & 0.69 & 0.74 & \underline{0.76} & \underline{0.76} & \textbf{0.84} \\
VQAv2 & 0.50 & 0.61 & 0.62 & 0.67 & 0.78 & 0.60 & \underline{0.83} & \textbf{0.87} & 0.50 & 0.65 & 0.68 & 0.70 & 0.78 & 0.72 & \underline{0.84} & \textbf{0.88} \\
OK-VQA & 0.49 & 0.53 & 0.57 & 0.57 & 0.66 & 0.57 & \underline{0.72} & \textbf{0.73} & 0.51 & 0.55 & 0.55 & 0.64 & 0.66 & 0.63 & \underline{0.75} & \textbf{0.78} \\
MMB & 0.50 & 0.76 & 0.74 & n/a & 0.77 & \underline{0.91} & 0.85 & \textbf{0.94} & 0.50 & 0.76 & 0.72 & n/a & 0.77 & \underline{0.91} & 0.87 & \textbf{0.94} \\
MMB-CN & 0.49 & 0.71 & 0.69 & 0.66 & 0.74 & \underline{0.92} & 0.80 & \textbf{0.93} & 0.51 & 0.73 & 0.67 & 0.66 & 0.75 & \underline{0.90} & 0.82 & \textbf{0.93} \\
MME & 0.49 & 0.72 & 0.65 & 0.65 & 0.74 & \underline{0.88} & 0.79 & \textbf{0.91} & 0.50 & 0.71 & 0.71 & 0.62 & 0.71 & \underline{0.90} & 0.78 & \textbf{0.92} \\
POPE & 0.50 & 0.82 & 0.79 & 0.70 & 0.83 & \underline{0.93} & 0.85 & \textbf{0.96} & 0.49 & 0.82 & 0.80 & 0.71 & 0.82 & \underline{0.94} & 0.82 & \textbf{0.96} \\
SEED & 0.50 & 0.73 & 0.70 & 0.64 & 0.76 & \underline{0.84} & 0.81 & \textbf{0.90} & 0.51 & 0.74 & 0.71 & 0.65 & 0.76 & \underline{0.84} & 0.82 & \textbf{0.90} \\
AI2D & 0.51 & 0.72 & 0.75 & 0.70 & 0.76 & \underline{0.87} & 0.82 & \textbf{0.91} & 0.50 & 0.70 & 0.74 & 0.67 & 0.76 & \underline{0.85} & 0.83 & \textbf{0.91} \\
SQA-I & 0.48 & 0.74 & 0.82 & 0.68 & 0.91 & 0.92 & \underline{0.95} & \textbf{0.99} & 0.50 & 0.76 & 0.81 & 0.71 & 0.91 & 0.92 & \underline{0.95} & \textbf{0.98} \\
\bottomrule
\end{tabular}
\end{table*}

%% file: sections/appx/tab_app_signif.tex
\begin{table*}[t]
\centering\scriptsize\setlength{\tabcolsep}{6.53pt}
\caption{\textbf{Significance against IF-Prune and SGL.} Differences between PRUDENT (Ours) and IF-Prune or SGL in Tables~\ref{tab:main-38b} and~\ref{tab:main-8b}, in points of each benchmark at the same budget, with MME on its 2800-point scale. An asterisk marks a difference whose paired bootstrap 95\% interval over questions excludes zero. Table~\ref{tab:app-signif-counts} counts the significant settings against all eight existing methods.}
\label{tab:app-signif}
\begin{tabular}{@{}lcccccccccccc@{}}
\toprule
 & \multicolumn{6}{c}{InternVL3-38B} & \multicolumn{6}{c}{InternVL3-8B} \\
\cmidrule(lr){2-7}\cmidrule(lr){8-13}
 & \multicolumn{3}{c}{vs.\ IF-Prune} & \multicolumn{3}{c}{vs.\ SGL} & \multicolumn{3}{c}{vs.\ IF-Prune} & \multicolumn{3}{c}{vs.\ SGL} \\
\cmidrule(lr){2-4}\cmidrule(lr){5-7}\cmidrule(lr){8-10}\cmidrule(lr){11-13}
Benchmark & 2.5\% & 5\% & 10\% & 2.5\% & 5\% & 10\% & 2.5\% & 5\% & 10\% & 2.5\% & 5\% & 10\% \\
\midrule
\multicolumn{13}{@{}l}{\textit{Text-rich}} \\
ChartQA & $+19.6^{*}$ & $+14.8^{*}$ & $+7.3^{*}$ & $+11.0^{*}$ & $+9.9^{*}$ & $+9.6^{*}$ & $+16.1^{*}$ & $+8.4^{*}$ & $+5.6^{*}$ & $+14.8^{*}$ & $+9.0^{*}$ & $+8.0^{*}$ \\
DocVQA & $+8.4^{*}$ & $+5.8^{*}$ & $+4.0^{*}$ & $+4.0^{*}$ & $+3.5^{*}$ & $+3.3^{*}$ & $+10.3^{*}$ & $+8.0^{*}$ & $+5.6^{*}$ & $+2.5^{*}$ & $+2.1^{*}$ & $+2.6^{*}$ \\
InfoVQA & $+5.1^{*}$ & $+4.7^{*}$ & $+2.7^{*}$ & $+3.2^{*}$ & $+4.0^{*}$ & $+3.4^{*}$ & $+7.3^{*}$ & $+5.2^{*}$ & $+5.6^{*}$ & $+1.3$ & $+1.5^{*}$ & $+3.3^{*}$ \\
TextVQA & $+5.3^{*}$ & $+3.1^{*}$ & $+1.5^{*}$ & $+3.7^{*}$ & $+2.4^{*}$ & $+1.6^{*}$ & $+5.1^{*}$ & $+3.5^{*}$ & $+2.3^{*}$ & $+3.7^{*}$ & $+1.8^{*}$ & $+0.8^{*}$ \\
OCRB. & $+0.9$ & $+2.3^{*}$ & $+1.3$ & $+18.2^{*}$ & $+16.1^{*}$ & $+9.1^{*}$ & $+1.8$ & $+1.6$ & $+2.0^{*}$ & $+19.6^{*}$ & $+16.3^{*}$ & $+11.1^{*}$ \\
WebSRC & $+6.8^{*}$ & $+4.5^{*}$ & $+2.6^{*}$ & $+10.0^{*}$ & $+7.2^{*}$ & $+5.4^{*}$ & $+8.1^{*}$ & $+6.4^{*}$ & $+5.6^{*}$ & $+9.4^{*}$ & $+6.6^{*}$ & $+5.0^{*}$ \\
\midrule
\multicolumn{13}{@{}l}{\textit{General}} \\
GQA & $+1.9^{*}$ & $+1.3^{*}$ & $+0.6^{*}$ & $+5.0^{*}$ & $+3.0^{*}$ & $+0.9^{*}$ & $+2.2^{*}$ & $+1.6^{*}$ & $+0.4$ & $+5.7^{*}$ & $+2.7^{*}$ & $+1.3^{*}$ \\
VQAv2 & $+3.4^{*}$ & $+3.5^{*}$ & $+2.4^{*}$ & $+7.3^{*}$ & $+4.8^{*}$ & $+2.4^{*}$ & $+2.3^{*}$ & $+1.9^{*}$ & $+1.0^{*}$ & $+6.4^{*}$ & $+3.8^{*}$ & $+1.3^{*}$ \\
OK-VQA & $+2.8^{*}$ & $+1.7^{*}$ & $+0.9^{*}$ & $+4.8^{*}$ & $+2.2^{*}$ & $+1.2^{*}$ & $+0.2$ & $+0.9^{*}$ & $+0.7^{*}$ & $+8.1^{*}$ & $+3.7^{*}$ & $+1.3^{*}$ \\
MMB & $+1.3^{*}$ & $+0.5$ & $+0.9^{*}$ & $+9.5^{*}$ & $+4.2^{*}$ & $+1.2$ & $+1.5^{*}$ & $+0.3$ & $+1.5^{*}$ & $+10.6^{*}$ & $+4.6^{*}$ & $+1.8^{*}$ \\
MMB-CN & $+1.4^{*}$ & $+1.1^{*}$ & $+1.0$ & $+10.1^{*}$ & $+4.2^{*}$ & $+1.1$ & $+0.9$ & $+1.2^{*}$ & $+0.3$ & $+10.7^{*}$ & $+5.9^{*}$ & $+0.9$ \\
MME & $+90.3^{*}$ & $+57.1^{*}$ & $+18.5$ & $+494.5^{*}$ & $+366.2^{*}$ & $+276.0^{*}$ & $+61.7$ & $+24.4$ & $+35.0$ & $+500.4^{*}$ & $+309.9^{*}$ & $+217.7^{*}$ \\
POPE & $+0.4^{*}$ & $+0.4^{*}$ & $+0.1$ & $+6.9^{*}$ & $+2.4^{*}$ & $+0.6^{*}$ & $+0.5^{*}$ & $0.0$ & $0.0$ & $+7.0^{*}$ & $+2.3^{*}$ & $+0.4$ \\
SEED & $+0.9^{*}$ & $+0.6^{*}$ & $+0.6^{*}$ & $+6.3^{*}$ & $+3.4^{*}$ & $+1.3^{*}$ & $+1.3^{*}$ & $+0.5^{*}$ & $+0.7^{*}$ & $+5.6^{*}$ & $+3.0^{*}$ & $+1.0^{*}$ \\
AI2D & $+3.3^{*}$ & $+2.4^{*}$ & $+1.6^{*}$ & $+5.0^{*}$ & $+4.6^{*}$ & $+3.4^{*}$ & $+2.9^{*}$ & $+1.9^{*}$ & $+2.0^{*}$ & $+4.7^{*}$ & $+3.5^{*}$ & $+2.6^{*}$ \\
SQA-I & $+2.4^{*}$ & $+2.6^{*}$ & $+1.5^{*}$ & $+8.0^{*}$ & $+5.3^{*}$ & $+1.6^{*}$ & $+2.9^{*}$ & $+1.5^{*}$ & $+1.5^{*}$ & $+7.3^{*}$ & $+4.0^{*}$ & $+1.1^{*}$ \\
\midrule
\multicolumn{13}{@{}l}{\textit{Grounding}} \\
RefC-A & $+3.7^{*}$ & $+2.7^{*}$ & $+2.3^{*}$ & $+7.4^{*}$ & $+7.7^{*}$ & $+4.1^{*}$ & $+2.5^{*}$ & $+1.0^{*}$ & $+0.9^{*}$ & $+7.5^{*}$ & $+5.9^{*}$ & $+1.9^{*}$ \\
RefC-B & $+7.7^{*}$ & $+3.9^{*}$ & $+3.9^{*}$ & $+7.1^{*}$ & $+7.2^{*}$ & $+2.4^{*}$ & $+2.0^{*}$ & $+1.5^{*}$ & $+1.9^{*}$ & $+10.7^{*}$ & $+7.7^{*}$ & $+2.8^{*}$ \\
\bottomrule
\end{tabular}
\end{table*}

%% file: sections/appx/tab_app_signif_counts_wide.tex
\begin{table*}[t]
\centering\scriptsize\setlength{\tabcolsep}{7pt}
\caption{\textbf{Significant settings against each method.} Each value counts the settings of Tables~\ref{tab:main-38b} and~\ref{tab:main-8b} in which PRUDENT (Ours) is significantly better than the method at the same budget (paired bootstrap over questions, 95\%). A setting is one benchmark at one budget. PRUDENT is significantly worse in no setting. The last row applies a Holm correction over all 864 comparisons, with $p$ from the bootstrap standard error, and PRUDENT is still significantly worse in none.}
\label{tab:app-signif-counts}
\begin{tabular}{@{}lcccccccc@{}}
\toprule
Ours vs. & ToMe & VisionZip & ZOO-Prune & FastV & CDPruner & MMTok & SGL & IF-Prune \\
\midrule
InternVL3-38B (54 settings) & 54 & 54 & 54 & 54 & 54 & 53 & 52 & 48 \\
InternVL3-8B (54 settings) & 54 & 54 & 54 & 54 & 54 & 53 & 51 & 42 \\
Both (108 settings) & 108 & 108 & 108 & 108 & 108 & 106 & 103 & 90 \\
Both, Holm-corrected & 108 & 108 & 108 & 107 & 108 & 106 & 97 & 70 \\
\bottomrule
\end{tabular}
\end{table*}

%% file: sections/appx/tab_app_frontier.tex
\begin{table*}[t]
\centering\scriptsize\setlength{\tabcolsep}{14.58pt}
\caption{\textbf{Scores at 20\% and 40\%.} These are the budgets of Figure~\ref{fig:hero} beyond those of Tables~\ref{tab:main-38b} and~\ref{tab:main-8b}, on the eight benchmarks of Figure~\ref{fig:app-frontier}. Avg.\ is the mean ratio to the full-token score over the eight benchmarks. On GQA, InternVL3-38B is scored on the questions of Table~\ref{tab:app-phrase}. \textbf{Best} in bold, \protect\underline{second}{} underlined, per budget.}
\label{tab:app-frontier}
\begin{tabular}{@{}lccccccccc@{}}
\toprule
Method & TextVQA & ChartQA & DocVQA & InfoVQA & OCRB. & WebSRC & GQA & MMB & Avg. \\
\midrule
\multicolumn{10}{@{}l}{\textit{InternVL3-38B}} \\
\rowcolor{gray!12} Full tokens & 83.7 & 88.9 & 93.4 & 77.4 & 85.4 & 94.3 & 67.5 & 89.2 & 100.0 \\
\multicolumn{10}{@{}l}{\textit{Retain 20\% of the visual tokens}} \\
ToMe & 64.0 & 47.6 & 44.1 & 39.5 & 44.3 & 54.4 & 65.6 & 85.4 & 66.3 \\
VisionZip & 62.3 & 70.4 & 54.3 & 39.6 & 38.7 & 57.4 & 64.2 & 84.9 & 69.9 \\
ZOO-Prune & 57.1 & 72.0 & 52.3 & 38.3 & 39.6 & 64.7 & 65.3 & 85.5 & 70.3 \\
FastV & 77.4 & 57.6 & 52.5 & 41.0 & 59.4 & 61.9 & 64.2 & 87.0 & 74.3 \\
CDPruner & 62.9 & 61.2 & 47.4 & 39.5 & 55.7 & 61.8 & 63.2 & 85.1 & 70.7 \\
MMTok & 66.6 & 58.0 & 50.0 & 41.9 & 54.8 & 61.4 & 66.2 & 86.3 & 72.1 \\
SGL & \underline{82.7} & 81.3 & \underline{90.6} & 69.7 & \underline{82.3} & 86.6 & \underline{66.8} & \textbf{88.1} & 95.4 \\
IF-Prune & 82.5 & \underline{84.6} & 90.4 & \underline{70.9} & \textbf{84.6} & \underline{89.2} & \underline{66.8} & 87.0 & \underline{96.5} \\
\rowcolor{oursbg} \textbf{Ours} & \textbf{83.5} & \textbf{87.0} & \textbf{92.4} & \textbf{71.7} & \textbf{84.6} & \textbf{91.8} & \textbf{67.4} & \underline{87.9} & \textbf{98.0} \\
\multicolumn{10}{@{}l}{\textit{Retain 40\% of the visual tokens}} \\
ToMe & 77.6 & 76.2 & 75.6 & 55.1 & 69.3 & 78.1 & 66.5 & 88.1 & 86.5 \\
VisionZip & 77.1 & 85.1 & 80.7 & 55.6 & 66.6 & 80.7 & 66.7 & 87.9 & 88.4 \\
ZOO-Prune & 70.2 & 83.3 & 73.0 & 50.7 & 62.0 & 80.5 & 65.8 & 88.0 & 84.4 \\
FastV & 81.5 & 79.2 & 71.5 & 54.7 & 73.9 & 78.4 & 66.1 & 87.8 & 87.4 \\
CDPruner & 74.3 & 79.6 & 66.3 & 51.8 & 71.5 & 77.2 & 65.0 & 87.1 & 84.5 \\
MMTok & 74.5 & 79.2 & 69.1 & 54.0 & 69.3 & 77.6 & 66.9 & 87.2 & 85.3 \\
SGL & 83.0 & 87.3 & \underline{92.8} & \underline{74.5} & 84.8 & 90.6 & 67.1 & \underline{88.3} & \underline{98.4} \\
IF-Prune & \underline{83.2} & \underline{87.5} & 92.0 & 74.0 & \underline{85.2} & \underline{92.0} & \underline{67.2} & 87.9 & \underline{98.4} \\
\rowcolor{oursbg} \textbf{Ours} & \textbf{83.8} & \textbf{88.4} & \textbf{93.3} & \textbf{75.2} & \textbf{85.9} & \textbf{93.5} & \textbf{67.5} & \textbf{88.7} & \textbf{99.5} \\
\midrule
\multicolumn{10}{@{}l}{\textit{InternVL3-8B}} \\
\rowcolor{gray!12} Full tokens & 80.9 & 86.6 & 91.7 & 71.2 & 84.2 & 90.8 & 62.9 & 85.9 & 100.0 \\
\multicolumn{10}{@{}l}{\textit{Retain 20\% of the visual tokens}} \\
ToMe & 57.0 & 42.0 & 32.6 & 32.4 & 39.4 & 43.6 & 60.8 & 81.8 & 60.8 \\
VisionZip & 61.4 & 57.9 & 51.4 & 38.9 & 46.2 & 55.9 & 60.1 & 80.8 & 69.9 \\
ZOO-Prune & 55.9 & 68.0 & 49.8 & 35.2 & 44.5 & 59.2 & 59.6 & 81.4 & 69.9 \\
FastV & 77.6 & 70.4 & 56.2 & 36.4 & 62.2 & 59.7 & 61.5 & \underline{84.5} & 78.2 \\
CDPruner & 61.8 & 62.5 & 45.8 & 35.1 & 49.5 & 57.1 & 61.0 & 82.5 & 70.3 \\
MMTok & 58.7 & 53.8 & 44.0 & 36.8 & 43.6 & 50.4 & 62.3 & 82.6 & 67.1 \\
SGL & \underline{80.9} & 80.8 & \underline{89.1} & \underline{64.8} & 80.1 & 83.4 & 62.8 & \textbf{85.4} & 96.0 \\
IF-Prune & 80.2 & \underline{82.8} & 87.3 & 64.2 & \underline{81.6} & \underline{84.5} & \textbf{63.2} & \underline{84.5} & \underline{96.1} \\
\rowcolor{oursbg} \textbf{Ours} & \textbf{81.5} & \textbf{84.1} & \textbf{90.6} & \textbf{67.0} & \textbf{83.4} & \textbf{88.3} & \underline{63.1} & 84.2 & \textbf{98.2} \\
\multicolumn{10}{@{}l}{\textit{Retain 40\% of the visual tokens}} \\
ToMe & 72.5 & 73.2 & 63.8 & 44.7 & 58.8 & 69.9 & 62.4 & 83.8 & 81.3 \\
VisionZip & 77.3 & 80.2 & 78.2 & 53.4 & 73.4 & 78.7 & 62.1 & 84.1 & 89.9 \\
ZOO-Prune & 67.7 & 81.1 & 70.0 & 48.2 & 63.2 & 76.8 & 61.1 & 83.2 & 84.4 \\
FastV & 80.2 & 82.6 & 78.6 & 52.1 & 76.0 & 80.0 & 62.3 & \underline{85.9} & 91.3 \\
CDPruner & 72.5 & 76.8 & 66.4 & 45.4 & 65.0 & 71.9 & 62.0 & 84.5 & 83.4 \\
MMTok & 68.4 & 76.8 & 66.5 & 49.4 & 61.0 & 71.5 & 62.7 & 84.5 & 83.0 \\
SGL & \underline{81.3} & 84.6 & \underline{91.3} & \underline{69.2} & 82.5 & 87.2 & \underline{62.8} & \textbf{86.0} & \underline{98.6} \\
IF-Prune & 81.1 & \underline{84.7} & 89.5 & 67.5 & \underline{83.5} & \underline{87.8} & \textbf{63.1} & 84.4 & 98.1 \\
\rowcolor{oursbg} \textbf{Ours} & \textbf{81.6} & \textbf{85.6} & \textbf{91.5} & \textbf{70.1} & \textbf{83.6} & \textbf{90.2} & \textbf{63.1} & 85.4 & \textbf{99.5} \\
\bottomrule
\end{tabular}
\end{table*}

%% file: sections/appx/tab_app_more_bench.tex
\begin{table}[t]
\centering\scriptsize\setlength{\tabcolsep}{1.19pt}
\caption{\textbf{Four more benchmarks.} PRUDENT (Ours) and existing methods on InternVL3-38B and 8B at the budgets of the main tables. MMMU uses its single-image validation questions, and RWQA is RealWorldQA. DeepSeek-V4.1-Flash scores MM-Vet for every method, the full-token run included. \textbf{Best} in bold, \protect\underline{second}{} underlined, per model and budget.}
\label{tab:app-bench}
\begin{tabular}{@{}lcccccccc@{}}
\toprule
 & \multicolumn{4}{c}{InternVL3-38B} & \multicolumn{4}{c}{InternVL3-8B} \\
\cmidrule(lr){2-5}\cmidrule(lr){6-9}
Method & MMMU & MMStar & RWQA & MM-Vet & MMMU & MMStar & RWQA & MM-Vet \\
\midrule
\rowcolor{gray!12} Full tokens & 63.2 & 72.3 & 73.7 & 78.8 & 55.4 & 68.4 & 69.0 & 76.6 \\
\midrule
\multicolumn{9}{@{}l}{\textit{Retain 2.5\% of the visual tokens}} \\
ToMe & 54.1 & 42.5 & 51.1 & 30.9 & 51.3 & 38.6 & 46.7 & 27.8 \\
VisionZip & 53.3 & 37.5 & 46.4 & 24.2 & 50.3 & 34.0 & 41.6 & 20.9 \\
ZOO-Prune & 53.5 & 44.6 & 55.4 & 30.7 & 50.1 & 47.5 & 57.1 & 30.2 \\
FastV & 54.4 & 41.8 & 50.2 & 35.6 & 51.2 & 42.0 & 48.6 & 25.8 \\
CDPruner & 55.1 & 43.4 & 49.7 & 36.6 & 51.7 & 46.5 & 51.4 & 35.7 \\
MMTok & 54.1 & 45.9 & 52.8 & 33.8 & 50.9 & 43.5 & 51.2 & 31.0 \\
SGL & \underline{55.4} & 49.1 & 62.4 & 46.8 & 50.9 & 48.1 & 63.5 & 42.4 \\
IF-Prune & \textbf{57.6} & \underline{55.8} & \textbf{68.9} & \textbf{56.9} & \underline{53.1} & \underline{57.2} & \underline{65.8} & \underline{52.1} \\
\rowcolor{oursbg} \textbf{Ours} & \textbf{57.6} & \textbf{56.5} & \underline{68.4} & \underline{54.1} & \textbf{55.3} & \textbf{57.4} & \textbf{66.7} & \textbf{55.1} \\
\midrule
\multicolumn{9}{@{}l}{\textit{Retain 5\% of the visual tokens}} \\
ToMe & 53.4 & 43.5 & 53.6 & 38.5 & 51.2 & 41.1 & 47.3 & 31.2 \\
VisionZip & 53.7 & 44.9 & 52.4 & 33.4 & 51.5 & 43.1 & 46.8 & 33.1 \\
ZOO-Prune & 54.5 & 49.7 & 60.1 & 42.8 & 49.6 & 49.3 & 59.5 & 36.4 \\
FastV & 56.0 & 47.5 & 54.9 & 43.6 & 52.7 & 47.7 & 59.2 & 43.5 \\
CDPruner & 54.9 & 48.3 & 50.6 & 46.1 & 52.4 & 50.4 & 54.9 & 42.6 \\
MMTok & 55.7 & 50.1 & 58.8 & 44.7 & 51.5 & 46.7 & 56.7 & 39.2 \\
SGL & 58.5 & \underline{56.0} & 66.1 & 54.6 & 54.8 & 54.1 & 66.1 & 52.6 \\
IF-Prune & \underline{59.0} & \textbf{61.5} & \underline{70.5} & \underline{60.9} & \underline{56.0} & \underline{59.5} & \underline{66.5} & \textbf{61.9} \\
\rowcolor{oursbg} \textbf{Ours} & \textbf{59.6} & \textbf{61.5} & \textbf{71.1} & \textbf{61.8} & \textbf{57.0} & \textbf{60.8} & \textbf{68.4} & \underline{58.7} \\
\midrule
\multicolumn{9}{@{}l}{\textit{Retain 10\% of the visual tokens}} \\
ToMe & 55.6 & 50.5 & 59.3 & 46.3 & 52.7 & 46.7 & 57.5 & 40.6 \\
VisionZip & 55.3 & 49.6 & 62.1 & 40.5 & 52.3 & 50.2 & 51.8 & 41.3 \\
ZOO-Prune & 55.6 & 56.7 & 65.1 & 49.7 & 52.1 & 55.6 & 63.0 & 42.5 \\
FastV & 58.0 & 53.3 & 59.2 & 50.0 & 52.8 & 55.9 & 65.2 & 54.3 \\
CDPruner & 56.3 & 53.8 & 54.5 & 48.8 & 53.5 & 54.1 & 60.8 & 49.3 \\
MMTok & 56.8 & 55.1 & 62.5 & 51.3 & 52.3 & 54.4 & 61.6 & 47.6 \\
SGL & 58.8 & 60.3 & 70.1 & 65.9 & 54.6 & 60.3 & 68.1 & 63.1 \\
IF-Prune & \textbf{60.3} & \underline{64.7} & \underline{71.5} & \underline{68.2} & \underline{56.3} & \underline{61.5} & \textbf{69.2} & \textbf{68.6} \\
\rowcolor{oursbg} \textbf{Ours} & \underline{59.7} & \textbf{65.0} & \textbf{72.0} & \textbf{69.0} & \textbf{56.7} & \textbf{62.7} & \underline{68.5} & \underline{65.9} \\
\bottomrule
\end{tabular}
\end{table}

%% file: sections/appx/tab_app_lowres.tex
\begin{table}[t]
\centering\scriptsize\setlength{\tabcolsep}{2.05pt}
\caption{\textbf{Comparison with lowering the resolution.} Lower res.\ feeds only the 448-pixel thumbnail of 256 visual tokens per image, and Tokens gives its share of the visual tokens, at least 1.5 times the 10\% budget of PRUDENT (Ours). Text-rich benchmarks lose much of their score at the lower resolution, and most of the others stay close to full tokens with these extra tokens.}
\label{tab:app-lowres}
\begin{tabular}{@{}lccccccc@{}}
\toprule
 & & \multicolumn{3}{c}{InternVL3-8B} & \multicolumn{3}{c}{InternVL3-38B} \\
\cmidrule(lr){3-5}\cmidrule(lr){6-8}
Benchmark & Tokens (\%) & Full & Lower res. & Ours 10\% & Full & Lower res. & Ours 10\% \\
\midrule
\multicolumn{8}{@{}l}{\textit{Text-rich}} \\
ChartQA & 18 & 86.6 & 78.8 & 83.3 & 88.9 & 79.8 & 83.8 \\
DocVQA & 15 & 91.7 & 65.8 & 88.4 & 93.4 & 64.3 & 91.1 \\
InfoVQA & 20 & 71.2 & 40.0 & 64.3 & 77.4 & 46.0 & 68.2 \\
TextVQA & 16 & 80.9 & 70.9 & 81.2 & 83.7 & 72.7 & 82.9 \\
OCRBench & 26 & 84.2 & 76.1 & 81.8 & 85.4 & 73.1 & 82.5 \\
WebSRC & 24 & 90.8 & 78.6 & 84.9 & 94.3 & 81.8 & 87.8 \\
\midrule
\multicolumn{8}{@{}l}{\textit{General}} \\
GQA & 18 & 62.9 & 62.3 & 63.2 & 67.5 & 65.5 & 67.0 \\
VQAv2 & 20 & 79.7 & 78.3 & 78.9 & 83.4 & 75.5 & 81.2 \\
OK-VQA & 22 & 49.0 & 47.6 & 48.6 & 58.0 & 55.5 & 56.2 \\
MMB & 36 & 85.9 & 85.5 & 84.2 & 89.2 & 89.4 & 87.0 \\
MMB-CN & 36 & 84.4 & 84.0 & 81.2 & 88.8 & 88.3 & 86.4 \\
MME & 21 & 2384 & 2348 & 2403 & 2466 & 2479 & 2477 \\
POPE & 20 & 91.0 & 90.1 & 90.5 & 90.8 & 89.6 & 90.3 \\
SEED & 16 & 78.1 & 76.9 & 76.7 & 79.2 & 77.8 & 77.7 \\
AI2D & 37 & 84.1 & 83.7 & 80.4 & 87.8 & 86.9 & 83.9 \\
SQA-I & 34 & 98.4 & 97.9 & 95.5 & 99.4 & 99.1 & 96.5 \\
\midrule
\multicolumn{8}{@{}l}{\textit{Grounding}} \\
RefC-A & 21 & 94.3 & 93.8 & 92.9 & 94.6 & 94.4 & 89.7 \\
RefC-B & 24 & 86.5 & 87.0 & 85.4 & 88.5 & 89.2 & 82.3 \\
\bottomrule
\end{tabular}
\end{table}

%% file: sections/appx/tab_app_size_14b.tex
\begin{table*}[t]
\centering\scriptsize\setlength{\tabcolsep}{10.94pt}
\caption{\textbf{InternVL3-14B on nine benchmarks.} PRUDENT (Ours) and the existing methods keep the same budget, on the nine benchmarks of Figure~\ref{fig:analysis}a. Avg.\ is the mean ratio to the full-token score. \textbf{Best} in bold, \protect\underline{second}{} underlined, per budget. PRUDENT is significantly better than every existing method in most settings and significantly worse in none (paired bootstrap over questions, 95\%).}
\label{tab:app-size-14b}
\begin{tabular}{lcccccccccc}
\toprule
 & \multicolumn{6}{c}{Text-rich} & \multicolumn{3}{c}{General} & \\
\cmidrule(lr){2-7}\cmidrule(lr){8-10}
Method & ChartQA & DocVQA & InfoVQA & TextVQA & OCRB. & WebSRC & GQA & VQAv2 & MMB & Avg. \\
\midrule
\rowcolor{gray!12} Full tokens & 87.9 & 93.2 & 74.6 & 81.0 & 84.5 & 93.8 & 63.6 & 79.8 & 87.1 & 100.0 \\
\midrule
\multicolumn{11}{@{}l}{\textit{Retain 2.5\% of the visual tokens}} \\
ToMe & 16.2 & 11.2 & 26.4 & 17.7 & 4.0 & 19.9 & 49.1 & 57.3 & 59.2 & 36.7 \\
VisionZip & 15.6 & 11.0 & 26.9 & 11.6 & 3.0 & 19.0 & 42.4 & 47.3 & 41.5 & 30.8 \\
ZOO-Prune & 23.9 & 14.2 & 25.8 & 19.7 & 5.3 & 23.3 & 50.6 & 61.0 & 62.0 & 40.0 \\
FastV & 21.8 & 18.1 & 27.8 & 39.0 & 16.5 & 27.6 & 51.2 & 61.2 & 69.4 & 46.2 \\
CDPruner & 30.2 & 16.7 & 27.4 & 31.0 & 12.9 & 30.1 & 51.4 & 61.5 & 64.7 & 45.2 \\
MMTok & 22.2 & 14.9 & 27.8 & 31.2 & 13.8 & 21.3 & 55.5 & 66.6 & 67.9 & 45.0 \\
SGL & \underline{61.6} & \underline{81.3} & \underline{53.0} & \underline{73.7} & 43.9 & 68.2 & 57.3 & 70.0 & 68.7 & 77.9 \\
IF-Prune & 57.2 & 74.9 & 49.8 & 73.0 & \underline{63.2} & \underline{70.0} & \underline{61.8} & \underline{76.4} & \underline{80.3} & \underline{81.9} \\
\rowcolor{oursbg} \textbf{Ours} & \textbf{74.5} & \textbf{84.8} & \textbf{55.8} & \textbf{78.7} & \textbf{65.2} & \textbf{78.0} & \textbf{63.5} & \textbf{77.7} & \textbf{82.0} & \textbf{88.8} \\
\midrule
\multicolumn{11}{@{}l}{\textit{Retain 5\% of the visual tokens}} \\
ToMe & 19.8 & 13.5 & 27.2 & 21.9 & 4.6 & 22.5 & 52.4 & 60.8 & 63.2 & 40.1 \\
VisionZip & 23.2 & 16.5 & 28.6 & 18.1 & 5.4 & 22.3 & 50.9 & 58.4 & 62.0 & 39.9 \\
ZOO-Prune & 32.8 & 20.0 & 26.0 & 26.7 & 8.6 & 31.0 & 55.5 & 67.0 & 70.7 & 46.9 \\
FastV & 28.3 & 24.8 & 30.9 & 53.4 & 27.0 & 35.0 & 55.4 & 67.8 & 75.7 & 54.9 \\
CDPruner & 40.6 & 23.7 & 27.5 & 40.5 & 20.9 & 39.2 & 55.3 & 66.3 & 73.4 & 53.2 \\
MMTok & 29.2 & 22.2 & 31.2 & 44.5 & 25.8 & 32.3 & 59.8 & 71.5 & 76.2 & 54.4 \\
SGL & \underline{68.5} & \underline{84.2} & \underline{58.8} & 76.7 & 57.5 & 77.0 & 61.2 & 74.2 & 76.1 & 85.4 \\
IF-Prune & 66.5 & 81.7 & 56.0 & \underline{76.8} & \underline{71.8} & \underline{77.1} & \underline{63.3} & \underline{77.9} & \underline{81.6} & \underline{87.9} \\
\rowcolor{oursbg} \textbf{Ours} & \textbf{78.9} & \textbf{88.5} & \textbf{61.5} & \textbf{79.4} & \textbf{76.0} & \textbf{83.8} & \textbf{64.0} & \textbf{78.9} & \textbf{83.4} & \textbf{93.3} \\
\midrule
\multicolumn{11}{@{}l}{\textit{Retain 10\% of the visual tokens}} \\
ToMe & 24.3 & 19.5 & 30.7 & 38.1 & 16.7 & 30.7 & 57.1 & 69.4 & 73.8 & 50.1 \\
VisionZip & 39.2 & 29.1 & 32.4 & 30.9 & 10.1 & 31.5 & 57.8 & 69.4 & 75.3 & 51.9 \\
ZOO-Prune & 49.5 & 30.9 & 28.8 & 35.9 & 16.0 & 44.9 & 59.2 & 72.3 & 77.5 & 56.9 \\
FastV & 43.6 & 36.6 & 33.9 & 66.4 & 43.0 & 46.9 & 58.9 & 73.7 & 79.2 & 65.9 \\
CDPruner & 56.1 & 35.1 & 31.9 & 50.0 & 31.6 & 52.3 & 58.6 & 70.3 & 78.9 & 63.3 \\
MMTok & 42.5 & 32.7 & 34.6 & 56.2 & 38.6 & 43.1 & 62.1 & 75.1 & 80.3 & 63.9 \\
SGL & 74.1 & \underline{88.2} & \underline{63.3} & \underline{79.1} & 72.4 & 81.8 & 62.7 & 77.1 & 80.8 & 91.4 \\
IF-Prune & \underline{77.0} & 87.0 & \underline{63.3} & \underline{79.1} & \underline{80.1} & \underline{83.2} & \underline{63.8} & \underline{78.8} & \underline{83.2} & \underline{93.5} \\
\rowcolor{oursbg} \textbf{Ours} & \textbf{83.8} & \textbf{90.7} & \textbf{66.2} & \textbf{80.4} & \textbf{82.0} & \textbf{87.4} & \textbf{64.2} & \textbf{79.4} & \textbf{83.7} & \textbf{96.4} \\
\bottomrule
\end{tabular}
\end{table*}

%% file: sections/appx/tab_app_size_26b.tex
\begin{table*}[t]
\centering\scriptsize\setlength{\tabcolsep}{10.94pt}
\caption{\textbf{InternVL2.5-26B on nine benchmarks.} PRUDENT (Ours) and the existing methods keep the same budget, on the nine benchmarks of Figure~\ref{fig:analysis}a. Avg.\ is the mean ratio to the full-token score. \textbf{Best} in bold, \protect\underline{second}{} underlined, per budget. PRUDENT is significantly better than every existing method in most settings and significantly worse in none (paired bootstrap over questions, 95\%).}
\label{tab:app-size-26b}
\begin{tabular}{lcccccccccc}
\toprule
 & \multicolumn{6}{c}{Text-rich} & \multicolumn{3}{c}{General} & \\
\cmidrule(lr){2-7}\cmidrule(lr){8-10}
Method & ChartQA & DocVQA & InfoVQA & TextVQA & OCRB. & WebSRC & GQA & VQAv2 & MMB & Avg. \\
\midrule
\rowcolor{gray!12} Full tokens & 86.6 & 92.2 & 72.6 & 83.1 & 82.4 & 91.0 & 64.9 & 83.5 & 85.6 & 100.0 \\
\midrule
\multicolumn{11}{@{}l}{\textit{Retain 2.5\% of the visual tokens}} \\
ToMe & 17.0 & 15.1 & 28.3 & 23.2 & 8.1 & 26.0 & 49.1 & 59.0 & 60.3 & 39.8 \\
VisionZip & 17.1 & 11.4 & 27.0 & 14.7 & 2.6 & 18.4 & 44.1 & 52.8 & 45.1 & 32.7 \\
ZOO-Prune & 28.2 & 20.3 & 26.6 & 29.4 & 13.1 & 28.3 & 53.0 & 66.1 & 63.3 & 45.4 \\
FastV & 16.3 & 11.1 & 26.3 & 12.2 & 3.9 & 19.6 & 38.4 & 43.2 & 32.2 & 28.5 \\
CDPruner & 31.0 & 20.8 & 28.3 & 44.4 & 18.7 & 30.4 & 48.8 & 60.9 & 59.8 & 47.2 \\
MMTok & 20.8 & 14.6 & 26.9 & 25.9 & 10.7 & 22.5 & 53.6 & 65.7 & 66.3 & 42.7 \\
SGL & \underline{62.9} & \underline{81.4} & \underline{53.3} & \underline{77.2} & 49.6 & 66.4 & 56.4 & 70.5 & 69.0 & 79.1 \\
IF-Prune & 54.4 & 74.4 & 49.9 & 76.0 & \underline{64.8} & \underline{67.8} & \underline{60.2} & \underline{75.7} & \underline{78.0} & \underline{81.3} \\
\rowcolor{oursbg} \textbf{Ours} & \textbf{66.9} & \textbf{84.6} & \textbf{55.6} & \textbf{80.5} & \textbf{67.7} & \textbf{75.5} & \textbf{62.7} & \textbf{77.9} & \textbf{80.7} & \textbf{88.0} \\
\midrule
\multicolumn{11}{@{}l}{\textit{Retain 5\% of the visual tokens}} \\
ToMe & 21.9 & 18.5 & 29.4 & 30.1 & 10.7 & 28.6 & 53.6 & 65.8 & 67.3 & 45.2 \\
VisionZip & 22.1 & 14.7 & 27.8 & 22.3 & 10.0 & 22.3 & 52.2 & 63.4 & 64.1 & 41.6 \\
ZOO-Prune & 41.2 & 31.0 & 28.7 & 42.7 & 26.3 & 36.3 & 57.6 & 72.2 & 73.3 & 56.1 \\
FastV & 19.4 & 16.1 & 27.5 & 26.7 & 9.8 & 23.5 & 39.9 & 45.9 & 48.5 & 35.6 \\
CDPruner & 43.4 & 28.7 & 30.4 & 57.8 & 33.1 & 36.6 & 54.3 & 68.6 & 68.9 & 57.7 \\
MMTok & 25.1 & 20.6 & 29.3 & 37.9 & 21.2 & 28.1 & 59.1 & 72.0 & 74.6 & 50.9 \\
SGL & \underline{69.8} & \underline{84.8} & \underline{57.6} & \underline{79.8} & 61.8 & 74.5 & 61.5 & 76.0 & 75.5 & 86.5 \\
IF-Prune & 61.0 & 81.1 & 55.7 & 79.0 & \underline{72.5} & \underline{75.2} & \underline{62.4} & \underline{78.1} & \underline{80.8} & \underline{87.2} \\
\rowcolor{oursbg} \textbf{Ours} & \textbf{76.8} & \textbf{87.8} & \textbf{61.2} & \textbf{82.0} & \textbf{76.3} & \textbf{81.6} & \textbf{64.0} & \textbf{80.3} & \textbf{82.2} & \textbf{93.3} \\
\midrule
\multicolumn{11}{@{}l}{\textit{Retain 10\% of the visual tokens}} \\
ToMe & 29.7 & 28.3 & 31.8 & 48.4 & 29.1 & 37.1 & 58.7 & 73.9 & 77.9 & 57.0 \\
VisionZip & 32.8 & 25.2 & 30.3 & 35.5 & 17.7 & 31.0 & 57.3 & 71.8 & 75.3 & 51.9 \\
ZOO-Prune & 59.6 & 45.3 & 32.8 & 56.3 & 39.3 & 49.0 & 61.8 & 76.8 & 78.7 & 67.9 \\
FastV & 38.8 & 31.9 & 30.9 & 56.5 & 31.0 & 34.1 & 50.5 & 62.0 & 70.4 & 55.5 \\
CDPruner & 59.6 & 43.1 & 34.9 & 68.6 & 47.8 & 45.8 & 58.8 & 74.8 & 76.7 & 69.4 \\
MMTok & 33.2 & 30.7 & 33.5 & 51.8 & 29.4 & 36.8 & 62.1 & 76.0 & 78.8 & 59.4 \\
SGL & \underline{76.5} & \underline{87.8} & \underline{62.6} & \underline{81.8} & 70.8 & 79.3 & \underline{63.8} & 79.9 & \underline{82.6} & 92.4 \\
IF-Prune & 72.6 & 86.1 & 61.9 & 81.4 & \underline{78.8} & \underline{81.2} & 63.7 & \underline{80.4} & 82.2 & \underline{92.9} \\
\rowcolor{oursbg} \textbf{Ours} & \textbf{82.9} & \textbf{90.2} & \textbf{63.9} & \textbf{82.6} & \textbf{80.5} & \textbf{84.7} & \textbf{64.9} & \textbf{81.8} & \textbf{83.5} & \textbf{96.4} \\
\bottomrule
\end{tabular}
\end{table*}

%% file: sections/appx/tab_app_size_78b.tex
\begin{table*}[t]
\centering\scriptsize\setlength{\tabcolsep}{10.94pt}
\caption{\textbf{InternVL3-78B on nine benchmarks.} PRUDENT (Ours) and the existing methods keep the same budget, on the nine benchmarks of Figure~\ref{fig:analysis}a. Avg.\ is the mean ratio to the full-token score. \textbf{Best} in bold, \protect\underline{second}{} underlined, per budget. PRUDENT is significantly better than every existing method in most settings and significantly worse in none (paired bootstrap over questions, 95\%).}
\label{tab:app-size-78b}
\begin{tabular}{lcccccccccc}
\toprule
 & \multicolumn{6}{c}{Text-rich} & \multicolumn{3}{c}{General} & \\
\cmidrule(lr){2-7}\cmidrule(lr){8-10}
Method & ChartQA & DocVQA & InfoVQA & TextVQA & OCRB. & WebSRC & GQA & VQAv2 & MMB & Avg. \\
\midrule
\rowcolor{gray!12} Full tokens & 89.2 & 94.0 & 79.6 & 85.4 & 86.5 & 93.8 & 65.7 & 83.6 & 89.6 & 100.0 \\
\midrule
\multicolumn{11}{@{}l}{\textit{Retain 2.5\% of the visual tokens}} \\
ToMe & 22.6 & 16.8 & 31.5 & 28.0 & 10.1 & 28.1 & 54.8 & 65.2 & 72.2 & 44.3 \\
VisionZip & 20.2 & 13.4 & 29.5 & 17.9 & 4.8 & 21.8 & 48.1 & 55.2 & 54.3 & 36.0 \\
ZOO-Prune & 24.6 & 18.3 & 28.7 & 26.8 & 10.0 & 28.6 & 55.0 & 68.0 & 71.0 & 44.5 \\
FastV & 18.2 & 12.8 & 30.5 & 31.8 & 16.0 & 24.7 & 45.1 & 51.1 & 52.2 & 38.1 \\
CDPruner & 38.0 & 23.2 & 30.4 & 51.1 & 29.4 & 31.2 & 52.1 & 65.2 & 71.6 & 52.2 \\
MMTok & 23.8 & 16.1 & 29.0 & 32.1 & 15.1 & 25.8 & 58.1 & 69.3 & 74.1 & 46.3 \\
SGL & 64.0 & \underline{83.0} & \underline{57.0} & \underline{79.4} & 53.3 & 69.5 & 59.2 & 73.3 & 75.3 & 80.2 \\
IF-Prune & \underline{65.6} & 78.0 & 54.2 & 78.0 & \underline{67.3} & \underline{71.1} & \underline{63.2} & \underline{77.6} & \underline{84.2} & \underline{83.6} \\
\rowcolor{oursbg} \textbf{Ours} & \textbf{77.6} & \textbf{86.7} & \textbf{60.4} & \textbf{82.7} & \textbf{72.6} & \textbf{78.6} & \textbf{64.6} & \textbf{80.1} & \textbf{85.1} & \textbf{89.9} \\
\midrule
\multicolumn{11}{@{}l}{\textit{Retain 5\% of the visual tokens}} \\
ToMe & 24.6 & 20.1 & 32.4 & 34.2 & 13.8 & 31.2 & 57.1 & 68.7 & 74.9 & 47.9 \\
VisionZip & 26.4 & 18.7 & 30.9 & 26.7 & 12.3 & 25.8 & 55.5 & 65.7 & 72.0 & 45.0 \\
ZOO-Prune & 38.8 & 27.0 & 31.7 & 38.0 & 21.4 & 38.6 & 59.3 & 72.5 & 78.9 & 54.1 \\
FastV & 21.2 & 18.4 & 31.4 & 52.6 & 27.9 & 30.1 & 51.3 & 63.1 & 68.0 & 48.7 \\
CDPruner & 51.7 & 33.6 & 33.3 & 63.2 & 41.2 & 39.5 & 56.0 & 70.6 & 79.3 & 62.0 \\
MMTok & 25.7 & 20.0 & 32.9 & 38.9 & 21.8 & 30.3 & 61.6 & 73.8 & 80.8 & 51.9 \\
SGL & 71.0 & \underline{86.0} & \underline{62.1} & \underline{82.0} & 65.0 & 77.4 & 62.6 & 77.4 & 81.6 & 86.9 \\
IF-Prune & \underline{75.3} & 84.1 & \underline{62.1} & \underline{82.0} & \underline{76.2} & \underline{80.0} & \underline{64.5} & \underline{79.7} & \underline{85.6} & \underline{90.0} \\
\rowcolor{oursbg} \textbf{Ours} & \textbf{82.5} & \textbf{89.7} & \textbf{65.6} & \textbf{84.1} & \textbf{79.8} & \textbf{85.2} & \textbf{65.4} & \textbf{81.4} & \textbf{87.7} & \textbf{94.1} \\
\midrule
\multicolumn{11}{@{}l}{\textit{Retain 10\% of the visual tokens}} \\
ToMe & 33.6 & 29.6 & 34.9 & 51.2 & 30.8 & 40.7 & 61.2 & 74.7 & 82.3 & 58.5 \\
VisionZip & 39.9 & 29.1 & 33.7 & 40.3 & 21.3 & 34.5 & 59.7 & 73.5 & 79.6 & 54.9 \\
ZOO-Prune & 59.3 & 41.7 & 34.0 & 50.8 & 39.3 & 52.9 & 62.4 & 75.7 & 83.7 & 66.0 \\
FastV & 37.1 & 30.1 & 34.8 & 68.1 & 44.5 & 40.3 & 58.0 & 72.2 & 80.0 & 61.7 \\
CDPruner & 69.4 & 47.9 & 38.4 & 73.4 & 54.8 & 52.0 & 60.0 & 75.7 & 84.8 & 73.1 \\
MMTok & 32.4 & 27.5 & 35.8 & 49.4 & 31.7 & 38.4 & 63.9 & 76.9 & 83.2 & 58.7 \\
SGL & 77.1 & \underline{89.0} & 67.4 & \underline{83.5} & 76.5 & 82.7 & 64.5 & 80.3 & 87.0 & 92.4 \\
IF-Prune & \underline{83.6} & 88.2 & \underline{67.8} & \underline{83.5} & \underline{83.1} & \underline{86.1} & \underline{65.4} & \underline{81.5} & \textbf{87.7} & \underline{94.8} \\
\rowcolor{oursbg} \textbf{Ours} & \textbf{86.0} & \textbf{91.7} & \textbf{71.3} & \textbf{84.5} & \textbf{85.1} & \textbf{88.7} & \textbf{65.7} & \textbf{82.2} & \underline{87.5} & \textbf{96.8} \\
\bottomrule
\end{tabular}
\end{table*}

%% file: sections/appx/tab_app_family_q8b.tex
\begin{table*}[t]
\centering\scriptsize\setlength{\tabcolsep}{9.25pt}
\caption{\textbf{Qwen3-VL-8B on 10 benchmarks.} PRUDENT (Ours) and existing pruning methods at 10\% of the visual tokens. Table~\ref{tab:family} shows five of the benchmarks. Avg.\ is the mean ratio to the full-token score. \textbf{Best} in bold, \protect\underline{second}{} underlined.}
\label{tab:app-family-q8b}
\begin{tabular}{lccccccccccc}
\toprule
 & \multicolumn{5}{c}{Text-rich} & \multicolumn{5}{c}{General} & \\
\cmidrule(lr){2-6}\cmidrule(lr){7-11}
Method & ChartQA & DocVQA & InfoVQA & TextVQA & OCRB. & GQA & VQAv2 & OK-VQA & MMB & POPE & Avg. \\
\midrule
\rowcolor{gray!12} Full tokens & 85.2 & 95.5 & 82.0 & 82.1 & 85.5 & 61.5 & 80.7 & 50.7 & 84.6 & 90.8 & 100.0 \\
\midrule
\multicolumn{12}{@{}l}{\textit{Retain 10\% of the visual tokens}} \\
ToMe & 25.2 & 28.3 & 28.8 & 36.2 & 27.7 & 51.5 & 66.0 & 34.8 & 64.9 & 81.9 & 57.2 \\
VisionZip & 35.1 & 44.1 & 32.3 & 49.1 & 19.3 & 49.5 & 63.6 & 34.1 & 65.6 & 81.1 & 60.3 \\
ZOO-Prune & 37.9 & 45.5 & 31.0 & 44.9 & 34.3 & 54.1 & 70.1 & 37.3 & 68.8 & 88.7 & 65.2 \\
FastV & 23.1 & 29.0 & 27.6 & 35.5 & 7.1 & 44.7 & 55.1 & 23.0 & 57.8 & 76.3 & 48.2 \\
CDPruner & 31.4 & 40.5 & 32.6 & 51.1 & \underline{40.7} & 50.7 & 68.5 & 39.2 & 70.3 & 85.2 & 65.0 \\
MMTok & 43.5 & 59.7 & 39.8 & 60.8 & 30.9 & 55.4 & 69.7 & 36.7 & 68.7 & 88.9 & 70.0 \\
SGL & \underline{74.9} & \underline{93.0} & \underline{73.8} & \underline{80.0} & \textbf{51.1} & \underline{59.7} & \underline{77.1} & \textbf{46.1} & \underline{73.5} & \textbf{90.7} & \underline{90.3} \\
\rowcolor{oursbg} \textbf{Ours} & \textbf{77.7} & \textbf{94.1} & \textbf{75.5} & \textbf{81.3} & \textbf{51.1} & \textbf{60.9} & \textbf{78.4} & \underline{43.0} & \textbf{78.9} & \underline{90.0} & \textbf{91.4} \\
\bottomrule
\end{tabular}
\end{table*}

%% file: sections/appx/tab_app_family_q32b.tex
\begin{table*}[t]
\centering\scriptsize\setlength{\tabcolsep}{9.25pt}
\caption{\textbf{Qwen3-VL-32B on 10 benchmarks.} PRUDENT (Ours) and existing pruning methods at 10\% of the visual tokens. Table~\ref{tab:family} shows five of the benchmarks. Avg.\ is the mean ratio to the full-token score. \textbf{Best} in bold, \protect\underline{second}{} underlined.}
\label{tab:app-family-q32b}
\begin{tabular}{lccccccccccc}
\toprule
 & \multicolumn{5}{c}{Text-rich} & \multicolumn{5}{c}{General} & \\
\cmidrule(lr){2-6}\cmidrule(lr){7-11}
Method & ChartQA & DocVQA & InfoVQA & TextVQA & OCRB. & GQA & VQAv2 & OK-VQA & MMB & POPE & Avg. \\
\midrule
\rowcolor{gray!12} Full tokens & 87.3 & 96.2 & 86.4 & 83.5 & 86.3 & 62.5 & 82.4 & 51.0 & 88.1 & 91.1 & 100.0 \\
\midrule
\multicolumn{12}{@{}l}{\textit{Retain 10\% of the visual tokens}} \\
ToMe & 21.2 & 30.2 & 29.5 & 39.5 & 28.2 & 54.4 & 69.2 & 40.8 & 73.7 & 82.6 & 59.5 \\
VisionZip & 33.1 & 40.1 & 31.7 & 50.5 & 24.9 & 53.2 & 68.9 & 39.3 & 77.1 & 83.8 & 63.1 \\
ZOO-Prune & 35.2 & 55.8 & 34.3 & 56.7 & 38.6 & 56.1 & 72.0 & \underline{43.3} & 77.1 & 86.1 & 69.5 \\
FastV & 22.4 & 33.4 & 30.5 & 60.3 & 34.3 & 46.7 & 63.4 & 33.7 & 71.0 & 72.5 & 58.5 \\
CDPruner & 22.2 & 34.8 & 28.0 & 47.2 & 28.7 & 49.7 & 62.7 & 35.2 & 74.7 & 78.4 & 57.9 \\
MMTok & 41.0 & 60.9 & 42.0 & 62.7 & 38.3 & 57.5 & 73.4 & 42.6 & 77.7 & 87.3 & 72.7 \\
SGL & \underline{60.0} & \underline{93.2} & \underline{76.1} & \underline{80.8} & \underline{52.4} & \underline{59.9} & \underline{77.5} & \textbf{45.7} & \underline{78.5} & \underline{89.8} & \underline{87.8} \\
\rowcolor{oursbg} \textbf{Ours} & \textbf{62.0} & \textbf{94.5} & \textbf{78.2} & \textbf{82.2} & \textbf{53.2} & \textbf{61.0} & \textbf{79.2} & 42.7 & \textbf{82.0} & \textbf{91.0} & \textbf{89.0} \\
\bottomrule
\end{tabular}
\end{table*}

%% file: sections/appx/tab_app_family_g31b.tex
\begin{table*}[t]
\centering\scriptsize\setlength{\tabcolsep}{9.25pt}
\caption{\textbf{Gemma 4 31B on 10 benchmarks.} PRUDENT (Ours) and existing pruning methods at 20\% of the visual tokens. Table~\ref{tab:family} shows five of the benchmarks. Avg.\ is the mean ratio to the full-token score. \textbf{Best} in bold, \protect\underline{second}{} underlined. ZOO-Prune is not defined on Gemma 4 (n/a).}
\label{tab:app-family-g31b}
\begin{tabular}{lccccccccccc}
\toprule
 & \multicolumn{5}{c}{Text-rich} & \multicolumn{5}{c}{General} & \\
\cmidrule(lr){2-6}\cmidrule(lr){7-11}
Method & ChartQA & DocVQA & InfoVQA & TextVQA & OCRB. & GQA & VQAv2 & OK-VQA & MMB & POPE & Avg. \\
\midrule
\rowcolor{gray!12} Full tokens & 68.0 & 93.0 & 78.4 & 75.8 & 83.9 & 47.3 & 64.5 & 38.0 & 84.1 & 91.0 & 100.0 \\
\midrule
\multicolumn{12}{@{}l}{\textit{Retain 20\% of the visual tokens}} \\
ToMe & 39.5 & 40.3 & 25.5 & 60.8 & 50.5 & 44.7 & \underline{62.2} & \underline{37.8} & 82.7 & 88.8 & 76.1 \\
VisionZip & \textbf{65.4} & 60.0 & 44.4 & 62.6 & 67.2 & 44.4 & 62.0 & 37.5 & 83.6 & 89.5 & 86.6 \\
ZOO-Prune & n/a & n/a & n/a & n/a & n/a & n/a & n/a & n/a & n/a & n/a & n/a \\
FastV & 36.6 & 23.1 & 22.4 & 49.6 & 44.1 & 40.8 & 55.2 & 32.2 & 77.3 & 84.4 & 66.6 \\
CDPruner & 63.5 & 61.4 & 35.7 & 68.2 & 64.9 & 44.9 & 61.8 & 37.3 & \underline{83.7} & 89.4 & 85.9 \\
MMTok & \underline{65.3} & 73.4 & 47.7 & \underline{74.0} & \textbf{73.8} & \textbf{45.1} & \textbf{62.7} & \textbf{38.1} & \textbf{84.6} & \textbf{90.9} & \underline{91.4} \\
SGL & 52.0 & \underline{85.5} & \underline{66.4} & 73.5 & \underline{67.9} & 44.8 & 61.4 & 36.4 & 80.1 & \underline{90.6} & 91.1 \\
\rowcolor{oursbg} \textbf{Ours} & 56.0 & \textbf{89.4} & \textbf{68.6} & \textbf{75.3} & 67.0 & \underline{45.0} & 61.7 & 34.9 & 82.4 & 90.3 & \textbf{92.5} \\
\bottomrule
\end{tabular}
\end{table*}

%% file: sections/appx/tab_app_plugin.tex
\begin{table}[t]
\centering\scriptsize\setlength{\tabcolsep}{4.0pt}
\caption{\textbf{IF-Prune and SGL as the first pass, per benchmark.} Alone is the method itself at the budget $K$, and First uses its ranking in both passes of PRUDENT (Ours) at the same average budget. Figure~\ref{fig:analysis}b averages the columns of 10\% and 20\%. An asterisk marks a difference from Alone whose paired bootstrap 95\% interval excludes zero.}
\label{tab:app-plugin}
\begin{tabular}{@{}lcccccccccccc@{}}
\toprule
 & \multicolumn{2}{c}{IF-Prune, 5\%} & \multicolumn{2}{c}{IF-Prune, 10\%} & \multicolumn{2}{c}{IF-Prune, 20\%} & \multicolumn{2}{c}{SGL, 5\%} & \multicolumn{2}{c}{SGL, 10\%} & \multicolumn{2}{c}{SGL, 20\%} \\
\cmidrule(lr){2-3}\cmidrule(lr){4-5}\cmidrule(lr){6-7}\cmidrule(lr){8-9}\cmidrule(lr){10-11}\cmidrule(lr){12-13}
Benchmark & Alone & First & Alone & First & Alone & First & Alone & First & Alone & First & Alone & First \\
\midrule
\multicolumn{13}{@{}l}{\textit{InternVL3-38B}} \\
ChartQA & 63.4 & 63.5 & 76.5 & 76.7 & 84.6 & 86.5$^*$ & 68.3 & 72.2$^*$ & 74.2 & 79.4$^*$ & 81.3 & 84.9$^*$ \\
DocVQA & 82.7 & 84.3$^*$ & 87.1 & 89.7$^*$ & 90.4 & 91.7$^*$ & 85.0 & 87.4$^*$ & 87.9 & 90.3$^*$ & 90.6 & 92.0$^*$ \\
InfoVQA & 59.2 & 58.3 & 65.5 & 66.2 & 70.9 & 71.1 & 60.0 & 60.8 & 64.9 & 65.1 & 69.7 & 70.3 \\
TextVQA & 79.0 & 81.4$^*$ & 81.5 & 82.7$^*$ & 82.5 & 83.2$^*$ & 79.7 & 81.4$^*$ & 81.4 & 82.8$^*$ & 82.7 & 83.2$^*$ \\
WebSRC & 79.5 & 81.8$^*$ & 85.2 & 87.8$^*$ & 89.2 & 90.8$^*$ & 76.8 & 78.8$^*$ & 82.4 & 84.6$^*$ & 86.6 & 89.3$^*$ \\
GQA & 65.5 & 66.6$^*$ & 66.4 & 66.8$^*$ & 66.8 & 67.2$^*$ & 63.7 & 65.4$^*$ & 66.1 & 66.9$^*$ & 66.8 & 67.0 \\
MMB & 85.6 & 86.1 & 86.2 & 87.0$^*$ & 87.0 & 87.9 & 81.9 & 85.7$^*$ & 85.8 & 87.7$^*$ & 88.1 & 88.3 \\
\midrule
\multicolumn{13}{@{}l}{\textit{InternVL3-8B}} \\
ChartQA & 70.2 & 70.7 & 77.6 & 80.2$^*$ & 82.8 & 84.1$^*$ & 69.5 & 72.0$^*$ & 75.3 & 79.3$^*$ & 80.8 & 83.4$^*$ \\
DocVQA & 77.5 & 80.6$^*$ & 82.8 & 85.5$^*$ & 87.3 & 89.0$^*$ & 83.4 & 85.8$^*$ & 85.9 & 88.4$^*$ & 89.1 & 90.7$^*$ \\
InfoVQA & 53.2 & 52.3 & 58.7 & 58.4 & 64.2 & 64.5 & 56.9 & 57.1 & 61.0 & 62.3$^*$ & 64.8 & 65.5 \\
TextVQA & 76.5 & 78.8$^*$ & 79.0 & 80.3$^*$ & 80.2 & 81.3$^*$ & 78.1 & 80.3$^*$ & 80.4 & 81.2$^*$ & 80.9 & 81.8$^*$ \\
WebSRC & 74.9 & 76.7$^*$ & 79.3 & 82.9$^*$ & 84.5 & 86.7$^*$ & 74.7 & 76.2$^*$ & 79.9 & 82.0$^*$ & 83.4 & 86.3$^*$ \\
GQA & 61.4 & 62.4$^*$ & 62.8 & 63.1 & 63.2 & 63.2 & 60.3 & 61.2$^*$ & 61.9 & 62.9$^*$ & 62.8 & 62.9 \\
MMB & 82.6 & 83.0 & 82.6 & 84.2$^*$ & 84.5 & 84.2 & 78.4 & 81.8$^*$ & 82.4 & 85.1$^*$ & 85.4 & 86.2 \\
\bottomrule
\end{tabular}
\end{table}

%% file: sections/appx/tab_app_checker_means.tex
\begin{table}[t]
\centering\scriptsize\setlength{\tabcolsep}{5.77pt}
\renewcommand{\arraystretch}{0.9}
\caption{\textbf{Other checkers per large VLM.} The guide InternVL2.5-1B, InternVL3-2B (the default) or InternVL3-8B as the checker, with IF-Prune for reference. Mean ratio to the full-token score. \textbf{Best} in bold, \protect\underline{second}{} underlined, per model.}
\label{tab:app-checker}
\begin{tabular}{lcccccc}
\toprule
 & \multicolumn{3}{c}{Text-rich} & \multicolumn{3}{c}{General} \\
\cmidrule(lr){2-4}\cmidrule(lr){5-7}
Method & 2.5\% & 5\% & 10\% & 2.5\% & 5\% & 10\% \\
\midrule
\multicolumn{7}{l}{\textit{InternVL3-38B}} \\
IF-Prune & 75.4 & 83.7 & 91.1 & 91.2 & 94.2 & 96.6 \\
Ours, 1B checker & \textbf{84.3} & 89.8 & 94.5 & 93.1 & 95.9 & 97.5 \\
\rowcolor{oursbg} Ours, 2B checker (default) & \underline{84.1} & \underline{90.4} & \underline{94.8} & \underline{93.8} & \underline{96.2} & \underline{97.9} \\
Ours, 8B checker & \underline{84.1} & \textbf{90.9} & \textbf{95.4} & \textbf{94.1} & \textbf{96.3} & \textbf{98.0} \\
\midrule
\multicolumn{7}{l}{\textit{InternVL3-8B}} \\
IF-Prune & 76.1 & 84.3 & 90.4 & 93.9 & 95.9 & 97.2 \\
Ours, 1B checker & \underline{85.3} & \underline{90.7} & \underline{94.8} & \underline{95.6} & \underline{96.9} & \underline{98.2} \\
\rowcolor{oursbg} Ours, 2B checker (default) & \textbf{85.6} & \textbf{90.8} & \textbf{95.7} & \textbf{96.0} & \textbf{97.3} & \textbf{98.4} \\
\bottomrule
\end{tabular}
\end{table}

%% file: sections/appx/tab_app_checker_38b.tex
\begin{table*}[t]
\centering\scriptsize\setlength{\tabcolsep}{7.35pt}
\caption{\textbf{Other checkers on InternVL3-38B per benchmark.} PRUDENT (Ours) with the guide InternVL2.5-1B, InternVL3-2B (the default) or InternVL3-8B as the checker, against IF-Prune, on the benchmarks of Table~\ref{tab:app-checker}. Full is the full-token score, and MME is out of 2800. The last two rows are the mean ratio to the full-token score, as in Table~\ref{tab:app-checker}. \textbf{Best} in bold, \protect\underline{second}{} underlined, per budget. GQA and VQAv2 omit the questions where full tokens give a stock phrase (Appendix~\ref{app:setup}).}
\label{tab:app-checker-38b}
\begin{tabular}{@{}lccccccccccccc@{}}
\toprule
 & & \multicolumn{4}{c}{Retain 2.5\%} & \multicolumn{4}{c}{Retain 5\%} & \multicolumn{4}{c}{Retain 10\%} \\
\cmidrule(lr){3-6}\cmidrule(lr){7-10}\cmidrule(lr){11-14}
Benchmark & Full & IF-Prune & 1B & 2B & 8B & IF-Prune & 1B & 2B & 8B & IF-Prune & 1B & 2B & 8B \\
\midrule
\multicolumn{14}{@{}l}{\textit{Text-rich}} \\
ChartQA & 88.9 & 52.2 & \textbf{73.6} & 71.9 & \underline{72.0} & 63.4 & 77.9 & \textbf{78.2} & \underline{78.1} & 76.5 & 83.6 & \underline{83.8} & \textbf{85.2} \\
DocVQA & 93.4 & 77.2 & \underline{84.8} & \textbf{85.5} & \textbf{85.5} & 82.7 & 88.0 & \underline{88.4} & \textbf{89.4} & 87.1 & 90.6 & \underline{91.1} & \textbf{91.4} \\
InfoVQA & 77.4 & 52.5 & \textbf{57.9} & \underline{57.6} & \underline{57.6} & 59.2 & \underline{62.2} & \textbf{63.9} & \textbf{63.9} & 65.5 & 68.1 & \underline{68.2} & \textbf{68.4} \\
TextVQA & 83.7 & 75.6 & 80.3 & \underline{80.8} & \textbf{81.1} & 79.0 & 82.0 & \underline{82.1} & \textbf{82.7} & 81.5 & 82.8 & \underline{82.9} & \textbf{83.1} \\
OCRB. & 85.4 & 65.4 & \textbf{67.2} & \underline{66.3} & \underline{66.3} & 74.5 & 76.7 & \underline{76.8} & \textbf{77.6} & 81.2 & \underline{82.9} & 82.5 & \textbf{83.2} \\
WebSRC & 94.3 & 71.7 & 78.2 & \underline{78.5} & \textbf{78.6} & 79.5 & 83.7 & \underline{84.0} & \textbf{84.7} & 85.2 & 86.9 & \underline{87.8} & \textbf{88.4} \\
\midrule
\multicolumn{14}{@{}l}{\textit{General}} \\
GQA & 67.5 & 63.9 & 65.8 & \underline{65.9} & \textbf{66.3} & 65.5 & 66.7 & \underline{66.8} & \textbf{66.9} & 66.4 & \underline{67.0} & \underline{67.0} & \textbf{67.2} \\
VQAv2 & 83.4 & 74.2 & 76.8 & \underline{77.7} & \textbf{78.0} & 76.1 & 78.9 & \underline{79.6} & \textbf{79.9} & 78.8 & 80.7 & \underline{81.2} & \textbf{81.7} \\
OK-VQA & 58.0 & 44.8 & 46.6 & \textbf{47.5} & \underline{47.2} & 51.0 & \textbf{52.8} & \underline{52.7} & 52.2 & 55.3 & \underline{55.7} & \textbf{56.2} & \underline{55.7} \\
MMB & 89.2 & 84.0 & 84.6 & \underline{85.3} & \textbf{85.9} & 85.6 & 85.7 & \underline{86.1} & \textbf{86.2} & 86.2 & \underline{86.9} & \textbf{87.0} & \underline{86.9} \\
MMB-CN & 88.8 & \underline{82.7} & 82.3 & \textbf{84.1} & \textbf{84.1} & 84.0 & \underline{85.0} & \textbf{85.1} & \textbf{85.1} & 85.4 & 85.7 & \underline{86.4} & \textbf{86.7} \\
MME & 2466.0 & 2292.7 & 2364.0 & \underline{2382.9} & \textbf{2395.5} & 2398.6 & 2448.0 & \underline{2455.8} & \textbf{2456.2} & 2458.5 & \underline{2478.4} & 2477.0 & \textbf{2479.1} \\
POPE & 90.8 & 89.5 & \textbf{90.0} & \underline{89.9} & \textbf{90.0} & 89.8 & \textbf{90.3} & \underline{90.2} & \underline{90.2} & \underline{90.2} & \underline{90.2} & \textbf{90.3} & \underline{90.2} \\
SEED & 79.2 & 75.3 & \underline{76.1} & \textbf{76.3} & \textbf{76.3} & 76.4 & \underline{77.0} & \underline{77.0} & \textbf{77.1} & 77.0 & \underline{77.5} & \textbf{77.7} & \textbf{77.7} \\
AI2D & 87.8 & 76.0 & 78.6 & \underline{79.3} & \textbf{80.1} & 79.0 & 80.7 & \underline{81.4} & \textbf{82.7} & 82.3 & 83.1 & \underline{83.9} & \textbf{85.1} \\
SQA-I & 99.4 & 90.1 & \underline{92.5} & \underline{92.5} & \textbf{92.6} & 92.4 & 94.8 & \textbf{95.0} & \underline{94.9} & 95.0 & \underline{96.4} & \textbf{96.5} & \textbf{96.5} \\
\midrule
Avg.\ text-rich & 100.0 & 75.4 & \textbf{84.3} & \underline{84.1} & \underline{84.1} & 83.7 & 89.8 & \underline{90.4} & \textbf{90.9} & 91.1 & 94.5 & \underline{94.8} & \textbf{95.4} \\
Avg.\ general & 100.0 & 91.2 & 93.1 & \underline{93.8} & \textbf{94.1} & 94.2 & 95.9 & \underline{96.2} & \textbf{96.3} & 96.6 & 97.5 & \underline{97.9} & \textbf{98.0} \\
\bottomrule
\end{tabular}
\end{table*}

%% file: sections/appx/tab_app_checker_8b.tex
\begin{table*}[t]
\centering\scriptsize\setlength{\tabcolsep}{12.44pt}
\caption{\textbf{Other checkers on InternVL3-8B per benchmark.} PRUDENT (Ours) with the guide InternVL2.5-1B or InternVL3-2B (the default) as the checker, against IF-Prune, on the benchmarks of Table~\ref{tab:app-checker}. Full is the full-token score, and MME is out of 2800. The last two rows are the mean ratio to the full-token score, as in Table~\ref{tab:app-checker}. \textbf{Best} in bold, \protect\underline{second}{} underlined, per budget.}
\label{tab:app-checker-8b}
\begin{tabular}{@{}lcccccccccc@{}}
\toprule
 & & \multicolumn{3}{c}{Retain 2.5\%} & \multicolumn{3}{c}{Retain 5\%} & \multicolumn{3}{c}{Retain 10\%} \\
\cmidrule(lr){3-5}\cmidrule(lr){6-8}\cmidrule(lr){9-11}
Benchmark & Full & IF-Prune & 1B & 2B & IF-Prune & 1B & 2B & IF-Prune & 1B & 2B \\
\midrule
\multicolumn{11}{@{}l}{\textit{Text-rich}} \\
ChartQA & 86.6 & 60.0 & \underline{75.6} & \textbf{76.1} & 70.2 & \textbf{79.1} & \underline{78.5} & 77.6 & \underline{82.5} & \textbf{83.3} \\
DocVQA & 91.7 & 72.5 & \underline{82.4} & \textbf{82.9} & \underline{77.5} & \textbf{85.5} & \textbf{85.5} & 82.8 & \underline{88.2} & \textbf{88.4} \\
InfoVQA & 71.2 & 45.8 & \underline{52.8} & \textbf{53.1} & 53.2 & \underline{57.8} & \textbf{58.4} & 58.7 & \underline{63.3} & \textbf{64.3} \\
TextVQA & 80.9 & 73.3 & \underline{77.7} & \textbf{78.4} & 76.5 & \textbf{80.0} & \underline{79.9} & 79.0 & \underline{80.9} & \textbf{81.2} \\
OCRB. & 84.2 & 65.8 & \textbf{68.3} & \underline{67.6} & 74.7 & \underline{76.2} & \textbf{76.3} & 79.8 & \underline{81.0} & \textbf{81.8} \\
WebSRC & 90.8 & 68.0 & \underline{75.7} & \textbf{76.1} & 74.9 & \underline{80.9} & \textbf{81.3} & 79.3 & \underline{84.0} & \textbf{84.9} \\
\midrule
\multicolumn{11}{@{}l}{\textit{General}} \\
GQA & 62.9 & 60.4 & \underline{62.3} & \textbf{62.6} & 61.4 & \underline{62.9} & \textbf{63.0} & 62.8 & \underline{63.1} & \textbf{63.2} \\
VQAv2 & 79.7 & 74.9 & \underline{76.9} & \textbf{77.2} & 76.6 & \underline{78.1} & \textbf{78.5} & 77.9 & \underline{78.7} & \textbf{78.9} \\
OK-VQA & 49.0 & \underline{46.0} & 45.6 & \textbf{46.2} & 46.9 & \underline{47.3} & \textbf{47.8} & 47.9 & \underline{48.5} & \textbf{48.6} \\
MMB & 85.9 & 80.8 & \underline{82.0} & \textbf{82.3} & 82.6 & \underline{82.7} & \textbf{83.0} & 82.6 & \underline{84.1} & \textbf{84.2} \\
MMB-CN & 84.4 & 78.4 & \underline{78.7} & \textbf{79.4} & 79.6 & \underline{80.2} & \textbf{80.8} & 80.9 & \textbf{81.4} & \underline{81.2} \\
MME & 2383.9 & 2242.3 & \textbf{2309.6} & \underline{2304.0} & 2328.0 & \textbf{2362.9} & \underline{2352.3} & 2367.8 & \underline{2381.9} & \textbf{2402.8} \\
POPE & 91.0 & 90.1 & \textbf{90.7} & \underline{90.6} & \textbf{90.6} & \underline{90.5} & \textbf{90.6} & \textbf{90.5} & \underline{90.4} & \textbf{90.5} \\
SEED & 78.1 & 74.7 & \underline{75.6} & \textbf{76.0} & 75.8 & \underline{76.1} & \textbf{76.3} & 76.0 & \underline{76.6} & \textbf{76.7} \\
AI2D & 84.1 & 74.0 & \underline{76.5} & \textbf{77.0} & 76.7 & \underline{77.5} & \textbf{78.6} & 78.5 & \underline{80.1} & \textbf{80.4} \\
SQA-I & 98.4 & 90.0 & \underline{92.7} & \textbf{92.9} & 92.6 & \underline{93.8} & \textbf{94.1} & 94.0 & \underline{95.3} & \textbf{95.5} \\
\midrule
Avg.\ text-rich & 100.0 & 76.1 & \underline{85.3} & \textbf{85.6} & 84.3 & \underline{90.7} & \textbf{90.8} & 90.4 & \underline{94.8} & \textbf{95.7} \\
Avg.\ general & 100.0 & 93.9 & \underline{95.6} & \textbf{96.0} & 95.9 & \underline{96.9} & \textbf{97.3} & 97.2 & \underline{98.2} & \textbf{98.4} \\
\bottomrule
\end{tabular}
\end{table*}

%% file: sections/appx/tab_app_checker_auroc.tex
\begin{table}[t]
\centering\scriptsize\setlength{\tabcolsep}{5.26pt}
\caption{\textbf{Detecting pruning failures with each checker.} AUROC on InternVL3-38B per benchmark and first-pass budget. The first passes of 1\%, 2.5\% and 5\% belong to the budgets of 2.5\%, 5\% and 10\%. The order of PRUDENT (Ours) ranks the questions by the disagreement of the checker and then by the confidence of the large VLM. On GQA and VQAv2 the questions are those whose full-token answer is not the stock phrase of Table~\ref{tab:app-phrase}. The 1B checker is the guide, InternVL2.5-1B.}
\label{tab:app-checker-auroc}
\begin{tabular}{@{}lccccccccc@{}}
\toprule
 & \multicolumn{3}{c}{First pass 1\%} & \multicolumn{3}{c}{First pass 2.5\%} & \multicolumn{3}{c}{First pass 5\%} \\
\cmidrule(lr){2-4}\cmidrule(lr){5-7}\cmidrule(lr){8-10}
Benchmark & 1B & 2B & 8B & 1B & 2B & 8B & 1B & 2B & 8B \\
\midrule
\multicolumn{10}{@{}l}{\textit{Text-rich}} \\
ChartQA & 0.90 & 0.90 & 0.92 & 0.88 & 0.89 & 0.91 & 0.90 & 0.91 & 0.93 \\
DocVQA & 0.89 & 0.91 & 0.93 & 0.89 & 0.91 & 0.93 & 0.87 & 0.90 & 0.92 \\
InfoVQA & 0.74 & 0.80 & 0.82 & 0.75 & 0.80 & 0.83 & 0.77 & 0.82 & 0.84 \\
TextVQA & 0.87 & 0.89 & 0.91 & 0.86 & 0.88 & 0.90 & 0.86 & 0.88 & 0.89 \\
OCRB. & 0.88 & 0.90 & 0.90 & 0.84 & 0.87 & 0.87 & 0.82 & 0.88 & 0.90 \\
WebSRC & 0.92 & 0.94 & 0.95 & 0.86 & 0.89 & 0.90 & 0.85 & 0.88 & 0.90 \\
\midrule
\multicolumn{10}{@{}l}{\textit{General}} \\
GQA & 0.79 & 0.82 & 0.84 & 0.80 & 0.82 & 0.85 & 0.81 & 0.83 & 0.86 \\
VQAv2 & 0.83 & 0.87 & 0.88 & 0.82 & 0.86 & 0.87 & 0.82 & 0.86 & 0.87 \\
OK-VQA & 0.66 & 0.72 & 0.74 & 0.69 & 0.73 & 0.73 & 0.72 & 0.75 & 0.75 \\
MMB & 0.90 & 0.94 & 0.96 & 0.90 & 0.94 & 0.96 & 0.92 & 0.95 & 0.96 \\
MMB-CN & 0.89 & 0.94 & 0.96 & 0.89 & 0.93 & 0.97 & 0.89 & 0.93 & 0.96 \\
MME & 0.89 & 0.92 & 0.94 & 0.89 & 0.91 & 0.94 & 0.89 & 0.92 & 0.94 \\
POPE & 0.96 & 0.96 & 0.96 & 0.96 & 0.96 & 0.97 & 0.96 & 0.96 & 0.97 \\
SEED & 0.87 & 0.90 & 0.91 & 0.88 & 0.91 & 0.92 & 0.87 & 0.90 & 0.92 \\
AI2D & 0.88 & 0.91 & 0.93 & 0.88 & 0.91 & 0.93 & 0.87 & 0.91 & 0.93 \\
SQA-I & 0.98 & 0.99 & 0.99 & 0.98 & 0.99 & 0.99 & 0.97 & 0.99 & 0.99 \\
\midrule
Median & 0.88 & 0.91 & 0.92 & 0.88 & 0.90 & 0.91 & 0.87 & 0.90 & 0.92 \\
\bottomrule
\end{tabular}
\end{table}

%% file: sections/appx/tab_app_checker_qwen.tex
\begin{table}[t]
\centering\footnotesize\setlength{\tabcolsep}{4pt}
\caption{\textbf{Other checkers on Qwen3-VL.} PRUDENT (Ours) with Qwen3-VL-2B (the default) or Qwen3-VL-4B as the checker on Qwen3-VL-8B and 32B, at the 10\% budget of Table~\ref{tab:family} and on five text-rich benchmarks. Full is the full-token score. An asterisk marks a difference from SGL whose paired bootstrap 95\% interval excludes zero. \textbf{Best} in bold, \protect\underline{second}{} underlined, per model.}
\label{tab:app-checker-qwen}
\begin{tabular}{@{}lrrrrrrrr@{}}
    \toprule
     & \multicolumn{4}{c}{Qwen3-VL-8B} & \multicolumn{4}{c}{Qwen3-VL-32B} \\
    \cmidrule(lr){2-5}\cmidrule(lr){6-9}
     & Full & SGL & 2B & 4B & Full & SGL & 2B & 4B \\
    \midrule
    ChartQA & 85.2 & 74.9 & \textbf{77.7}$^*$ & \underline{76.7}$^*$ & 87.3 & 60.0 & \textbf{62.0}$^*$ & \underline{61.5} \\
    DocVQA & 95.5 & 93.0 & \underline{94.1}$^*$ & \textbf{94.5}$^*$ & 96.2 & 93.2 & \underline{94.5}$^*$ & \textbf{94.9}$^*$ \\
    InfoVQA & 82.0 & 73.8 & \underline{75.5}$^*$ & \textbf{75.6}$^*$ & 86.4 & 76.1 & \underline{78.2}$^*$ & \textbf{78.4}$^*$ \\
    TextVQA & 82.1 & 80.0 & \underline{81.3}$^*$ & \textbf{81.8}$^*$ & 83.5 & 80.8 & \underline{82.2}$^*$ & \textbf{82.5}$^*$ \\
    OCRBench & 85.5 & \underline{51.1} & \underline{51.1} & \textbf{51.2} & 86.3 & 52.4 & \underline{53.2} & \textbf{53.5} \\
    \midrule
    Mean & 86.1 & 74.5 & \textbf{76.0} & \underline{75.9} & 87.9 & 72.5 & \underline{74.0} & \textbf{74.2} \\
    \bottomrule
  \end{tabular}
\end{table}

%% file: sections/appx/tab_app_cost.tex
\begin{table*}[t]
\centering\footnotesize\setlength{\tabcolsep}{7.21pt}
\caption{\textbf{Cost at a 5\% budget.} Means over the seven benchmarks that we timed. Acc.\ is the score as a share of the full-token score, Latency is per question at batch size 1, Memory is the peak GPU memory, and FLOPs count the vision encoder and the prefill as a share of full tokens. Memory and FLOPs include the small VLMs. In the rows with + draft, the large VLM decodes speculatively from an answer of the guide, and PRUDENT (Ours) does so by default. Ours, 1B check uses the guide as the checker. For the existing methods without a small VLM, latency is that of our pipeline without the small VLMs plus their own selection time, measured alone. For FastV this is a lower bound, since its ranking needs the first layers on all tokens. \textbf{Best} accuracy in bold, \protect\underline{second}{} underlined.}
\label{tab:app-cost}
\begin{tabular}{lcccccccc}
\toprule
 & \multicolumn{4}{c}{InternVL3-38B} & \multicolumn{4}{c}{InternVL3-78B} \\
\cmidrule(lr){2-5}\cmidrule(lr){6-9}
Method & Acc.\ (\%) & Latency (ms) & Memory (GB) & FLOPs (\%) & Acc.\ (\%) & Latency (ms) & Memory (GB) & FLOPs (\%) \\
\midrule
\rowcolor{gray!12} Full tokens & 100.0 & 499 & 75.2 & 100.0 & 100.0 & 895 & 150.7 & 100.0 \\
Full tokens + draft & 100.0 & 475 & 77.1 & 103.2 & 100.0 & 811 & 152.7 & 101.8 \\
\midrule
ToMe & 42.2 & 336 & 73.2 & 45.2 & 45.2 & 536 & 148.3 & 29.3 \\
VisionZip & 40.9 & 336 & 73.2 & 45.2 & 42.6 & 536 & 148.3 & 29.3 \\
ZOO-Prune & 47.5 & 487 & 73.2 & 66.3 & 51.4 & 814 & 148.3 & 51.8 \\
FastV & 53.1 & 336 & 73.2 & 48.0 & 47.2 & 536 & 148.3 & 32.2 \\
CDPruner & 49.7 & 345 & 73.2 & 45.2 & 61.6 & 545 & 148.3 & 29.3 \\
MMTok & 52.8 & 341 & 73.2 & 45.2 & 49.5 & 541 & 148.3 & 29.3 \\
SGL & 85.4 & 368 & 75.2 & 48.4 & 86.7 & 567 & 150.3 & 31.1 \\
IF-Prune & 87.3 & 383 & 75.2 & 48.4 & 89.9 & 578 & 150.3 & 31.1 \\
IF-Prune + draft & 87.3 & 349 & 75.2 & 49.3 & 89.9 & 487 & 150.3 & 31.6 \\
\rowcolor{oursbg} \textbf{Ours} & \textbf{92.7} & 421 & 79.5 & 54.6 & \textbf{94.1} & 569 & 154.7 & 34.9 \\
Ours, 1B check & \underline{92.1} & 380 & 75.2 & 49.7 & \underline{93.7} & 527 & 150.3 & 32.1 \\
\bottomrule
\end{tabular}
\end{table*}

%% file: sections/appx/tab_app_ablate_38b.tex
\begin{table*}[t]
\centering\scriptsize\setlength{\tabcolsep}{1.30pt}
\caption{\textbf{Ablations on InternVL3-38B per benchmark.} The first two blocks of Table~\ref{tab:ablation} on each benchmark. The first row of each budget is the ratio of PRUDENT (Ours) to the full-token score. Every other row changes one thing and gives the change in points of the full-token score. Text and Gen.\ are the means over the text-rich and general benchmarks. RefCOCO is left out because the checker abstains on almost all of its questions. GQA and VQAv2 omit the questions where full tokens give a stock phrase (Appendix~\ref{app:setup}).}
\label{tab:app-ablate-38b}
\begin{tabular}{@{}>{\raggedright\arraybackslash}p{0.95in}cccccccccccccccccc@{}}
\toprule
 & \multicolumn{6}{c}{Text-rich} & \multicolumn{10}{c}{General} & \multicolumn{2}{c}{Mean} \\
\cmidrule(lr){2-7}\cmidrule(lr){8-17}\cmidrule(lr){18-19}
 & ChartQA & DocVQA & InfoVQA & TextVQA & OCRB. & WebSRC & GQA & VQAv2 & OK-VQA & MMB & MMB-CN & MME & POPE & SEED & AI2D & SQA-I & Text & Gen. \\
\midrule
\multicolumn{19}{@{}l}{\textit{Retain 2.5\% of the visual tokens}} \\
\rowcolor{oursbg} \textbf{PRUDENT (Ours)} & \textbf{80.8} & \textbf{91.5} & \textbf{74.4} & \textbf{96.6} & \textbf{77.6} & \textbf{83.3} & \textbf{97.5} & \textbf{93.1} & \textbf{81.9} & \textbf{95.7} & \textbf{94.7} & \textbf{96.6} & \textbf{99.0} & \textbf{96.3} & \textbf{90.3} & \textbf{93.1} & \textbf{84.1} & \textbf{93.8} \\
\multicolumn{19}{@{}l}{\textit{Change one part}} \\
IF-Prune's ranking in both passes & $-$18.8 & $-$7.3 & $-$7.3 & $-$3.3 & $-$4.3 & $-$5.3 & $-$0.9 & +0.5 & $-$6.7 & 0.0 & 0.0 & 0.0 & 0.0 & 0.0 & 0.0 & 0.0 & $-$7.7 & $-$0.7 \\
No checker, one pass at $K$ & $-$3.5 & $-$3.7 & +0.5 & $-$3.2 & $-$0.2 & $-$1.8 & $-$1.2 & $-$3.8 & +1.4 & $-$1.4 & $-$1.5 & $-$3.7 & $-$0.4 & $-$1.2 & $-$3.8 & $-$2.4 & $-$2.0 & $-$1.8 \\
The checker never abstains & $-$3.0 & $-$0.3 & $-$0.6 & $-$0.5 & $-$0.5 & $-$0.4 & $-$0.5 & $-$1.3 & $-$5.0 & +0.1 & $-$0.5 & +0.4 & 0.0 & 0.0 & +0.5 & 0.0 & $-$0.9 & $-$0.6 \\
IF-Prune's ranking in the re-answers & $-$2.1 & $-$1.5 & $-$1.0 & $-$1.1 & $-$0.5 & $-$0.4 & $-$0.3 & $-$0.3 & $-$0.6 & 0.0 & 0.0 & 0.0 & 0.0 & 0.0 & 0.0 & 0.0 & $-$1.1 & $-$0.1 \\
\multicolumn{19}{@{}l}{\textit{Choose the re-answered questions by}} \\
Random choice & $-$7.7 & $-$5.6 & $-$3.8 & $-$6.1 & $-$7.0 & $-$7.9 & $-$3.9 & $-$5.9 & $-$4.1 & $-$9.9 & $-$9.5 & $-$8.4 & $-$1.0 & $-$3.6 & $-$4.8 & $-$5.4 & $-$6.4 & $-$5.6 \\
SGL's early-exit score & $-$4.9 & $-$4.6 & $-$3.3 & $-$4.3 & $-$3.6 & $-$5.3 & $-$3.3 & $-$5.2 & $-$4.1 & $-$6.3 & $-$6.7 & $-$5.1 & $-$0.8 & $-$2.7 & $-$3.5 & $-$2.0 & $-$4.3 & $-$4.0 \\
Large VLM's confidence & $-$3.7 & $-$3.0 & $-$1.5 & $-$3.9 & $-$2.6 & $-$1.8 & $-$2.2 & $-$5.3 & $-$5.9 & $-$0.5 & $-$0.7 & $-$1.3 & +0.1 & $-$1.3 & $-$1.4 & $-$0.6 & $-$2.8 & $-$1.9 \\
Disagreement, in random order & $-$0.1 & $-$0.6 & +0.3 & 0.0 & $-$2.6 & $-$1.7 & 0.0 & $-$0.1 & 0.0 & $-$0.5 & $-$0.8 & $-$1.7 & +0.2 & +0.2 & +0.2 & +0.5 & $-$0.8 & $-$0.2 \\
\midrule
\multicolumn{19}{@{}l}{\textit{Retain 5\% of the visual tokens}} \\
\rowcolor{oursbg} \textbf{PRUDENT (Ours)} & \textbf{87.9} & \textbf{94.6} & \textbf{82.6} & \textbf{98.2} & \textbf{89.9} & \textbf{89.1} & \textbf{98.9} & \textbf{95.4} & \textbf{90.8} & \textbf{96.5} & \textbf{95.8} & \textbf{99.6} & \textbf{99.4} & \textbf{97.2} & \textbf{92.7} & \textbf{95.7} & \textbf{90.4} & \textbf{96.2} \\
\multicolumn{19}{@{}l}{\textit{Change one part}} \\
IF-Prune's ranking in both passes & $-$16.5 & $-$4.4 & $-$7.2 & $-$0.9 & $-$2.6 & $-$2.4 & $-$0.2 & $-$1.1 & $-$3.2 & 0.0 & 0.0 & 0.0 & 0.0 & 0.0 & 0.0 & 0.0 & $-$5.7 & $-$0.5 \\
No checker, one pass at $K$ & $-$4.8 & $-$2.6 & $-$1.8 & $-$1.6 & $-$0.5 & $-$2.6 & $-$1.3 & $-$2.9 & +0.7 & $-$0.6 & $-$1.3 & $-$2.3 & $-$0.5 & $-$0.8 & $-$2.7 & $-$2.6 & $-$2.3 & $-$1.4 \\
The checker never abstains & $-$2.5 & 0.0 & $-$1.9 & $-$0.2 & $-$3.0 & +0.3 & $-$0.4 & $-$1.8 & $-$3.8 & +0.3 & +0.3 & +0.2 & 0.0 & +0.1 & 0.0 & 0.0 & $-$1.2 & $-$0.5 \\
IF-Prune's ranking in the re-answers & $-$0.6 & $-$1.3 & $-$1.3 & $-$0.4 & $-$0.2 & +0.4 & $-$0.1 & $-$0.6 & $-$0.1 & 0.0 & 0.0 & 0.0 & 0.0 & 0.0 & 0.0 & 0.0 & $-$0.6 & $-$0.1 \\
\multicolumn{19}{@{}l}{\textit{Choose the re-answered questions by}} \\
Random choice & $-$7.1 & $-$4.7 & $-$3.4 & $-$3.1 & $-$4.9 & $-$5.0 & $-$1.9 & $-$4.0 & $-$2.1 & $-$2.0 & $-$3.6 & $-$5.3 & $-$0.6 & $-$1.6 & $-$4.7 & $-$3.7 & $-$4.7 & $-$3.0 \\
SGL's early-exit score & $-$5.1 & $-$3.1 & $-$2.2 & $-$2.2 & $-$3.9 & $-$3.1 & $-$1.7 & $-$3.5 & $-$1.5 & $-$1.3 & $-$2.7 & $-$2.4 & $-$0.4 & $-$1.2 & $-$2.7 & $-$2.0 & $-$3.3 & $-$1.9 \\
Large VLM's confidence & $-$3.0 & $-$2.7 & $-$1.9 & $-$2.0 & $-$2.9 & $-$1.7 & $-$1.4 & $-$4.0 & $-$2.7 & $-$0.2 & $-$0.1 & $-$0.9 & 0.0 & $-$0.4 & $-$0.8 & $-$0.4 & $-$2.4 & $-$1.1 \\
Disagreement, in random order & $-$1.3 & $-$0.4 & $-$0.7 & 0.0 & $-$0.4 & $-$0.8 & 0.0 & $-$0.4 & $-$0.4 & +0.2 & $-$0.9 & $-$0.8 & +0.2 & +0.2 & 0.0 & 0.0 & $-$0.6 & $-$0.2 \\
\midrule
\multicolumn{19}{@{}l}{\textit{Retain 10\% of the visual tokens}} \\
\rowcolor{oursbg} \textbf{PRUDENT (Ours)} & \textbf{94.2} & \textbf{97.5} & \textbf{88.2} & \textbf{99.2} & \textbf{96.6} & \textbf{93.0} & \textbf{99.1} & \textbf{97.4} & \textbf{96.8} & \textbf{97.6} & \textbf{97.3} & \textbf{100.4} & \textbf{99.4} & \textbf{98.0} & \textbf{95.5} & \textbf{97.2} & \textbf{94.8} & \textbf{97.9} \\
\multicolumn{19}{@{}l}{\textit{Change one part}} \\
IF-Prune's ranking in both passes & $-$8.0 & $-$1.5 & $-$2.6 & $-$0.3 & $-$0.9 & 0.0 & $-$0.2 & $-$0.5 & $-$2.1 & 0.0 & 0.0 & 0.0 & 0.0 & 0.0 & 0.0 & 0.0 & $-$2.2 & $-$0.3 \\
No checker, one pass at $K$ & $-$4.2 & $-$2.2 & $-$0.5 & $-$0.9 & $-$0.1 & $-$2.2 & $-$0.6 & $-$2.0 & +0.1 & $-$1.0 & $-$1.2 & $-$0.8 & $-$0.1 & $-$0.8 & $-$1.8 & $-$1.5 & $-$1.7 & $-$1.0 \\
The checker never abstains & $-$1.5 & $-$0.3 & $-$1.5 & +0.3 & $-$1.6 & $-$0.1 & +0.1 & $-$1.6 & $-$2.4 & +0.1 & +0.4 & $-$0.4 & 0.0 & 0.0 & 0.0 & 0.0 & $-$0.8 & $-$0.4 \\
IF-Prune's ranking in the re-answers & $-$0.1 & $-$0.6 & $-$0.2 & $-$0.1 & 0.0 & +0.5 & $-$0.1 & $-$0.4 & $-$0.5 & 0.0 & 0.0 & 0.0 & 0.0 & 0.0 & 0.0 & 0.0 & $-$0.1 & $-$0.1 \\
\multicolumn{19}{@{}l}{\textit{Choose the re-answered questions by}} \\
Random choice & $-$7.6 & $-$3.6 & $-$3.8 & $-$1.8 & $-$3.4 & $-$3.7 & $-$1.2 & $-$3.6 & $-$1.8 & $-$1.7 & $-$2.5 & $-$1.9 & $-$0.4 & $-$1.2 & $-$3.8 & $-$2.8 & $-$4.0 & $-$2.1 \\
SGL's early-exit score & $-$5.7 & $-$2.4 & $-$2.8 & $-$1.1 & $-$2.7 & $-$2.2 & $-$1.2 & $-$3.1 & $-$1.7 & $-$1.2 & $-$2.2 & $-$1.1 & 0.0 & $-$0.9 & $-$1.8 & $-$1.8 & $-$2.8 & $-$1.5 \\
Large VLM's confidence & $-$3.4 & $-$2.3 & $-$1.4 & $-$1.1 & $-$2.2 & $-$1.5 & $-$0.8 & $-$3.4 & $-$1.8 & $-$0.1 & $-$0.1 & $-$0.2 & 0.0 & $-$0.3 & +0.1 & $-$0.2 & $-$2.0 & $-$0.7 \\
Disagreement, in random order & $-$1.1 & $-$0.1 & $-$1.2 & +0.1 & $-$0.1 & $-$0.3 & 0.0 & $-$0.1 & $-$0.5 & 0.0 & $-$0.2 & $-$0.9 & +0.1 & 0.0 & 0.0 & 0.0 & $-$0.5 & $-$0.2 \\
\bottomrule
\end{tabular}
\end{table*}

%% file: sections/appx/tab_app_ablate_8b.tex
\begin{table*}[t]
\centering\scriptsize\setlength{\tabcolsep}{1.32pt}
\caption{\textbf{Ablations on InternVL3-8B per benchmark.} The first two blocks of Table~\ref{tab:ablation} on each benchmark. The first row of each budget is the ratio of PRUDENT (Ours) to the full-token score. Every other row changes one thing and gives the change in points of the full-token score. Text and Gen.\ are the means over the text-rich and general benchmarks. RefCOCO is left out because the checker abstains on almost all of its questions.}
\label{tab:app-ablate-8b}
\begin{tabular}{@{}>{\raggedright\arraybackslash}p{0.95in}cccccccccccccccccc@{}}
\toprule
 & \multicolumn{6}{c}{Text-rich} & \multicolumn{10}{c}{General} & \multicolumn{2}{c}{Mean} \\
\cmidrule(lr){2-7}\cmidrule(lr){8-17}\cmidrule(lr){18-19}
 & ChartQA & DocVQA & InfoVQA & TextVQA & OCRB. & WebSRC & GQA & VQAv2 & OK-VQA & MMB & MMB-CN & MME & POPE & SEED & AI2D & SQA-I & Text & Gen. \\
\midrule
\multicolumn{19}{@{}l}{\textit{Retain 2.5\% of the visual tokens}} \\
\rowcolor{oursbg} \textbf{PRUDENT (Ours)} & \textbf{87.9} & \textbf{90.3} & \textbf{74.6} & \textbf{96.9} & \textbf{80.3} & \textbf{83.8} & \textbf{99.5} & \textbf{96.9} & \textbf{94.2} & \textbf{95.8} & \textbf{94.1} & \textbf{96.6} & \textbf{99.5} & \textbf{97.3} & \textbf{91.6} & \textbf{94.4} & \textbf{85.6} & \textbf{96.0} \\
\multicolumn{19}{@{}l}{\textit{Change one part}} \\
IF-Prune's ranking in both passes & $-$14.0 & $-$9.4 & $-$10.0 & $-$3.2 & $-$4.4 & $-$6.4 & $-$1.6 & $-$1.0 & +0.6 & 0.0 & 0.0 & 0.0 & 0.0 & 0.0 & 0.0 & 0.0 & $-$7.9 & $-$0.2 \\
No checker, one pass at $K$ & $-$6.4 & $-$3.6 & $-$1.5 & $-$3.8 & 0.0 & $-$1.5 & $-$2.4 & $-$1.8 & $-$0.5 & $-$1.8 & $-$1.1 & $-$2.6 & $-$0.5 & $-$1.7 & $-$3.5 & $-$2.9 & $-$2.8 & $-$1.9 \\
The checker never abstains & $-$2.3 & $-$0.2 & $-$1.4 & $-$0.6 & $-$3.1 & $-$0.5 & $-$0.4 & $-$0.1 & $-$1.9 & +0.2 & 0.0 & 0.0 & +0.1 & +0.1 & +0.3 & +0.1 & $-$1.3 & $-$0.2 \\
IF-Prune's ranking in the re-answers & $-$1.9 & $-$2.7 & $-$2.6 & $-$1.1 & $-$0.6 & $-$1.3 & $-$0.4 & $-$0.3 & $-$0.3 & 0.0 & 0.0 & 0.0 & 0.0 & 0.0 & 0.0 & 0.0 & $-$1.7 & $-$0.1 \\
\multicolumn{19}{@{}l}{\textit{Choose the re-answered questions by}} \\
Random choice & $-$8.4 & $-$5.9 & $-$4.8 & $-$6.7 & $-$5.8 & $-$7.4 & $-$3.7 & $-$3.8 & $-$3.1 & $-$6.3 & $-$5.6 & $-$3.2 & $-$1.0 & $-$3.6 & $-$5.9 & $-$5.7 & $-$6.5 & $-$4.2 \\
SGL's early-exit score & $-$5.1 & $-$4.3 & $-$3.5 & $-$5.2 & $-$4.3 & $-$4.7 & $-$3.0 & $-$3.2 & $-$2.7 & $-$4.0 & $-$4.6 & $-$2.7 & $-$0.5 & $-$2.4 & $-$3.9 & $-$1.8 & $-$4.5 & $-$2.9 \\
Large VLM's confidence & $-$4.0 & $-$2.7 & $-$2.9 & $-$4.1 & $-$3.3 & $-$2.9 & $-$2.6 & $-$2.6 & $-$4.0 & $-$0.3 & $-$0.4 & $-$0.6 & $-$0.1 & $-$1.2 & $-$2.6 & $-$0.9 & $-$3.3 & $-$1.5 \\
Disagreement, in random order & $-$0.3 & $-$0.9 & $-$0.5 & 0.0 & $-$1.4 & $-$1.9 & 0.0 & 0.0 & $-$0.2 & $-$0.7 & $-$0.7 & $-$0.6 & +0.2 & +0.1 & $-$0.2 & +0.5 & $-$0.8 & $-$0.2 \\
\midrule
\multicolumn{19}{@{}l}{\textit{Retain 5\% of the visual tokens}} \\
\rowcolor{oursbg} \textbf{PRUDENT (Ours)} & \textbf{90.7} & \textbf{93.1} & \textbf{82.0} & \textbf{98.8} & \textbf{90.6} & \textbf{89.5} & \textbf{100.1} & \textbf{98.5} & \textbf{97.6} & \textbf{96.6} & \textbf{95.7} & \textbf{98.7} & \textbf{99.5} & \textbf{97.7} & \textbf{93.5} & \textbf{95.6} & \textbf{90.8} & \textbf{97.3} \\
\multicolumn{19}{@{}l}{\textit{Change one part}} \\
IF-Prune's ranking in both passes & $-$9.1 & $-$5.3 & $-$8.5 & $-$1.3 & $-$2.0 & $-$5.0 & $-$0.8 & $-$0.9 & $-$0.5 & 0.0 & 0.0 & 0.0 & 0.0 & 0.0 & 0.0 & 0.0 & $-$5.2 & $-$0.2 \\
No checker, one pass at $K$ & $-$1.7 & $-$2.7 & +0.1 & $-$2.1 & $-$0.5 & $-$2.6 & $-$1.0 & $-$1.3 & $-$0.4 & $-$0.4 & $-$1.4 & $-$1.0 & 0.0 & $-$0.6 & $-$2.3 & $-$1.6 & $-$1.6 & $-$1.0 \\
The checker never abstains & $-$1.7 & $-$0.5 & $-$1.6 & $-$0.2 & $-$2.7 & $-$0.1 & +0.1 & $-$0.4 & $-$1.6 & $-$0.4 & +0.3 & +0.3 & 0.0 & $-$0.1 & $-$0.3 & 0.0 & $-$1.1 & $-$0.2 \\
IF-Prune's ranking in the re-answers & $-$0.5 & $-$1.3 & $-$1.5 & $-$0.6 & $-$0.5 & $-$0.6 & $-$0.1 & $-$0.3 & 0.0 & 0.0 & 0.0 & 0.0 & 0.0 & 0.0 & 0.0 & 0.0 & $-$0.8 & 0.0 \\
\multicolumn{19}{@{}l}{\textit{Choose the re-answered questions by}} \\
Random choice & $-$6.7 & $-$4.1 & $-$4.2 & $-$4.0 & $-$4.8 & $-$4.9 & $-$1.9 & $-$2.3 & $-$1.6 & $-$2.5 & $-$3.3 & $-$4.2 & $-$0.3 & $-$1.6 & $-$3.7 & $-$3.0 & $-$4.8 & $-$2.5 \\
SGL's early-exit score & $-$3.3 & $-$3.2 & $-$3.5 & $-$3.1 & $-$3.2 & $-$3.2 & $-$1.5 & $-$1.8 & $-$1.5 & $-$2.2 & $-$2.0 & $-$1.7 & 0.0 & $-$1.2 & $-$3.2 & $-$1.3 & $-$3.2 & $-$1.6 \\
Large VLM's confidence & $-$2.4 & $-$2.7 & $-$2.2 & $-$2.3 & $-$3.0 & $-$2.3 & $-$0.9 & $-$1.4 & $-$2.4 & $-$0.4 & $-$0.3 & +0.3 & 0.0 & $-$0.7 & $-$1.5 & $-$0.6 & $-$2.5 & $-$0.8 \\
Disagreement, in random order & $-$0.1 & $-$0.3 & $-$0.8 & 0.0 & $-$1.9 & $-$0.6 & 0.0 & +0.1 & +0.4 & +0.4 & $-$0.2 & $-$0.6 & 0.0 & +0.2 & $-$0.1 & +0.2 & $-$0.6 & 0.0 \\
\midrule
\multicolumn{19}{@{}l}{\textit{Retain 10\% of the visual tokens}} \\
\rowcolor{oursbg} \textbf{PRUDENT (Ours)} & \textbf{96.2} & \textbf{96.4} & \textbf{90.3} & \textbf{100.5} & \textbf{97.1} & \textbf{93.5} & \textbf{100.5} & \textbf{98.9} & \textbf{99.1} & \textbf{98.0} & \textbf{96.2} & \textbf{100.8} & \textbf{99.4} & \textbf{98.1} & \textbf{95.7} & \textbf{97.1} & \textbf{95.7} & \textbf{98.4} \\
\multicolumn{19}{@{}l}{\textit{Change one part}} \\
IF-Prune's ranking in both passes & $-$3.5 & $-$3.2 & $-$8.3 & $-$1.1 & $-$2.7 & $-$2.3 & $-$0.2 & $-$0.4 & $-$0.8 & 0.0 & 0.0 & 0.0 & 0.0 & 0.0 & 0.0 & 0.0 & $-$3.5 & $-$0.1 \\
No checker, one pass at $K$ & $-$3.2 & $-$2.6 & $-$1.0 & $-$1.9 & $-$0.4 & $-$2.7 & $-$0.1 & $-$0.6 & $-$0.4 & $-$1.8 & $-$0.3 & $-$1.5 & 0.0 & $-$0.9 & $-$2.3 & $-$1.6 & $-$2.0 & $-$0.9 \\
The checker never abstains & $-$1.5 & $-$0.1 & $-$2.1 & +0.4 & $-$3.1 & $-$0.3 & 0.0 & +0.2 & $-$1.4 & +0.3 & +0.7 & $-$0.5 & 0.0 & +0.1 & +0.2 & 0.0 & $-$1.1 & $-$0.1 \\
IF-Prune's ranking in the re-answers & $-$0.1 & $-$0.6 & $-$1.7 & $-$0.8 & 0.0 & $-$0.3 & 0.0 & $-$0.1 & $-$0.2 & 0.0 & 0.0 & 0.0 & 0.0 & 0.0 & 0.0 & 0.0 & $-$0.6 & 0.0 \\
\multicolumn{19}{@{}l}{\textit{Choose the re-answered questions by}} \\
Random choice & $-$4.6 & $-$4.1 & $-$3.1 & $-$2.7 & $-$2.4 & $-$4.6 & $-$1.0 & $-$1.2 & $-$1.2 & $-$2.2 & $-$1.3 & $-$1.2 & $-$0.1 & $-$1.1 & $-$3.4 & $-$2.8 & $-$3.6 & $-$1.5 \\
SGL's early-exit score & $-$2.7 & $-$3.0 & $-$2.8 & $-$2.3 & $-$3.0 & $-$2.6 & $-$0.7 & $-$1.1 & $-$0.9 & $-$2.2 & $-$0.9 & $-$0.6 & 0.0 & $-$1.0 & $-$2.7 & $-$1.7 & $-$2.7 & $-$1.2 \\
Large VLM's confidence & $-$2.3 & $-$2.5 & $-$2.0 & $-$1.7 & $-$2.3 & $-$2.2 & $-$0.2 & $-$1.0 & $-$1.6 & $-$0.5 & +0.2 & 0.0 & 0.0 & $-$0.3 & $-$0.8 & $-$0.2 & $-$2.1 & $-$0.4 \\
Disagreement, in random order & 0.0 & $-$0.1 & $-$1.1 & 0.0 & $-$0.4 & $-$0.2 & +0.1 & +0.1 & $-$0.1 & +0.1 & $-$0.6 & $-$1.4 & +0.2 & +0.2 & $-$0.2 & +0.6 & $-$0.3 & $-$0.1 \\
\bottomrule
\end{tabular}
\end{table*}

%% file: sections/appx/tab_app_abstain_38b.tex
\begin{table*}[t]
\centering\scriptsize\setlength{\tabcolsep}{1.30pt}
\caption{\textbf{Abstention of the checker on InternVL3-38B.} The checker abstains when the least token probability of its answer is below 0.5 or the answer is longer than 10 words. Each row moves one of the two on every benchmark of Table~\ref{tab:ablation}, and the first row of each budget is the ratio of PRUDENT (Ours) to the full-token score. The other rows give the change in points of the full-token score. Text and Gen.\ are the means over the text-rich and general benchmarks. GQA and VQAv2 omit the questions where full tokens give a stock phrase (Appendix~\ref{app:setup}).}
\label{tab:app-abstain-38b}
\begin{tabular}{@{}>{\raggedright\arraybackslash}p{0.95in}cccccccccccccccccc@{}}
\toprule
 & \multicolumn{6}{c}{Text-rich} & \multicolumn{10}{c}{General} & \multicolumn{2}{c}{Mean} \\
\cmidrule(lr){2-7}\cmidrule(lr){8-17}\cmidrule(lr){18-19}
 & ChartQA & DocVQA & InfoVQA & TextVQA & OCRB. & WebSRC & GQA & VQAv2 & OK-VQA & MMB & MMB-CN & MME & POPE & SEED & AI2D & SQA-I & Text & Gen. \\
\midrule
\multicolumn{19}{@{}l}{\textit{Retain 2.5\% of the visual tokens}} \\
\rowcolor{oursbg} \textbf{PRUDENT (Ours)} & \textbf{80.8} & \textbf{91.5} & \textbf{74.4} & \textbf{96.6} & \textbf{77.6} & \textbf{83.3} & \textbf{97.5} & \textbf{93.1} & \textbf{81.9} & \textbf{95.7} & \textbf{94.7} & \textbf{96.6} & \textbf{99.0} & \textbf{96.3} & \textbf{90.3} & \textbf{93.1} & \textbf{84.1} & \textbf{93.8} \\
\multicolumn{19}{@{}l}{\textit{Change when the checker abstains, default probability 0.5 and 10 words}} \\
Below probability 0.3 & $-$1.6 & $-$0.1 & $-$0.5 & $-$0.1 & $-$0.2 & $-$0.3 & $-$0.3 & $-$0.6 & $-$3.3 & +0.1 & $-$0.5 & +0.4 & 0.0 & 0.0 & +0.5 & 0.0 & $-$0.5 & $-$0.4 \\
Below probability 0.4 & $-$0.8 & 0.0 & $-$0.4 & 0.0 & $-$0.2 & $-$0.2 & $-$0.1 & $-$0.1 & $-$1.8 & 0.0 & $-$0.4 & +0.4 & 0.0 & 0.0 & +0.3 & 0.0 & $-$0.3 & $-$0.2 \\
Below probability 0.6 & +0.4 & $-$0.3 & +0.2 & $-$0.4 & +1.3 & 0.0 & $-$0.1 & $-$0.7 & +1.1 & 0.0 & $-$0.4 & $-$0.1 & 0.0 & 0.0 & $-$0.3 & 0.0 & +0.2 & 0.0 \\
Below probability 0.7 & +1.5 & $-$0.6 & +0.8 & $-$1.0 & +1.2 & +0.1 & $-$0.3 & $-$1.3 & +1.7 & $-$0.4 & $-$0.3 & $-$1.1 & +0.1 & 0.0 & $-$1.0 & 0.0 & +0.3 & $-$0.3 \\
Below probability 0.8 & +1.8 & $-$1.3 & +1.3 & $-$1.5 & +0.8 & $-$0.1 & $-$0.2 & $-$2.0 & +2.0 & $-$0.3 & $-$1.2 & $-$1.1 & $-$0.1 & $-$0.3 & $-$1.4 & $-$0.1 & +0.2 & $-$0.5 \\
Above 5 words & 0.0 & +0.1 & 0.0 & $-$0.1 & +1.1 & +0.3 & 0.0 & 0.0 & 0.0 & 0.0 & 0.0 & 0.0 & 0.0 & 0.0 & 0.0 & 0.0 & +0.2 & 0.0 \\
Above 20 words & 0.0 & 0.0 & 0.0 & 0.0 & +0.1 & 0.0 & 0.0 & 0.0 & 0.0 & 0.0 & 0.0 & 0.0 & 0.0 & 0.0 & 0.0 & 0.0 & 0.0 & 0.0 \\
\midrule
\multicolumn{19}{@{}l}{\textit{Retain 5\% of the visual tokens}} \\
\rowcolor{oursbg} \textbf{PRUDENT (Ours)} & \textbf{87.9} & \textbf{94.6} & \textbf{82.6} & \textbf{98.2} & \textbf{89.9} & \textbf{89.1} & \textbf{98.9} & \textbf{95.4} & \textbf{90.8} & \textbf{96.5} & \textbf{95.8} & \textbf{99.6} & \textbf{99.4} & \textbf{97.2} & \textbf{92.7} & \textbf{95.7} & \textbf{90.4} & \textbf{96.2} \\
\multicolumn{19}{@{}l}{\textit{Change when the checker abstains, default probability 0.5 and 10 words}} \\
Below probability 0.3 & $-$0.9 & $-$0.1 & $-$1.2 & $-$0.1 & $-$0.2 & +0.3 & $-$0.3 & $-$1.1 & $-$2.6 & +0.3 & +0.3 & +0.2 & 0.0 & +0.1 & 0.0 & 0.0 & $-$0.4 & $-$0.3 \\
Below probability 0.4 & $-$0.4 & $-$0.2 & $-$0.5 & +0.2 & +0.1 & +0.1 & $-$0.2 & $-$0.4 & $-$1.1 & +0.1 & +0.2 & +0.2 & 0.0 & +0.1 & +0.1 & 0.0 & $-$0.1 & $-$0.1 \\
Below probability 0.6 & +0.2 & +0.2 & $-$0.1 & $-$0.3 & +0.7 & 0.0 & $-$0.3 & $-$0.3 & +0.7 & +0.2 & $-$0.5 & $-$0.2 & $-$0.1 & $-$0.1 & $-$0.2 & $-$0.2 & +0.1 & $-$0.1 \\
Below probability 0.7 & +0.2 & $-$0.1 & $-$0.2 & $-$0.6 & +0.6 & $-$0.2 & $-$0.5 & $-$0.8 & +1.1 & 0.0 & $-$0.6 & $-$0.7 & $-$0.2 & $-$0.1 & $-$0.2 & $-$0.4 & $-$0.1 & $-$0.2 \\
Below probability 0.8 & +0.8 & $-$0.6 & $-$0.2 & $-$0.9 & +1.1 & $-$0.4 & $-$0.8 & $-$1.3 & +1.3 & $-$0.2 & $-$0.7 & $-$1.1 & $-$0.3 & $-$0.2 & $-$0.4 & $-$0.5 & 0.0 & $-$0.4 \\
Above 5 words & $-$0.1 & 0.0 & 0.0 & $-$0.1 & +1.2 & +0.1 & 0.0 & 0.0 & 0.0 & 0.0 & 0.0 & 0.0 & 0.0 & 0.0 & 0.0 & 0.0 & +0.2 & 0.0 \\
Above 20 words & 0.0 & 0.0 & 0.0 & 0.0 & $-$0.8 & 0.0 & 0.0 & 0.0 & 0.0 & 0.0 & 0.0 & 0.0 & 0.0 & 0.0 & 0.0 & 0.0 & $-$0.2 & 0.0 \\
\midrule
\multicolumn{19}{@{}l}{\textit{Retain 10\% of the visual tokens}} \\
\rowcolor{oursbg} \textbf{PRUDENT (Ours)} & \textbf{94.2} & \textbf{97.5} & \textbf{88.2} & \textbf{99.2} & \textbf{96.6} & \textbf{93.0} & \textbf{99.1} & \textbf{97.4} & \textbf{96.8} & \textbf{97.6} & \textbf{97.3} & \textbf{100.4} & \textbf{99.4} & \textbf{98.0} & \textbf{95.5} & \textbf{97.2} & \textbf{94.8} & \textbf{97.9} \\
\multicolumn{19}{@{}l}{\textit{Change when the checker abstains, default probability 0.5 and 10 words}} \\
Below probability 0.3 & $-$0.7 & $-$0.1 & $-$1.3 & +0.3 & +0.2 & $-$0.1 & +0.2 & $-$0.4 & $-$1.4 & +0.2 & +0.4 & $-$0.4 & 0.0 & 0.0 & 0.0 & 0.0 & $-$0.3 & $-$0.1 \\
Below probability 0.4 & $-$0.1 & $-$0.1 & $-$0.8 & +0.2 & +0.1 & $-$0.1 & +0.3 & $-$0.1 & $-$0.5 & +0.2 & +0.4 & $-$0.4 & 0.0 & +0.1 & +0.1 & 0.0 & $-$0.1 & 0.0 \\
Below probability 0.6 & $-$0.2 & $-$0.3 & +0.3 & $-$0.2 & +0.2 & 0.0 & 0.0 & $-$0.2 & +0.2 & $-$0.2 & $-$0.5 & $-$0.4 & 0.0 & $-$0.1 & +0.3 & 0.0 & 0.0 & $-$0.1 \\
Below probability 0.7 & 0.0 & $-$0.7 & +0.1 & $-$0.1 & +0.1 & $-$0.4 & $-$0.1 & $-$0.7 & +0.5 & $-$0.4 & $-$0.5 & $-$0.1 & 0.0 & $-$0.2 & $-$0.1 & $-$0.1 & $-$0.2 & $-$0.2 \\
Below probability 0.8 & $-$0.8 & $-$0.9 & +0.6 & $-$0.3 & +0.2 & $-$0.7 & $-$0.1 & $-$1.0 & +0.1 & $-$0.4 & $-$0.8 & $-$0.6 & 0.0 & $-$0.3 & $-$0.3 & $-$0.2 & $-$0.3 & $-$0.4 \\
Above 5 words & 0.0 & 0.0 & $-$0.1 & $-$0.1 & +0.2 & +0.1 & 0.0 & 0.0 & 0.0 & 0.0 & 0.0 & 0.0 & 0.0 & 0.0 & 0.0 & 0.0 & 0.0 & 0.0 \\
Above 20 words & 0.0 & 0.0 & 0.0 & 0.0 & $-$0.2 & 0.0 & 0.0 & 0.0 & 0.0 & 0.0 & 0.0 & 0.0 & 0.0 & 0.0 & 0.0 & 0.0 & 0.0 & 0.0 \\
\bottomrule
\end{tabular}
\end{table*}

%% file: sections/appx/tab_app_abstain_8b.tex
\begin{table*}[t]
\centering\scriptsize\setlength{\tabcolsep}{1.32pt}
\caption{\textbf{Abstention of the checker on InternVL3-8B.} The checker abstains when the least token probability of its answer is below 0.5 or the answer is longer than 10 words. Each row moves one of the two on every benchmark of Table~\ref{tab:ablation}, and the first row of each budget is the ratio of PRUDENT (Ours) to the full-token score. The other rows give the change in points of the full-token score. Text and Gen.\ are the means over the text-rich and general benchmarks.}
\label{tab:app-abstain-8b}
\begin{tabular}{@{}>{\raggedright\arraybackslash}p{0.95in}cccccccccccccccccc@{}}
\toprule
 & \multicolumn{6}{c}{Text-rich} & \multicolumn{10}{c}{General} & \multicolumn{2}{c}{Mean} \\
\cmidrule(lr){2-7}\cmidrule(lr){8-17}\cmidrule(lr){18-19}
 & ChartQA & DocVQA & InfoVQA & TextVQA & OCRB. & WebSRC & GQA & VQAv2 & OK-VQA & MMB & MMB-CN & MME & POPE & SEED & AI2D & SQA-I & Text & Gen. \\
\midrule
\multicolumn{19}{@{}l}{\textit{Retain 2.5\% of the visual tokens}} \\
\rowcolor{oursbg} \textbf{PRUDENT (Ours)} & \textbf{87.9} & \textbf{90.3} & \textbf{74.6} & \textbf{96.9} & \textbf{80.3} & \textbf{83.8} & \textbf{99.5} & \textbf{96.9} & \textbf{94.2} & \textbf{95.8} & \textbf{94.1} & \textbf{96.6} & \textbf{99.5} & \textbf{97.3} & \textbf{91.6} & \textbf{94.4} & \textbf{85.6} & \textbf{96.0} \\
\multicolumn{19}{@{}l}{\textit{Change when the checker abstains, default probability 0.5 and 10 words}} \\
Below probability 0.3 & $-$0.7 & 0.0 & $-$1.1 & $-$0.2 & $-$0.4 & $-$0.5 & $-$0.1 & +0.5 & $-$0.9 & +0.2 & 0.0 & 0.0 & +0.1 & +0.1 & +0.2 & +0.1 & $-$0.5 & 0.0 \\
Below probability 0.4 & $-$0.7 & +0.1 & $-$0.6 & +0.2 & 0.0 & $-$0.4 & 0.0 & +0.4 & $-$0.6 & 0.0 & $-$0.1 & $-$0.1 & +0.1 & +0.1 & 0.0 & +0.1 & $-$0.2 & 0.0 \\
Below probability 0.6 & +0.5 & $-$0.1 & +0.4 & $-$0.5 & +0.2 & +0.3 & $-$0.4 & $-$0.2 & +0.7 & $-$0.2 & $-$0.1 & $-$0.1 & $-$0.1 & $-$0.1 & $-$0.3 & $-$0.2 & +0.2 & $-$0.1 \\
Below probability 0.7 & +0.1 & $-$0.2 & +0.5 & $-$1.0 & +0.2 & +0.4 & $-$0.9 & $-$0.4 & +0.9 & $-$0.2 & $-$0.4 & +0.2 & 0.0 & $-$0.3 & $-$0.4 & $-$0.3 & 0.0 & $-$0.2 \\
Below probability 0.8 & $-$0.8 & $-$0.6 & +0.4 & $-$1.7 & +0.5 & +0.1 & $-$1.5 & $-$0.7 & +0.4 & $-$0.3 & $-$0.6 & $-$0.3 & $-$0.2 & $-$0.7 & $-$1.3 & $-$0.5 & $-$0.4 & $-$0.6 \\
Above 5 words & 0.0 & +0.1 & +0.1 & $-$0.1 & +0.8 & +0.3 & 0.0 & 0.0 & 0.0 & 0.0 & 0.0 & 0.0 & 0.0 & 0.0 & 0.0 & 0.0 & +0.2 & 0.0 \\
Above 20 words & 0.0 & 0.0 & 0.0 & 0.0 & $-$0.2 & +0.1 & 0.0 & 0.0 & 0.0 & 0.0 & 0.0 & 0.0 & 0.0 & 0.0 & 0.0 & 0.0 & 0.0 & 0.0 \\
\midrule
\multicolumn{19}{@{}l}{\textit{Retain 5\% of the visual tokens}} \\
\rowcolor{oursbg} \textbf{PRUDENT (Ours)} & \textbf{90.7} & \textbf{93.1} & \textbf{82.0} & \textbf{98.8} & \textbf{90.6} & \textbf{89.5} & \textbf{100.1} & \textbf{98.5} & \textbf{97.6} & \textbf{96.6} & \textbf{95.7} & \textbf{98.7} & \textbf{99.5} & \textbf{97.7} & \textbf{93.5} & \textbf{95.6} & \textbf{90.8} & \textbf{97.3} \\
\multicolumn{19}{@{}l}{\textit{Change when the checker abstains, default probability 0.5 and 10 words}} \\
Below probability 0.3 & $-$0.5 & $-$0.2 & $-$1.0 & +0.1 & $-$0.1 & $-$0.1 & 0.0 & +0.1 & $-$1.0 & $-$0.4 & +0.3 & +0.3 & 0.0 & 0.0 & $-$0.2 & 0.0 & $-$0.3 & $-$0.1 \\
Below probability 0.4 & $-$0.1 & $-$0.1 & $-$0.3 & +0.2 & $-$0.5 & $-$0.1 & +0.1 & +0.1 & $-$0.4 & $-$0.4 & +0.3 & +0.3 & 0.0 & $-$0.1 & $-$0.1 & 0.0 & $-$0.2 & 0.0 \\
Below probability 0.6 & +0.4 & +0.4 & +0.8 & $-$0.4 & +0.5 & +0.1 & $-$0.2 & $-$0.3 & +0.4 & +0.1 & 0.0 & $-$0.3 & 0.0 & $-$0.1 & $-$0.3 & +0.1 & +0.3 & $-$0.1 \\
Below probability 0.7 & +0.7 & +0.2 & +0.8 & $-$0.7 & $-$0.1 & +0.3 & $-$0.5 & $-$0.5 & +0.5 & 0.0 & $-$0.2 & $-$0.4 & +0.1 & 0.0 & $-$0.5 & 0.0 & +0.2 & $-$0.1 \\
Below probability 0.8 & +0.7 & $-$0.5 & +0.9 & $-$1.2 & $-$0.1 & $-$0.1 & $-$0.7 & $-$0.6 & +0.3 & $-$0.3 & $-$0.7 & $-$0.3 & 0.0 & $-$0.2 & $-$0.2 & +0.1 & $-$0.1 & $-$0.3 \\
Above 5 words & +0.1 & $-$0.1 & +0.1 & $-$0.1 & +0.1 & +0.2 & 0.0 & 0.0 & 0.0 & 0.0 & 0.0 & 0.0 & 0.0 & 0.0 & 0.0 & 0.0 & +0.1 & 0.0 \\
Above 20 words & 0.0 & 0.0 & 0.0 & 0.0 & $-$0.8 & 0.0 & 0.0 & 0.0 & 0.0 & 0.0 & 0.0 & 0.0 & 0.0 & 0.0 & 0.0 & 0.0 & $-$0.1 & 0.0 \\
\midrule
\multicolumn{19}{@{}l}{\textit{Retain 10\% of the visual tokens}} \\
\rowcolor{oursbg} \textbf{PRUDENT (Ours)} & \textbf{96.2} & \textbf{96.4} & \textbf{90.3} & \textbf{100.5} & \textbf{97.1} & \textbf{93.5} & \textbf{100.5} & \textbf{98.9} & \textbf{99.1} & \textbf{98.0} & \textbf{96.2} & \textbf{100.8} & \textbf{99.4} & \textbf{98.1} & \textbf{95.7} & \textbf{97.1} & \textbf{95.7} & \textbf{98.4} \\
\multicolumn{19}{@{}l}{\textit{Change when the checker abstains, default probability 0.5 and 10 words}} \\
Below probability 0.3 & $-$0.6 & $-$0.1 & $-$1.3 & +0.4 & $-$0.7 & $-$0.2 & +0.1 & +0.2 & $-$0.6 & +0.3 & +0.7 & $-$0.5 & 0.0 & +0.1 & +0.2 & 0.0 & $-$0.4 & 0.0 \\
Below probability 0.4 & $-$0.2 & $-$0.1 & $-$0.9 & +0.2 & $-$0.2 & $-$0.2 & +0.1 & +0.1 & $-$0.4 & +0.1 & +0.6 & $-$0.5 & 0.0 & +0.1 & +0.2 & 0.0 & $-$0.2 & 0.0 \\
Below probability 0.6 & $-$0.3 & $-$0.2 & $-$0.3 & $-$0.4 & $-$0.1 & +0.3 & $-$0.1 & $-$0.1 & 0.0 & $-$0.4 & $-$0.2 & $-$0.2 & 0.0 & 0.0 & $-$0.5 & $-$0.2 & $-$0.2 & $-$0.2 \\
Below probability 0.7 & $-$0.5 & $-$0.7 & $-$0.1 & $-$0.7 & $-$0.4 & $-$0.2 & $-$0.3 & $-$0.2 & +0.3 & $-$0.5 & $-$0.3 & +0.2 & 0.0 & $-$0.2 & $-$0.7 & $-$0.1 & $-$0.4 & $-$0.2 \\
Below probability 0.8 & $-$1.3 & $-$1.0 & +0.2 & $-$1.0 & $-$0.1 & $-$0.7 & $-$0.1 & $-$0.3 & 0.0 & $-$0.6 & $-$0.4 & $-$0.9 & +0.1 & $-$0.3 & $-$1.1 & $-$0.2 & $-$0.7 & $-$0.4 \\
Above 5 words & 0.0 & $-$0.1 & 0.0 & $-$0.1 & +0.7 & +0.2 & 0.0 & 0.0 & 0.0 & 0.0 & 0.0 & 0.0 & 0.0 & 0.0 & 0.0 & 0.0 & +0.1 & 0.0 \\
Above 20 words & 0.0 & +0.1 & $-$0.1 & 0.0 & $-$0.6 & $-$0.1 & 0.0 & 0.0 & 0.0 & 0.0 & 0.0 & 0.0 & 0.0 & 0.0 & 0.0 & 0.0 & $-$0.1 & 0.0 \\
\bottomrule
\end{tabular}
\end{table*}

%% file: sections/appx/tab_app_split10.tex
\begin{table}[t]
\centering\scriptsize\setlength{\tabcolsep}{3.64pt}
\caption{\textbf{Splitting a 10\% budget, per benchmark.} Each column fixes the fraction of the first pass and of the re-answers, and the quota keeps the average at 10\%. The default of PRUDENT (Ours) is a first pass of 5\% and re-answers of 20\%. Values are ratios to the full-token score. \textbf{Best} in bold, \protect\underline{second}{} underlined, per row.}
\label{tab:app-split10}
\begin{tabular}{@{}lccccccccc@{}}
\toprule
 & \multicolumn{3}{c}{First pass 2.5\%} & \multicolumn{3}{c}{First pass 5\%} & \multicolumn{3}{c}{First pass 7.5\%} \\
\cmidrule(lr){2-4}\cmidrule(lr){5-7}\cmidrule(lr){8-10}
Re-answer & 20\% & 30\% & 40\% & 20\% & 30\% & 40\% & 20\% & 30\% & 40\% \\
\midrule
\multicolumn{10}{@{}l}{\textit{InternVL3-38B}} \\
ChartQA & \textbf{94.2} & \underline{92.0} & 88.8 & \textbf{94.2} & 91.9 & 89.8 & 91.2 & 90.1 & 89.9 \\
DocVQA & \underline{97.4} & 96.9 & 96.0 & \textbf{97.5} & 96.7 & 96.2 & 96.6 & 96.3 & 96.1 \\
InfoVQA & \textbf{88.7} & 85.9 & 84.4 & \underline{88.2} & 86.6 & 85.4 & 87.5 & 87.1 & 86.6 \\
TextVQA & 98.9 & \textbf{99.2} & 98.7 & \textbf{99.2} & \textbf{99.2} & \underline{99.0} & 98.9 & 98.6 & 98.6 \\
OCRB. & 96.8 & 93.6 & 91.5 & 96.6 & 96.3 & 95.0 & \textbf{97.3} & \underline{97.1} & 96.6 \\
GQA & 98.8 & \textbf{99.2} & 99.0 & \underline{99.1} & 99.0 & 98.8 & 99.0 & 98.9 & 98.9 \\
MMB & 97.3 & 98.2 & 98.2 & 97.6 & \textbf{98.5} & \underline{98.4} & 98.0 & \underline{98.4} & 98.3 \\
\midrule
\multicolumn{10}{@{}l}{\textit{InternVL3-8B}} \\
ChartQA & \underline{95.6} & 94.1 & 92.4 & \textbf{96.2} & 95.0 & 94.5 & 94.9 & 93.9 & 93.3 \\
DocVQA & \textbf{96.5} & 96.1 & 94.9 & \underline{96.4} & 95.7 & 95.2 & 94.9 & 94.7 & 94.5 \\
InfoVQA & 88.9 & 86.9 & 84.5 & \textbf{90.3} & \underline{89.4} & 88.2 & 89.1 & 88.8 & 88.3 \\
TextVQA & 100.2 & \underline{100.4} & 99.5 & \textbf{100.5} & 100.1 & 99.6 & 99.6 & 99.2 & 99.0 \\
OCRB. & 96.4 & 94.4 & 93.7 & 97.1 & 96.1 & 96.0 & \underline{97.4} & \underline{97.4} & \textbf{97.5} \\
GQA & 100.4 & 100.4 & 100.0 & 100.5 & \underline{100.8} & 100.4 & \textbf{100.9} & 100.7 & 100.4 \\
MMB & 98.0 & 97.3 & 97.4 & 98.0 & 97.3 & \underline{98.1} & \textbf{98.2} & \textbf{98.2} & 98.0 \\
\bottomrule
\end{tabular}
\end{table}

%% file: sections/appx/tab_app_where.tex
\begin{table*}[t]
\centering\scriptsize\setlength{\tabcolsep}{6.49pt}
\caption{\textbf{Where the re-answers go, per benchmark.} Re.\ is the share of all questions that PRUDENT (Ours) re-answers, as in Figure~\ref{fig:analysis}g. Hurt is the share of the re-answered questions whose first answer scores below the full-token answer, and Kept the same share among the questions whose first answer PRUDENT keeps. All values are percentages.}
\label{tab:app-where}
\begin{tabular}{@{}lcccccccccccccccccc@{}}
\toprule
 & \multicolumn{9}{c}{InternVL3-38B} & \multicolumn{9}{c}{InternVL3-8B} \\
\cmidrule(lr){2-10}\cmidrule(lr){11-19}
 & \multicolumn{3}{c}{2.5\%} & \multicolumn{3}{c}{5\%} & \multicolumn{3}{c}{10\%} & \multicolumn{3}{c}{2.5\%} & \multicolumn{3}{c}{5\%} & \multicolumn{3}{c}{10\%} \\
\cmidrule(lr){2-4}\cmidrule(lr){5-7}\cmidrule(lr){8-10}\cmidrule(lr){11-13}\cmidrule(lr){14-16}\cmidrule(lr){17-19}
Benchmark & Re. & Hurt & Kept & Re. & Hurt & Kept & Re. & Hurt & Kept & Re. & Hurt & Kept & Re. & Hurt & Kept & Re. & Hurt & Kept \\
\midrule
\multicolumn{19}{@{}l}{\textit{Text-rich}} \\
ChartQA & 25.6 & 69.7 & 11.7 & 21.3 & 58.5 & 7.6 & 21.3 & 51.8 & 3.4 & 25.6 & 62.7 & 5.4 & 21.3 & 53.1 & 4.5 & 21.3 & 34.5 & 1.8 \\
DocVQA & 25.2 & 71.7 & 6.6 & 21.0 & 59.0 & 3.9 & 21.0 & 43.0 & 1.9 & 25.2 & 73.4 & 9.1 & 21.0 & 61.7 & 5.6 & 21.0 & 48.9 & 2.5 \\
InfoVQA & 23.0 & 63.7 & 26.4 & 19.2 & 53.9 & 20.5 & 19.2 & 52.4 & 14.8 & 23.0 & 61.6 & 29.7 & 19.2 & 55.9 & 24.3 & 19.2 & 49.3 & 15.3 \\
TextVQA & 25.4 & 45.4 & 2.0 & 21.2 & 31.5 & 1.6 & 21.2 & 21.3 & 1.1 & 25.4 & 47.3 & 2.0 & 21.2 & 35.5 & 1.6 & 21.2 & 23.1 & 1.0 \\
OCRB. & 16.8 & 80.4 & 21.0 & 14.0 & 50.0 & 10.4 & 14.0 & 31.4 & 2.6 & 16.8 & 69.6 & 21.0 & 14.0 & 46.4 & 8.7 & 14.0 & 29.3 & 3.5 \\
WebSRC & 26.2 & 92.6 & 18.4 & 21.8 & 73.5 & 9.8 & 21.8 & 61.4 & 6.0 & 26.2 & 87.6 & 19.6 & 21.8 & 71.6 & 8.4 & 21.8 & 59.7 & 5.4 \\
\midrule
\multicolumn{19}{@{}l}{\textit{General}} \\
GQA & 23.3 & 25.8 & 3.0 & 19.4 & 18.2 & 2.3 & 19.4 & 13.3 & 1.8 & 23.3 & 26.1 & 2.8 & 19.4 & 19.2 & 2.3 & 19.4 & 13.8 & 1.6 \\
VQAv2 & 23.4 & 43.3 & 2.2 & 19.5 & 34.5 & 2.2 & 19.5 & 26.1 & 1.6 & 23.4 & 37.9 & 2.0 & 19.5 & 27.6 & 1.3 & 19.5 & 19.3 & 1.1 \\
OK-VQA & 17.2 & 42.8 & 16.3 & 14.4 & 27.3 & 8.4 & 14.4 & 20.0 & 4.4 & 17.2 & 27.9 & 6.5 & 14.4 & 19.7 & 4.1 & 14.4 & 14.8 & 2.6 \\
MMB & 29.1 & 30.9 & 0.9 & 24.3 & 20.2 & 0.2 & 24.3 & 14.5 & 0.0 & 29.1 & 26.6 & 0.5 & 24.3 & 18.6 & 0.3 & 24.3 & 13.8 & 0.1 \\
MMB-CN & 27.9 & 35.8 & 1.0 & 23.3 & 21.7 & 0.3 & 23.3 & 15.8 & 0.1 & 27.9 & 27.2 & 0.8 & 23.3 & 21.9 & 0.4 & 23.3 & 17.6 & 0.3 \\
MME & 29.5 & 25.3 & 1.0 & 24.6 & 16.6 & 0.6 & 24.6 & 9.8 & 0.2 & 29.5 & 19.0 & 0.6 & 24.6 & 15.2 & 0.2 & 24.6 & 9.4 & 0.1 \\
POPE & 29.8 & 14.3 & 0.1 & 24.9 & 12.7 & 0.1 & 24.9 & 10.4 & 0.1 & 29.8 & 12.3 & 0.0 & 24.9 & 9.8 & 0.0 & 24.9 & 7.7 & 0.0 \\
SEED & 28.8 & 31.4 & 1.5 & 24.0 & 24.5 & 1.1 & 24.0 & 19.3 & 0.8 & 28.8 & 29.9 & 1.4 & 24.0 & 24.4 & 1.1 & 24.0 & 19.9 & 0.7 \\
AI2D & 27.2 & 48.6 & 3.4 & 22.6 & 44.6 & 3.5 & 22.6 & 34.3 & 2.6 & 27.2 & 49.6 & 2.8 & 22.6 & 43.8 & 2.6 & 22.6 & 34.3 & 2.0 \\
SQA-I & 29.9 & 45.6 & 0.1 & 24.9 & 38.0 & 0.1 & 24.9 & 29.1 & 0.0 & 29.9 & 44.3 & 0.1 & 24.9 & 34.7 & 0.0 & 24.9 & 24.7 & 0.0 \\
\bottomrule
\end{tabular}
\end{table*}

%% file: sections/appx/tab_app_lost_wide.tex
\begin{table*}[t]
\centering\scriptsize\setlength{\tabcolsep}{9pt}
\caption{\textbf{Correct answers lost to pruning.} Each value is the share of the questions that the full-token run answers correctly and that a method answers wrongly at the same budget, pooled over the 18 benchmarks of the main tables. PRUDENT (Ours) loses the fewest. \textbf{Lowest} in bold, \protect\underline{second}{} underlined, per row.}
\label{tab:app-lost}
\begin{tabular}{@{}lcccccccc>{\columncolor{oursbg}}c@{}}
\toprule
Budget & ToMe & VisionZip & ZOO-Prune & FastV & CDPruner & MMTok & SGL & IF-Prune & \textbf{Ours} \\
\midrule
\multicolumn{10}{@{}l}{\textit{InternVL3-38B}} \\
2.5\% & 43.8 & 51.8 & 42.2 & 44.5 & 44.9 & 39.9 & 20.0 & \underline{16.0} & \textbf{11.9} \\
5\% & 40.2 & 41.8 & 35.6 & 35.7 & 38.5 & 32.5 & 13.6 & \underline{11.0} & \textbf{8.0} \\
10\% & 31.6 & 31.1 & 27.6 & 25.1 & 31.0 & 25.2 & 8.2 & \underline{7.3} & \textbf{5.3} \\
\multicolumn{10}{@{}l}{\textit{InternVL3-8B}} \\
2.5\% & 49.3 & 56.0 & 38.4 & 46.1 & 39.0 & 42.0 & 17.6 & \underline{12.5} & \textbf{8.8} \\
5\% & 44.8 & 45.3 & 30.2 & 32.1 & 31.6 & 33.6 & 11.7 & \underline{9.2} & \textbf{6.6} \\
10\% & 34.2 & 33.3 & 22.4 & 19.7 & 24.0 & 24.5 & 7.2 & \underline{6.8} & \textbf{4.7} \\
\bottomrule
\end{tabular}
\end{table*}

%% file: sections/appx/tab_app_fixed.tex
\begin{table*}[t]
\centering\scriptsize\setlength{\tabcolsep}{6.72pt}
\caption{\textbf{Questions on which PRUDENT (Ours) and IF-Prune differ.} Both run at the same budget. Fix.\ is the share of all questions that IF-Prune answers wrongly and PRUDENT correctly, and Brk.\ the reverse. Rec.\ is the share of the pruning failures of IF-Prune that PRUDENT answers correctly.}
\label{tab:app-fixed}
\begin{tabular}{@{}lcccccccccccccccccc@{}}
\toprule
 & \multicolumn{9}{c}{InternVL3-38B} & \multicolumn{9}{c}{InternVL3-8B} \\
\cmidrule(lr){2-10}\cmidrule(lr){11-19}
 & \multicolumn{3}{c}{2.5\%} & \multicolumn{3}{c}{5\%} & \multicolumn{3}{c}{10\%} & \multicolumn{3}{c}{2.5\%} & \multicolumn{3}{c}{5\%} & \multicolumn{3}{c}{10\%} \\
\cmidrule(lr){2-4}\cmidrule(lr){5-7}\cmidrule(lr){8-10}\cmidrule(lr){11-13}\cmidrule(lr){14-16}\cmidrule(lr){17-19}
Benchmark & Fix. & Brk. & Rec. & Fix. & Brk. & Rec. & Fix. & Brk. & Rec. & Fix. & Brk. & Rec. & Fix. & Brk. & Rec. & Fix. & Brk. & Rec. \\
\midrule
\multicolumn{19}{@{}l}{\textit{Text-rich}} \\
ChartQA & 25.0 & 5.3 & 62.3 & 19.7 & 4.9 & 68.1 & 10.9 & 3.6 & 68.3 & 20.4 & 4.2 & 65.7 & 12.2 & 3.9 & 60.3 & 8.2 & 2.6 & 61.5 \\
DocVQA & 10.3 & 3.2 & 63.4 & 7.1 & 2.1 & 61.7 & 5.0 & 1.5 & 67.4 & 12.2 & 2.9 & 61.8 & 9.4 & 2.4 & 61.4 & 6.8 & 1.9 & 64.0 \\
InfoVQA & 14.1 & 9.2 & 43.8 & 11.6 & 7.4 & 48.1 & 8.7 & 6.1 & 45.5 & 15.1 & 8.3 & 44.3 & 12.7 & 7.7 & 49.7 & 11.2 & 5.5 & 53.2 \\
TextVQA & 8.0 & 2.7 & 61.8 & 5.2 & 2.1 & 55.6 & 3.3 & 1.9 & 50.0 & 7.8 & 2.5 & 59.3 & 5.3 & 1.7 & 53.6 & 4.0 & 1.5 & 60.2 \\
OCRBench & 8.6 & 7.7 & 34.4 & 6.8 & 4.5 & 49.6 & 3.7 & 2.4 & 49.2 & 9.9 & 8.1 & 41.5 & 6.4 & 4.8 & 48.7 & 4.8 & 2.8 & 60.0 \\
WebSRC & 12.3 & 5.2 & 54.0 & 8.2 & 3.6 & 55.9 & 4.8 & 2.5 & 52.9 & 13.9 & 5.0 & 54.6 & 10.4 & 3.9 & 57.3 & 8.1 & 2.6 & 59.5 \\
\midrule
\multicolumn{19}{@{}l}{\textit{General}} \\
GQA & 5.9 & 4.1 & 51.3 & 4.4 & 3.1 & 47.7 & 3.2 & 2.7 & 43.7 & 6.1 & 3.8 & 54.0 & 4.4 & 2.9 & 49.9 & 2.7 & 2.3 & 41.3 \\
VQAv2 & 8.1 & 4.5 & 49.5 & 6.9 & 3.1 & 51.2 & 5.8 & 2.3 & 54.6 & 5.0 & 2.6 & 47.2 & 3.8 & 1.9 & 46.9 & 2.6 & 1.6 & 45.6 \\
OK-VQA & 7.4 & 4.2 & 34.2 & 4.6 & 2.5 & 33.0 & 3.0 & 1.9 & 36.3 & 4.6 & 4.4 & 40.1 & 3.8 & 2.8 & 42.3 & 2.4 & 1.7 & 34.1 \\
MMB & 2.3 & 1.2 & 38.9 & 1.5 & 0.9 & 34.3 & 1.4 & 0.9 & 38.7 & 2.1 & 1.1 & 33.8 & 1.2 & 0.8 & 27.6 & 1.5 & 0.5 & 39.0 \\
MMB-CN & 2.6 & 1.2 & 39.0 & 1.8 & 0.9 & 31.6 & 1.7 & 0.9 & 45.1 & 2.1 & 1.3 & 27.2 & 1.3 & 0.9 & 21.2 & 1.4 & 0.8 & 24.7 \\
MME & 2.4 & 1.1 & 45.9 & 1.3 & 0.8 & 40.3 & 0.9 & 0.9 & 47.4 & 2.3 & 1.2 & 46.9 & 1.0 & 0.5 & 35.6 & 1.0 & 0.7 & 42.9 \\
POPE & 1.5 & 1.2 & 39.2 & 1.2 & 0.8 & 34.6 & 0.7 & 0.6 & 27.0 & 1.2 & 0.7 & 35.6 & 0.7 & 0.8 & 30.9 & 0.6 & 0.6 & 34.0 \\
SEED & 3.1 & 2.1 & 33.7 & 1.9 & 1.3 & 27.3 & 1.8 & 1.1 & 29.4 & 3.0 & 1.7 & 33.0 & 1.9 & 1.4 & 25.4 & 1.8 & 1.1 & 28.2 \\
AI2D & 5.9 & 2.6 & 35.4 & 4.5 & 2.1 & 34.5 & 3.4 & 1.8 & 33.7 & 5.6 & 2.7 & 33.9 & 4.2 & 2.3 & 32.4 & 3.5 & 1.5 & 34.9 \\
SQA-I & 3.8 & 1.4 & 38.3 & 3.6 & 1.0 & 47.9 & 2.1 & 0.6 & 45.3 & 3.7 & 0.8 & 41.5 & 2.4 & 0.8 & 38.4 & 2.6 & 1.0 & 53.7 \\
\midrule
\multicolumn{19}{@{}l}{\textit{Grounding}} \\
RefC-A & 13.0 & 9.4 & 45.4 & 8.6 & 5.9 & 50.5 & 5.3 & 3.0 & 55.1 & 6.0 & 3.5 & 57.1 & 3.6 & 2.7 & 59.1 & 2.3 & 1.4 & 52.4 \\
RefC-B & 17.4 & 9.7 & 45.4 & 10.8 & 6.9 & 48.5 & 8.1 & 4.2 & 51.2 & 8.2 & 6.2 & 56.6 & 5.9 & 4.4 & 55.3 & 4.8 & 2.9 & 56.0 \\
\bottomrule
\end{tabular}
\end{table*}